\documentclass[10pt,twocolumn,letterpaper]{article}

\usepackage[pagenumbers]{cvpr} %

\usepackage{epsfig}
\usepackage{graphicx}
\usepackage{amsmath}
\usepackage{amssymb}
\usepackage{booktabs}
\usepackage{epigraph}
\usepackage{url}            %
\usepackage{amsfonts}       %
\usepackage{nicefrac}       %
\usepackage{xcolor,colortbl}         %
\usepackage{xspace}
\usepackage{multirow}
\usepackage{mathtools}
\usepackage{overpic}
\usepackage{algpseudocode}
\usepackage{algorithm}

\usepackage{adjustbox}

\usepackage{caption}
\usepackage{dsfont}
\usepackage{soul}

\usepackage{array}
\newcolumntype{H}{>{\setbox0=\hbox\bgroup}c<{\egroup}@{}}

\DeclareMathOperator*{\argmax}{argmax}

\newcommand{\PAR}[1]{\vskip4pt \noindent {\bf #1~}}
\newcommand{\PARit}[1]{\vskip4pt \noindent {\it #1~}}

\newcommand{\method}{\textbf{SAL-4D}\@\xspace}

\newcommand{\task}{\textit{Zero-Shot 4D Lidar Panoptic Segmentation}\@\xspace}

\newcommand{\taskabbrev}{\textit{ZS-4D-LPS}\@\xspace}

\newcommand{\fdpls}{\textit{4D Panoptic Lidar Segmentation}\@\xspace}
\newcommand{\fdlps}{\textit{4D Lidar Panoptic Segmentation}\@\xspace}

\newcommand{\thing}{\textit{thing}\@\xspace}
\newcommand{\stuff}{\textit{stuff}\@\xspace}
\newcommand{\lps}{\textit{Lidar Panoptic Segmentation}\@\xspace}

\definecolor{red}{rgb}{0.95,0.4,0.4}
\definecolor{purered}{rgb}{1,0,0}
\definecolor{darkblue}{rgb}{0,0,0.8}
\definecolor{darkred}{rgb}{1,0,0}
\definecolor{darkgreen}{rgb}{0,0.5,0}
\definecolor{grey}{rgb}{0.6,0.6,0.6}
\definecolor{col1}{RGB}{232, 161, 148}
\definecolor{col2}{RGB}{148, 187, 232}
\definecolor{lightgrey}{rgb}{0.85,0.85,0.85}
\definecolor{lightlightgrey}{rgb}{0.9,0.9,0.9}
\definecolor{verylightBG}{rgb}{0.9,0.99,0.99}
\definecolor{darkgreen}{rgb}{0.3, 0.75, 0.3}
\definecolor{darkgrey}{rgb}{0.8,0.35,0.35}

\usepackage{array}
\newcolumntype{H}{>{\setbox0=\hbox\bgroup}c<{\egroup}@{}}
\newcolumntype{Z}{>{\setbox0=\hbox\bgroup}c<{\egroup}@{\hspace*{-\tabcolsep}}}

\definecolor{cvprblue}{rgb}{0.21,0.49,0.74}
\usepackage[pagebackref,breaklinks,colorlinks,citecolor=cvprblue]{hyperref}

\newif\ifarxiv

\newcommand{\arxiv}[1]{\ifarxiv{#1}\fi}
\newcommand{\conf}[1]{\ifarxiv\else{#1}\fi}

\def\paperID{1331} %
\def\confName{CVPR}
\def\confYear{2025}

\title{Zero-Shot 4D Lidar Panoptic Segmentation}

\author{
Yushan Zhang$^{1,2}$\thanks{Correspondance to \tt\small yushan.zhang@liu.se.}
\quad
Aljoša Ošep$^{1}$
\quad
Laura Leal-Taixé$^{1}$
\quad
Tim Meinhardt$^{1}$
\\
$^1$\text{NVIDIA}
\hspace{1cm}
$^2$\text{Linköping University}
}

\begin{document}

\twocolumn[{%
\renewcommand\twocolumn[1][]{#1}%
\maketitle
\captionsetup{type=figure}
\begin{center}
\vspace{-0.5cm}

\captionsetup[subfigure]{labelformat=empty,justification=centering}

\subcaptionbox{\textbf{\textit{Prior work:}}  Zero-Shot (3D) \\Lidar Panoptic Segmentation \label{fig:teaser_1}}[.23\textwidth]{%
  \begin{overpic}[width=\linewidth]{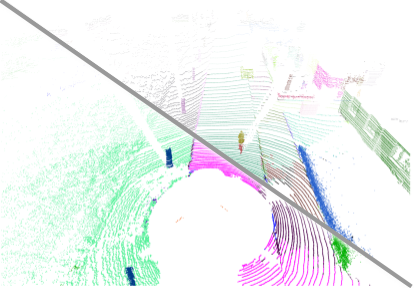}
  \put(57,58){{\small \colorbox{gray!10}
    {\scriptsize \textit{3D Instances}}}}
  \put(5,5){{\small \colorbox{gray!10}
    {\scriptsize \textit{Semantics}}}}
  \end{overpic}
}\hspace{1mm}
\subcaptionbox{\textbf{\textit{This work}}: Zero-Shot 4D \\Lidar Panoptic Segmentation \label{fig:teaser_2}}[.23\textwidth]{%
  \begin{overpic}[width=\linewidth]{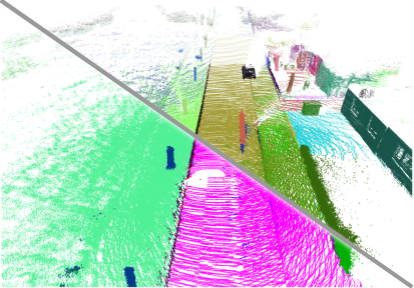}    \put(57,58){{\small \colorbox{gray!10}
    {\scriptsize \textit{4D Instances}}}}
  \put(5,5){{\small \colorbox{gray!10}
    {\scriptsize \textit{Semantics}}}}
  \end{overpic}
}\hspace{0.3mm}
\subcaptionbox{Text prompts: \\ \{\texttt{advertising stand}\}\label{fig:teaser_3}}[.23\textwidth]{%
  \includegraphics[width=\linewidth]{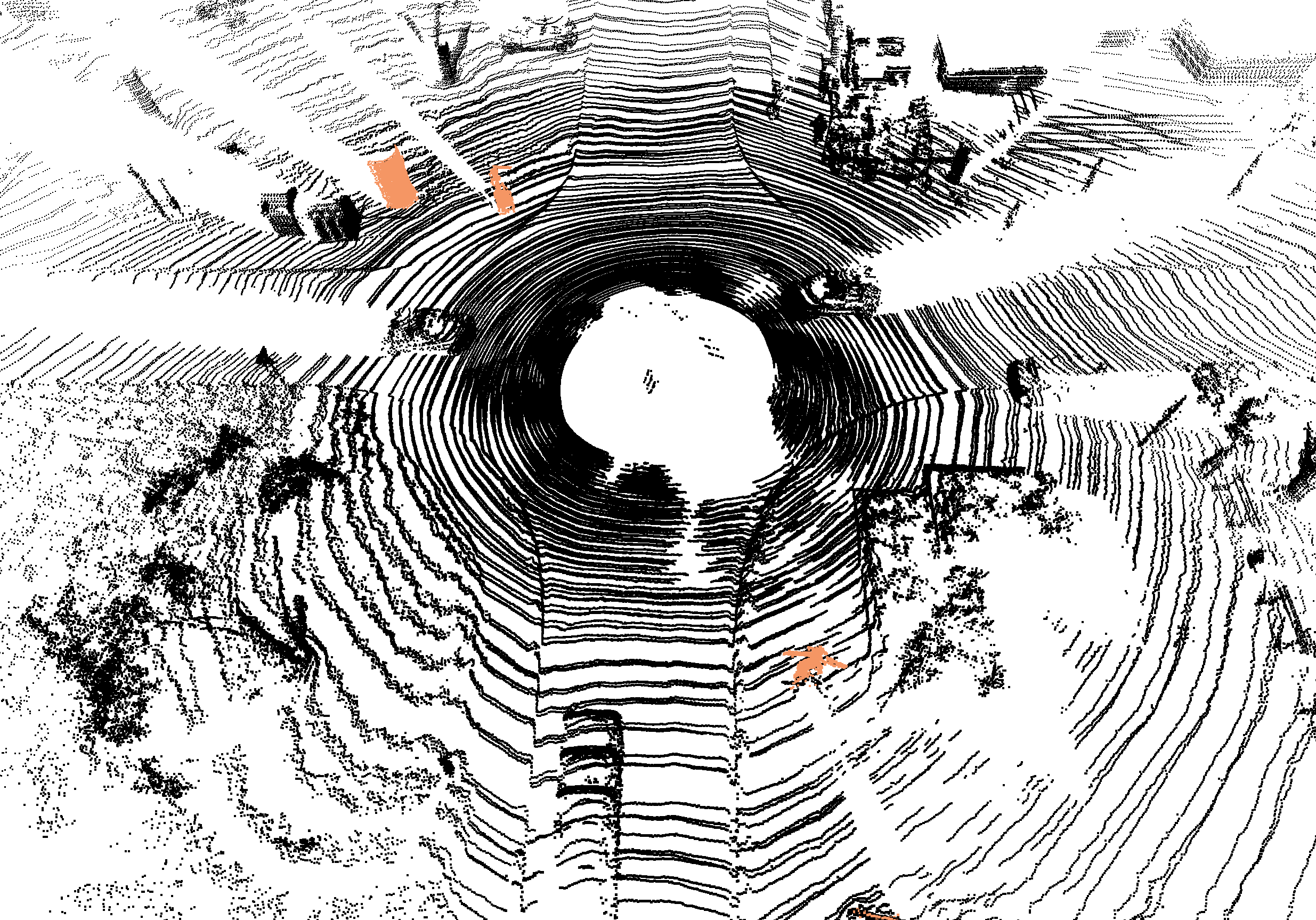}
}\hspace{0.3mm}
\subcaptionbox{Text prompts: \\ \{\texttt{car}\}\label{fig:teaser_4}}[.23\textwidth]{%
  \includegraphics[width=\linewidth]{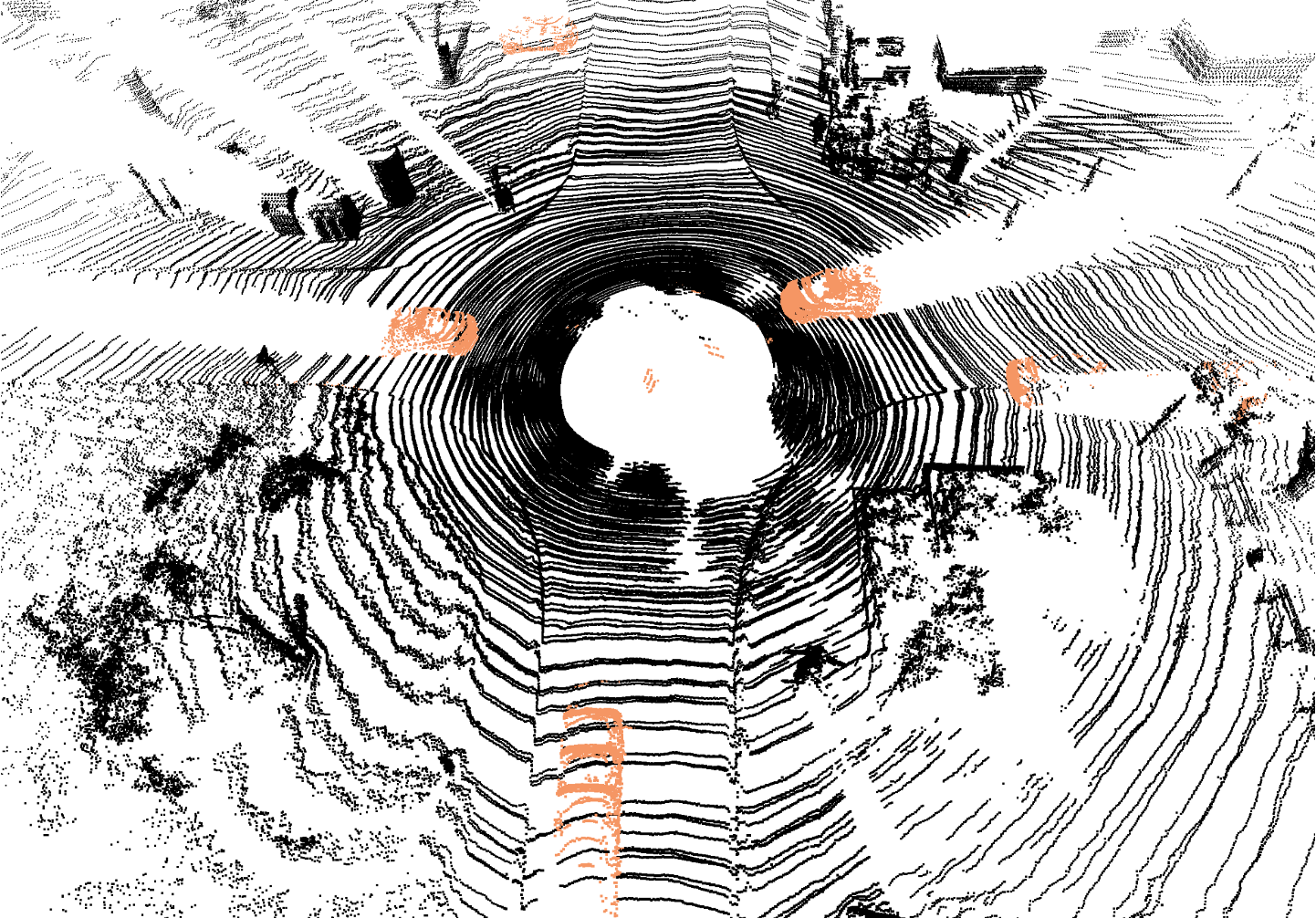}
}

\vspace{-0.2cm}
\caption{\textbf{Learning to Segment Anything in Lidar--4D:} 
Prior methods (\textit{left}) for zero-shot Lidar panoptic segmentation process individual (3D) point clouds in isolation. In contrast, our data-driven approach (\textit{right}) operates directly on sequences of point clouds, jointly performing object segmentation, tracking, and zero-shot recognition based on text prompts specified at test time. 
Our method localizes and tracks \textit{any} object and provides a temporally coherent semantic interpretation of dynamic scenes. We can \textit{correctly} segment canonical objects, such as \texttt{car}, and objects beyond the vocabularies of standard Lidar datasets, such as \texttt{advertising stand}. 
\textit{Best seen in color, zoomed.}
}
\label{fig:teaser}
\end{center}
}
]
\renewcommand{\thefootnote}{\fnsymbol{footnote}}
\footnotetext[1]{Work done during an internship at NVIDIA.}
\renewcommand{\thefootnote}{\arabic{footnote}}

\begin{abstract}
Zero-shot 4D segmentation and recognition of arbitrary objects in Lidar is crucial for embodied navigation, with applications ranging from streaming perception to semantic mapping and localization. 
However, the primary challenge in advancing research and developing generalized, versatile methods for spatio-temporal scene understanding in Lidar lies in the scarcity of datasets that provide the necessary diversity and scale of annotations.
To overcome these challenges, we propose \method (Segment Anything in Lidar--4D), a method that utilizes multi-modal robotic sensor setups as a bridge to distill recent developments in Video Object Segmentation (VOS) in conjunction with off-the-shelf Vision-Language foundation models to Lidar.  
We utilize VOS models to pseudo-label tracklets in short video sequences, annotate these tracklets with sequence-level CLIP tokens, and lift them to the 4D Lidar space using calibrated multi-modal sensory setups to distill them to our \method model. 
Due to temporal consistent predictions, we outperform prior art in 3D Zero-Shot Lidar Panoptic Segmentation (LPS) over $5$ PQ, and unlock Zero-Shot 4D-LPS.
\end{abstract}

\section{Introduction}

We tackle segmentation, tracking, and zero-shot recognition of any object in Lidar sequences. Such open-ended 4D spatio-temporal scene understanding is directly relevant for embodied navigation~\cite{thrun2006stanley}, semantic mapping~\cite{behley2019iccv, behley18rss, Wang_2023_CVPR},
localization~\cite{kolmet2022text2pos, huang2024semantics} and neural rendering~\cite{Ost_2021_CVPR}.

\PAR{Status quo.} %
In applications that demand precise spatial and dynamic situational scene understanding, \eg, autonomous driving~\cite{thrun2006stanley}, perception stacks 
rely on Lidar-based object detection~\cite{zhou2018voxelnet,yan2018second,liu2021iccv} and multi-object tracking~\cite{yin2021center, kim2022polarmot, Weng20iros, ding20233dmotformer} methods to localize objects, with recent trends moving towards holistic scene understanding via \fdlps (4D-LPS)~\cite{aygun21cvpr}. The progress in these areas has largely been fueled by data-driven methods~\cite{qi2017pointnet,qi2017pointnet++,choy20194d,Thomas19ICCV} that rely on manually labeled datasets~\cite{behley2019iccv, fong21ral, sun20CVPR}, limiting these methods to localizing instances of predefined object classes. On the other hand, recent developments in single-scan Lidar-based perception are moving towards utilizing vision foundation models for pre-training~\cite{sautier2022image, puy2023revisiting, puy2024three} and zero-shot segmentation~\cite{sal2024eccv,Peng2023OpenScene,xiao20243d}.
However, state-of-the-art methods can only detect~\cite{najibi2023unsupervised} and segment~\cite{sal2024eccv,xiao20243d} objects in individual scans. In contrast, embodied agents must continuously interpret sensory data and localize objects in a 4D continuum to understand the present and predict the future. 

\PAR{Towards 4D pseudo-labeling.} 
\textit{Can we perform \fdlps by distilling video-foundation models to Lidar?} %
Recent advances~\cite{ravi2024sam} suggest that Video Object Segmentation (VOS)~\cite{PontTuset17arxiv} generalize well to arbitrary objects. However, empirically, long-term segmentation stability remains a challenge~\cite{zhang2024evaluation, ding2024sam2long}, while data recorded from moving platforms presents unique challenges, such as rapid (ego) motion, objects commonly entering and exiting sensing areas, and frequent occlusions.

To train our \method for \textit{Zero-Shot \fdlps}, we present a pseudo-labeling engine that is built on the insight that we can reliably prompt state-of-the-art VOS models over short temporal horizons in videos and generate their corresponding sequence-level CLIP features to facilitate zero-shot recognition. 
To account for inherently noisy localization and possible tracking errors, we lift these masklets, localized in the video, to Lidar, where we leverage accurate spatial Lidar localization to associate masklets across windows and \textit{continually} localize individual object instances as they enter and leave the sensing area. Therefore, our pseudo-labeling engine provides precisely the supervisory signal for 4D Lidar segmentation models~\cite{aygun21cvpr,yilmaz2024mask4former}. 
Even though our pseudo-labeling approach is still prone to noise and errors, we empirically observe that they are sufficiently de-correlated, enabling us to distill a noisy supervisory signal into a strong, end-to-end trainable Lidar segmentation model that can segment, track, and recognize objects anywhere in Lidar in the absence of image features.

\PAR{Key findings.} 
Our method significantly improves the zero-shot recognition capabilities compared to the single-scan state-of-the-art \lps~\cite{sal2024eccv} due to temporal coherence, and, more importantly, \method unlocks new capabilities in Lidar perception. For the first time, we can segment objects beyond the predefined object classes of typical 4D-LPS benchmarks in a temporally coherent manner and open the door for future progress in learning to segment anything in Lidar sequences. 

\PAR{Main contributions.} We present the (i) first study on \task, and discuss multiple possible approaches for this task. 
Our analysis (ii) paves the road for a strong baseline, \method, that utilizes vision foundation models to construct temporal consistent annotations, that, when distilled to Lidar, allow us to segment, track, and recognize arbitrary objects. 
We (iii) thoroughly ablate our design decisions and analyze the remaining gap to supervised models on standard benchmarks.

\section{Related Work}
\label{sec:related}

This section discusses recent developments in segmentation, tracking, and zero-shot recognition in Lidar. 

\PAR{Lidar panoptic segmentation.}
Thanks to the advent of manually labeled Lidar-based datasets~\cite{behley2019iccv,fong21ral,le2024jrdb} we have made rapid progress in single-scan semantic~\cite{xiong11icra,Wu18ICRA,Wu19ICRA,Milioto19IROS,aksoy2020salsanet,razani2021lite,li2021multi,choy20194d,tang2020spvnas,zhu2020cylindrical,Agarwalla23iros} and panoptic segmentation~\cite{Behley21icra, zhou2021panoptic,hong2021lidar,razani2021gp,li2022panoptic, gasperini2020panoster} via supervised learning. In this setting, the task is to learn to classify points into a set of pre-defined semantic classes that follow class vocabulary defined prior to the data annotation process.

This formulation limits types of classes that can be recognized or segmented as individual instances. 
As labeled Lidar data is scarce, \cite{Peng2023OpenScene,xiao20243d} lift image features to 3D for zero-shot semantic \cite{Peng2023OpenScene} and panoptic \cite{xiao20243d} segmentation. Different from \cite{najibi2023unsupervised, sal2024eccv}, these are limited to segmenting Lidar points that are co-visible in cameras. \cite{najibi2023unsupervised} addresses zero-shot object detection for traffic participants, a subset of \textit{thing} classes, and SAL~\cite{sal2024eccv} distills vision foundation models to Lidar to segment and recognize instances of \thing and \stuff classes. 
However, all aforementioned can only segment individual scans, whereas temporal interpretation of sensory data is pivotal in embodied perception.

\PAR{Object tracking.} 
Multi-object tracking (MOT) is a long-standing problem commonly used for spatio-temporal understanding of Radar~\cite{Reid79TAC}, image~\cite{Zhang08CVPR,Leibe08TPAMI, dendorfer20ijcv}, and Lidar~\cite{Petrovskaya09AR} data. %
It is commonly addressed via tracking-by-detection, where an object detector is first trained for a pre-defined set of object classes~\cite{yin2021center, Lang19CVPR, zhou2018voxelnet,yan2018second,liu2021iccv}, that localize objects in individual frames, followed by cross-frame association. Image-based methods rely on learning robust appearance models~\cite{dendorfer20ijcv,seidenschwarz2023simple}, whereas Lidar-based trackers leverage accurate 3D localization in Lidar and rely on motion and geometry~\cite{kim2022polarmot,Weng20iros,yin2021center,ding20233dmotformer}. 
Unlike our pursuit of joint zero-shot segmentation and tracking of \textit{any} object, prior Lidar-based tracking methods focus on the cross-detection association to track instances of pre-defined classes as bounding boxes. 

Related to our work is class-agnostic multi-object tracking in videos~\cite{li2022tracking, Osep18ICRA, Dave19ICCVW, Osep20ICRA,liu2022opening}, recently addressed in conjunction with zero-shot recognition~\cite{li2023ovtrack,chu2024zero}. Like ours, these methods must track and, optionally, classify objects as they enter and exit the sensing area. In contrast to ours, these rely on (at least some) labeled data available in the image domain and focus on tracking \textit{thing} classes. These are also related to methods for single object tracking based on spatial prompts (Visual Object Tracking~\cite{kristan2016novel,kristan2015visual,wu2013online,huang2019got} and Video Object Segmentation~\cite{Perazzi16CVPR,Xu18ECCV}), which we utilize \cite{ravi2024sam} in our pseudo-labeling pipeline (\cref{subsec:pseudo-label-engine}).

\PAR{4D Lidar panoptic segmentation.}
4D Lidar Panoptic Segmentation~\cite{aygun21cvpr} addresses holistic, spatio-temporal understanding of (4D) Lidar data. Contemporary methods approach this task by segmenting short spatio-temporal (4D) volumes~\cite{aygun21cvpr, kreuzberg20224d, zhu20234d, yilmaz2024mask4former,hong2024unified,chen2023svqnet,athar20234d,wu2024taseg}, followed by cross-volume fusion, or follow the tracking-by-detection paradigm, established in MOT~\cite{hurtado2020mopt,Agarwalla23iros, marcuzzi2022contrastive, marcuzzi2023mask4d}. 
The aforementioned methods utilize manual supervision in the form of semantic spatio-temporal instance labels and are confined to pre-defined class vocabularies. 
Exceptions are early efforts, such as~\cite{Teichman11ICRA, Moosmann13ICRA, Held16RSS, Mitzel12ECCV, Osep16ICRA, Kochanov16IROS}, that utilize heuristic bottom-up grouping methods to segment arbitrary objects in individual Lidar scans, followed by tracking, and, optionally, semantic recognition of tracked objects (for which semantic annotations are available). 
Our approach follows the same principle and performs class-agnostic segmentation and tracking of any object in Lidar. However, we learn via self-supervision to track, segment, and recognize any object that occurs in the training data. 

\PAR{Zero-shot learning.}
Zero-shot learning (ZSL)~\cite{Xian18TPAMI} methods must recognize object classes for which labeled training data may not be available. \textit{Inductive} methods assume no available information about the target classes, whereas \textit{transductive} setting only restricts access to labels. We address 4D Lidar segmentation in \textit{transductive} setting, as usual in tasks beyond image recognition (\eg, object detection~\cite{Bansal18ECCV,miller18ICRA,Rahman18ACCV}, semantic/panoptic segmentation~\cite{ding2023open, xu2023open, bucher2019zero}), where imposing restrictions on the presence of semantic classes in images would be impractical.
Similarly to contemporary image-based methods~\cite{gu2021open, zareian2021open, zhong2022regionclip,li2022languagedriven,ghiasi2022scaling,rao2022denseclip, zhou2022maskclip, liang2023open, xu2023side,yuan2024open}, we rely on CLIP~\cite{radford2021learning} for zero-shot recognition of objects, however, we distill CLIP features directly to point cloud sequences. 
Our work is related to open-set recognition~\cite{scheirer2012toward} and open-world~\cite{bendale15CVPR} learning, which recognize classes not shown as labeled instances during the model training.

\section{Zero-Shot 4D Lidar Panoptic Segmentation}
\label{sec:method}

\begin{figure*}[ht]
    \begin{subfigure}[t]{0.35\linewidth}
        \centering
        \includegraphics[height=6cm]{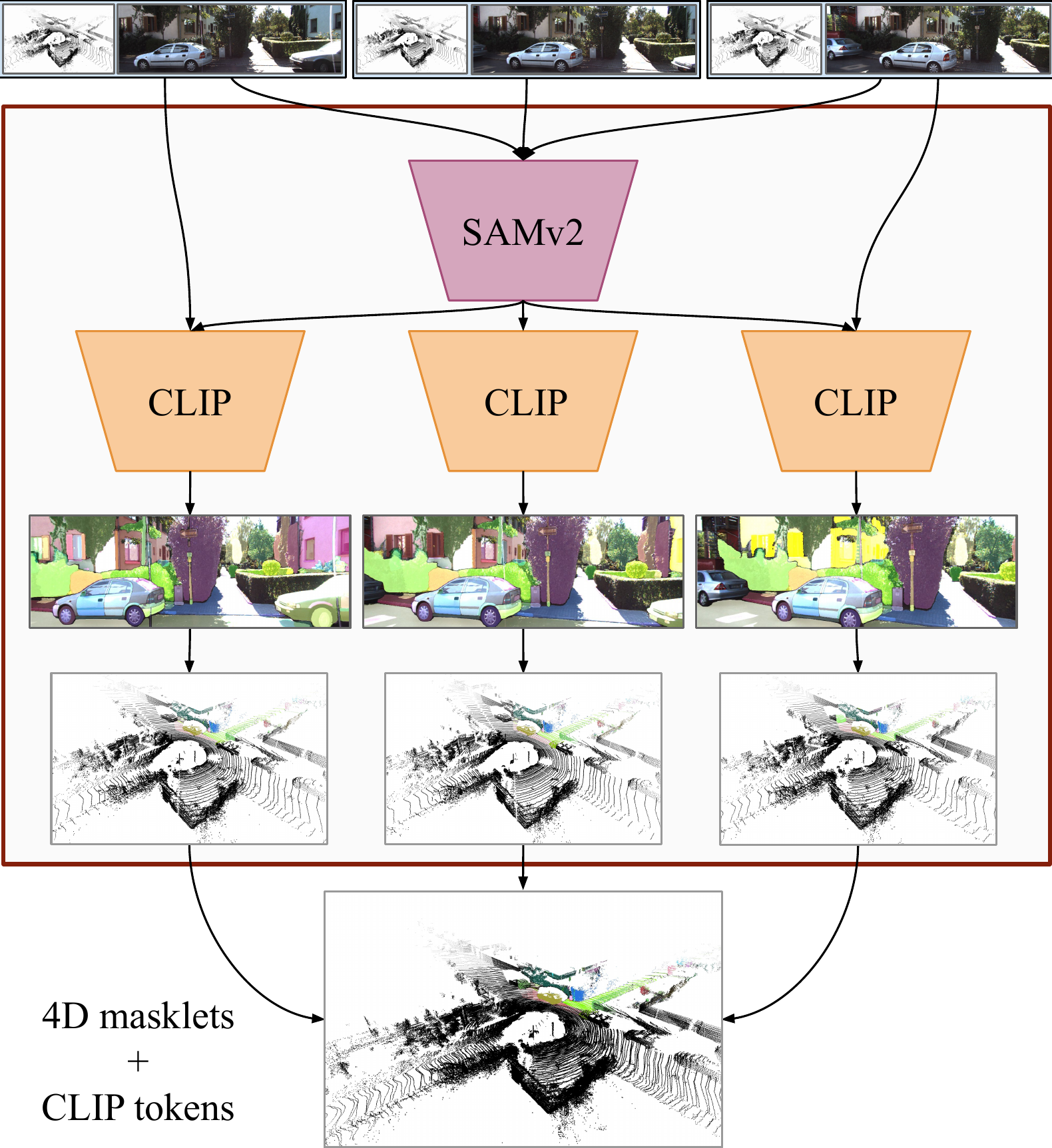}
        \caption{\textbf{Track--Lift--Flatten.}}
        \label{fig:sal4d-window}
    \end{subfigure}%
    \hfill
    \begin{subfigure}[t]{0.59\linewidth}
        \centering
        \includegraphics[height=6cm]{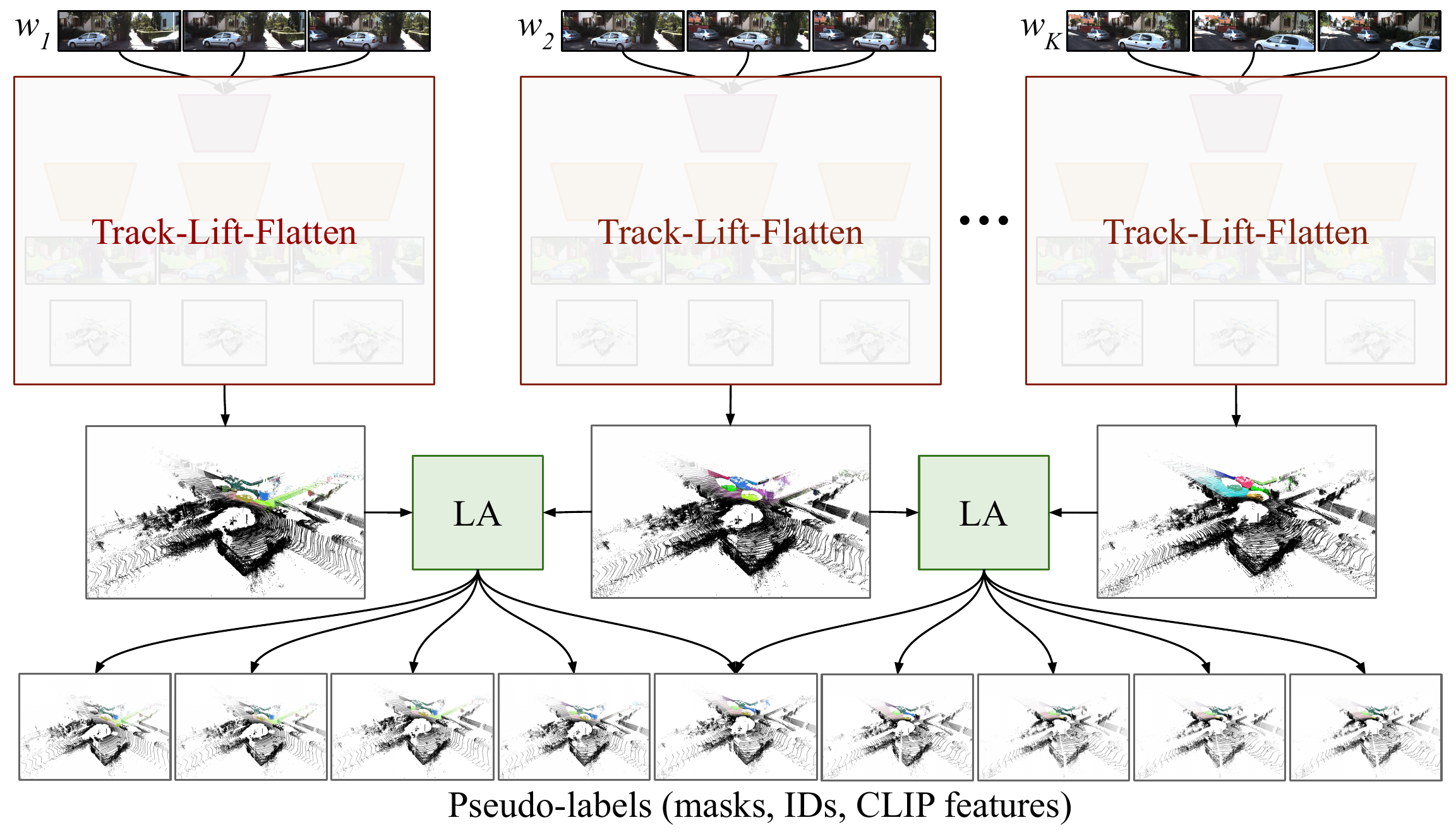}
        \caption{\textbf{Cross-window association.} }
        \label{fig:sal4d-cross-window}
    \end{subfigure}
    \caption{\textbf{\method pseudo-label engine.} 
    We first independently pseudo-label overlapping sliding windows (\cref{fig:sal4d-window}). We track and segment objects in the video using \cite{ravi2024sam}, generate their semantic features using CLIP, and lift labels from images to 4D Lidar space. Finally, we ``flatten'' masklets to obtain a unique non-overlapping set of masklets in Lidar for each temporal window. We associate masklets across windows via linear assignment (LA) to obtain pseudo-labels for full sequences and average their semantic features (\cref{fig:sal4d-cross-window}).}
    \label{fig:sal4d-autolabel-engine}
\end{figure*}

In this section, we formally state the \fdlps (4D-LPS) task and discuss its generalization to zero-shot setting (\cref{subsec:problem_statement}) for joint segmentation, tracking and recognition of \textit{any} object in Lidar. 
In \cref{subsec:model}, we describe our concrete instantiation of this approach, \method. 

\subsection{Problem Statement}
\label{subsec:problem_statement}

\PAR{4D Lidar panoptic segmentation.}
Let $\mathcal{P} = \{ P_t \}_{t=1}^T$ be a sequence of $T$ point clouds, where each $P_t \in \mathbb{R}^{N_t\times 4}$ is a point cloud observed at time $t$ containing $N_t$ points that consist of spatial coordinates and an intensity value.
For each point $p$, 4D-LPS methods estimate a semantic class $c \in \{1, \ldots, L\}$ with $L$ predefined classes, and an instance $\text{id} \in \mathbb{N}$ for \thing classes, or $\emptyset$ for \stuff classes. 
To this end, a function $f_\theta$, representing the segmentation model with parameters $\theta$, is usually trained on manually-labeled dataset $\mathcal{D}_{\text{train}}$ by minimizing an appropriate loss function. %

\PAR{Zero-shot 4D Lidar panoptic segmentation.} 
We address 4D-LPS in a zero-shot setting, intending to localize and recognize \textit{any} objects in 4D Lidar point cloud sequences. 
Similarly, we assign \textit{each} points $p \in  \mathcal{P}$ an instance identity $\text{id} \in \mathbb{N}$; however, we do not assume predefined semantic class vocabulary and (accordingly) labeled training set at train time.  
Instead, we assume a semantic vocabulary $\mathcal{C}_{test}$ is \textit{optionally} specified at test-time as a list of free-form descriptions of semantic classes. When specified, we assign points also to semantic classes $c \in \mathcal{C}_{test}$. 
As the separation between \thing and \stuff classes cannot be specified \textit{prior} to the model training, we drop this distinction.

\PAR{Method overview.} Our \method consists of two core components: (i) The \textbf{pseudo-label engine} (\cref{fig:sal4d-autolabel-engine}) constructs a proxy dataset $\mathcal{D}_{\text{proxy}}$, that consists of Lidar data and self-generated pseudo-labels that localize individual spatio-temporal instances and their semantic features. %
(ii) The \textbf{model} $f_\theta$ (\cref{fig:sal4d-model}) learns to segment individual instances in fixed-size 4D volumes by minimizing empirical risk on our proxy dataset $\mathcal{D}_{\text{proxy}}$. Our model and proxy dataset are constructed such that our model learns to segment and recognize a super-set of all objects labeled in existing datasets. %

\subsection{SAL-4D Pseudo-label Engine}
\label{subsec:pseudo-label-engine}

Our pseudo-label engine (\cref{fig:sal4d-autolabel-engine}) operates with a multi-modal sensory setup. We assume an input Lidar sequence $\mathcal{P} = \{ P_t \}_{t=1}^T$ along with $C$ unlabeled videos $\mathcal{V} = \{ \mathcal{V}^c \}_{c=1}^C$, where each video $\mathcal{V}^c = \{ I_t^c \}_{t=1}^T$ consists of images $I_t^c \in \mathbb{R}^{H \times W \times 3}$ of spatial dimensions $H \times W$, captured by camera $c$ at time $t$.
For each point cloud $P_t$, we produce pseudo-labels, comprising of tuples $\{ \tilde{m}_{i,t}, \text{id}_i, f_i \}_{i=1}^{M_t}$, where $\tilde{m}_{i,t} \in \{0, 1\}^{N_t}$ represents the binary segmentation mask for instance $i$ at time $t$ in the point cloud $P_t$, and $\text{id}_i \in \mathbb{N}$ is the unique object identifier for spatio-temporal instance $i$.  %
Finally, $f_i \in \mathbb{R}^d$ represents instance semantic features aggregated over time.

\subsubsection{Track--Lift--Flatten}
\label{subsubsec:track-lift-splat}

We proceed by sliding a temporal window of size $K$ with a stride $S$ over the sequence of length $T$. We first pseudo-label each temporal window (see Figure~\ref{fig:sal4d-window}), and then perform cross-window association (see Figure~\ref{fig:sal4d-cross-window}) to obtain pseudo-labels for sequences of arbitrary length. 
In a nutshell, for each temporal window, we track objects in video (\textit{track}), lift masks to 4D Lidar sequences (\textit{lift}), and, finally, ``\textit{flatten}'' overlapping masklets in the 4D volume. 
Our temporal windows \( w_k = \{ (P_t, I_t) \mid t \in T_k \} \) consist of Lidar point clouds and images over specific time frames. Here, \( T_k = \{ t_k, t_k + 1, \ldots, t_k + K - 1 \} \) is the set of time indices for window \( w_k \).
We drop the camera index \( c \) unless needed. 

\PAR{Track.}
For each video, we use a segmentation foundation model~\cite{kirillov2023segment} to perform grid-prompting in the first video frame of the window \( I_{t_k} \) to localize objects as masks \( \{ m_{i,t_k} \}_{i=1}^{M_{t_k}} \), \( m_{i,t_k} \in \{0,1\}^{H \times W} \), where $M_{t_k}$ denotes the number of discovered instances in $I_{t_k}$. 
We then propagate  masks through the entire window \( \{ I_t \mid t \in T_k \} \) using~\cite{ravi2024sam} to obtain masklets \( \{ m_{i,t} \mid t \in T_k \}_{i=1}^{M_{t_k}} \) for all instances discovered in ${I_{t_k}}$. 
This results in $M_{t_k}$ overlapping masklets in a 3D video volume of dimensions \( H \times W \times K \), representing objects visible in \( I_{t_k} \) across the window \( w_k \).

Given masklets  \(\{ m_{i,t} \mid t \in T_k \}_{i=1}^{M_{t_k}} \) and corresponding images \( \{ I_t \mid t \in T_k \} \), we compute semantic features \( f_{i,t} \) for each mask \( m_{i,t} \) using relative mask attention in the CLIP~\cite{radford2021learning} 
feature space and obtain masklets paired with their CLIP features \( \{ ( m_{i,t}, \text{id}_{i,k}, f_{i,t} ) \mid t \in T_k \} \) for each instance \( i \), where \( \text{id}_{i,k} \) is a local instance identifier within window \( w_k \). For details, we refer to \conf{Appendix A.1}\arxiv{\cref{appendix:pseudo_label_engine}}.

\PAR{Lift.}
We associate 3D points \( \{ P_t \mid t \in T_k \} \) with image masks \( m_{i,t} \) via Lidar-to-camera transformation and projection. 
We refine our lifted Lidar masklets to address sensor misalignment errors using density-based clustering~\cite{Ester96KDD}. We create an ensemble of DBSCAN clusters by varying the density parameter and replacing all lifted masks with DBSCAN masks with sufficient intersection-over-union (IoU) overlap~\cite{sal2024eccv}. We obtained the best results by performing this on a single-scan basis (\conf{Appendix C.1}\arxiv{\cref{appendix:subsection:pseudo_label_engine_ablation}}).  

We obtain sets \( \{ ( \tilde{m}_{i,t}^c, \text{id}_{i,k}^c, f_{i,t}^c ) \mid t \in T_k \} \) independently for each camera \( c \), and fuse instances with sufficient IoU overlap across cameras. We fuse their semantic features \( f_{i,t} \) via mask-area-based weighted average to obtain a set of tuples \( \{ ( \tilde{m}_{i,t}, \text{id}_{i,k}, f_{i,t} ) \mid t \in T_k \} \), that represent spatio-temporal instances localized in window \( w_k \).

\PAR{Flatten.}
The resulting set contains overlapping masklets in 4D space-time volume. To ensure each point is assigned to at most one instance, we perform spatio-temporal flattening as follows.
We compute the spatio-temporal volume \( V_i \) of each masklet \( \tilde{M}_i = \{ \tilde{m}_{i,t} \mid t \in T_k \} \) by summing the number of points across all frames: $V_i = \sum_{t \in T_k} \left| \tilde{m}_{i,t} \right|$,
where \( \left| \tilde{m}_{i,t} \right| \) denotes the number of points in mask \( \tilde{m}_{i,t} \). 
We sort the masklets in descending order based on their volumes \( V_i \), and incrementally suppress masklets with intersection-over-minimum larger than empirically determined threshold. %
With this flattening operation, we favor larger and temporally consistent instances (\ie, prefer larger volumes), and ensure unique point-to-instance assignments (via IoM-based suppression) in the 4D space-time volume. However, we obtain pseudo-labels \textit{only} for objects visible in the first video frame \( I_{t_k} \) of each window \( w_k \).

\subsubsection{Labeling Arbitrary-Length Sequences}
\label{subsubsec:labeling-arbitrary-sequences}
After labeling each temporal window, we obtain pseudo-labels for point clouds within overlapping windows of size $K$, with local instance identifiers \( \text{id}_{i,k} \).
To produce pseudo-labels for the full sequence of length $T$ and account for new objects entering the scene, we associate instances across windows in a near-online fashion (with stride $S$), resulting our final pseudo-labels \( \{ ( \tilde{m}_{i,t}, \text{id}_i, f_i ) \mid t \in T \} \)  (\cref{fig:sal4d-pseudo-labels}). 

For each pair of overlapping windows $(w_{k-1}, w_k)$, we perform association via linear assignment. We derive association costs from temporal instance overlaps (measured by 3D-IoU) in the overlapping frames \( T_{k-1} \cap T_k \):
\begin{equation}
c_{ij} = 1 - \text{IoU}_{\text{3D}}(\tilde{m}_{i,k-1}, \tilde{m}_{j,k}),
\end{equation}
where $\tilde{m}_{i,k-1}$ and $\tilde{m}_{j,k}$ are the aggregated Lidar masks of instances \( i \) and \( j \). 
After association, we update the global instance identifiers \( \text{id}_i \) for matched instances and aggregate their semantic features \( f_i \). As a final post-processing step, we remove instances that are shorter than a threshold $\tau$.

\subsection{SAL-4D Model}
\label{subsec:model}

\begin{figure}[t]
    \centering
    \includegraphics[width=\linewidth]{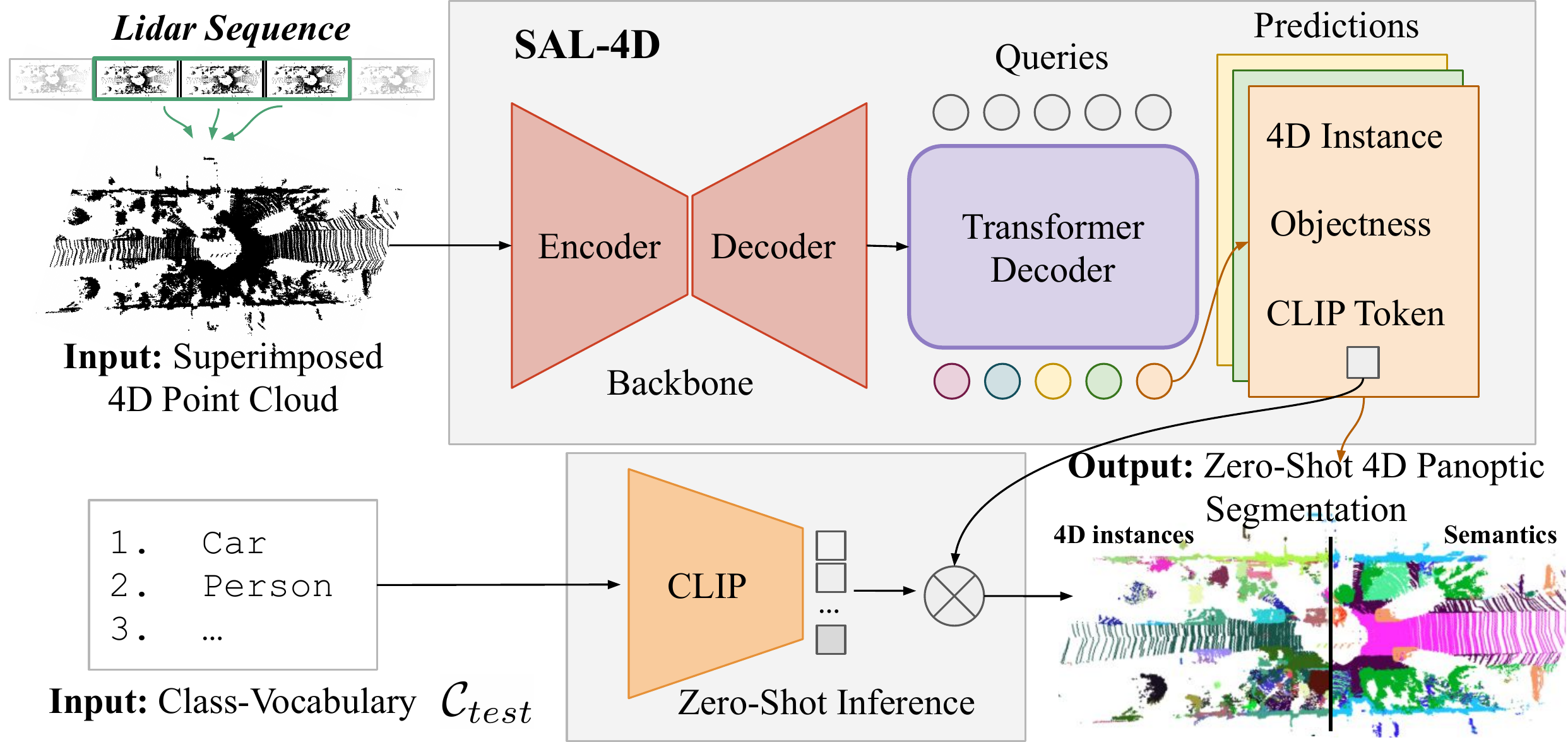}
    \caption{\textbf{\method model} segments individual spatio-temporal instances in 4D Lidar sequences and predicts per-track CLIP tokens that foster test-time zero-shot recognition via text prompts.}
    \label{fig:sal4d-model}
\end{figure}

\PAR{Overview.} 
We follow \textit{tracking-before-detection} design~\cite{Teichman11ICRA, Mitzel12ECCV, Osep18ICRA} and segment and track objects in a class-agnostic fashion. Once localized and tracked, objects can be recognized. 
To operationalize this, we employ a Transformer decoder-based architecture~\cite{carion2020end}.
In a nutshell, our network (\cref{fig:sal4d-model}) consists of a point cloud encoder-decoder network that encodes sequences of point clouds, followed by a Transformer-based object instance decoder that localizes objects in the 4D Lidar space (\cf, \cite{marcuzzi2023ral,yilmaz2024mask4former}). 

\PAR{Model.} 
Our model (\cref{fig:sal4d-model}) operates on point clouds $\mathcal{P}_{super} \in \mathbb{R}^{N \times 4}$, $N = N_{t_k}+\ldots+N_{t_k+K-1}$, superimposed over fixed-size temporal windows $w_k$. 
As in~\cite{sal2024eccv}, we encode superimposed sequences using Minkowski U-Net~\cite{choy20194d} backbone to learn a multi-resolution representation of our input using sparse 3D convolutions. 
For spatio-temporal reasoning, we augment voxel features with Fourier positional embeddings~\cite{fourier2020neurips,yilmaz2024mask4former} that encode 3D spatial and temporal coordinates.

Our segmentation decoder follows the design of~\cite{carion2020end,cheng2022masked,marcuzzi2023ral}. 
Inputs to the decoder are a set of $M$ learnable queries
that interact with voxel features, \ie, our (4D) spatio-temporal representation of the input sequence.  
For each query, we estimate a spatio-temporal mask
, an objectness score indicating how likely a query represents an object and a $d$-dimensional CLIP token capturing object semantics. %
For details, we refer to \conf{ Appendix A.2}\arxiv{\cref{appendix:model}}.

\PAR{Training.} 
Our network predicts a set of spatio-temporal instances, parametrized via segmentation masks over the superimposed point cloud: $\hat{m}_j \in \{0,1\}^{N}$, $j=1, \ldots, M$, obtained by sigmoid activating and thresholding the spatio-temporal mask $\mathcal{M}$. 
To train the network, we establish correspondences between predictions $\hat{m}_j$ and pseudo-labels $\Tilde{m}_i$ via bi-partite matching (following the standard practice~\cite{carion2020end,marcuzzi2023ral, yilmaz2024mask4former}) and evaluate the following loss: 
\begin{equation}
    \label{equ:loss}
    \mathcal{L}_{SAL-4D} = \mathcal{L}_{obj} + \mathcal{L}_{seg} + \mathcal{L}_{token},
\end{equation}
with a cross-entropy loss $\mathcal{L}_{obj}$ indicating whether a mask localizes an object, 
a segmentation loss $\mathcal{L}_{seg}$ (binary cross-entropy and a dice loss following~\cite{marcuzzi2023ral}),
and a CLIP token loss (cosine distance) $\mathcal{L}_{token}$. 
As all three terms are evaluated on a sequence rather than individual frame level, our network implicitly learns to segment and associate instances over time, encouraging temporal semantic coherence. %

\PAR{Inference.} We first decode masks by multiplying objectness scores with the spatio-temporal masks $\mathcal{M} \in \mathbb{R}^{M \times N}$, followed by argmax over each point (details in \conf{Appendix A.2}\arxiv{\cref{appendix:model}}.)
As our model directly processes superimposed point clouds within windows of size $K$, we perform \textit{near-online} inference~\cite{Choi15ICCV} by associating Lidar masklets across time based on 3D-IoU overlap via bi-partite matching (as described in \cref{subsubsec:labeling-arbitrary-sequences}).
For zero-shot prompting, 
we follow \cite{sal2024eccv} and first encode prompts specified in the semantic class vocabulary using a CLIP language encoder. Then, we perform argmax over scores, computed as a dot product between encoded queries and predicted CLIP features.

\section{Experimental Validation}
\label{sec:experiments}

This section first discusses datasets and evaluation protocol and metrics (\cref{subsec:experimental_setup}). 
In \cref{subsec:ablations}, we ablate our pseudo-label engine and model and justify our design decisions.  
In \cref{subsec:benchmarks}, we compare our \method with several zero-shot and supervised baselines on multiple benchmarks for 3D and 4D Lidar Panoptic Segmentation. 

\subsection{Experiments}
\label{subsec:experimental_setup}

\PAR{Datasets.} For evaluation, we utilize two datasets that provide semantic and spatio-temporal instance labels for Lidar, \textit{SemanticKITTI}~\cite{behley2019iccv} and \textit{Panoptic nuScenes}~\cite{fong21ral,Caesar20CVPR}. 
\PARit{SemanticKITTI} was recorded in Karlsruhe, Germany, using a 64-beam Velodyne Lidar sensor at $10Hz$ and provides Lidar and front RGB camera, which we use for pseudo-labeling ($14\%$ of all Lidar points are visible in camera). The dataset provides instance-level spatiotemporal labels for $8$ \thing and $11$ \stuff classes.

\PARit{Panoptic nuScenes} was recorded in Boston and Singapore using 32-beam Velodyne. It provides five cameras with $360^{\circ}$ coverage (covering $48\%$ of all points) at $2Hz$. Spatio-temporal labels are available for $8$ \thing and $8$ \stuff classes.

\PAR{Evaluation metrics.} We follow prior work in \fdlps~\cite{aygun21cvpr} and adopt $LSTQ$ as the core metric for evaluation. In a nutshell, $LSTQ = \sqrt{S_{assoc} \times S_{cls}}$ is defined as the geometric mean of two terms, association term $S_{assoc}$ assesses spatio-temporal segmentation quality, independently of semantics, whereas classification $S_{cls}$ assesses semantic recognition quality and establishes whether points were correctly classified. This separation between spatio-temporal segmentation and semantic recognition makes $LSTQ$ uniquely suitable for studying \taskabbrev. For per-scan evaluation, we adopt Panoptic Quality~\cite{Kirillov19CVPR}, which consists of Segmentation Score (SQ) and Recognition Score (RQ): $PQ = SQ \times RQ$.

\PAR{Frustum and stuff evaluation.} As our pseudo-labels only cover part of the point cloud co-visible in RGB cameras (``frustum''), we focus our ablations to camera view frustums and only report benchmark results on full point clouds. 
Furthermore, since our approach no longer distinguishes \thing and \stuff classes but treats both in a unified manner, we follow~\cite{sal2024eccv} and utilize zero-shot classification labels for merging instances with the same \stuff classes to evaluate on respective dataset class vocabularies.

\begin{table}[t]
\centering
\footnotesize
\begin{tabular}{Hccccccccc}
\toprule
Method & \# frames & Cross & LSTQ & $S_{assoc}$ & $S_{cls}$ & $IoU_{st}$ & $IoU_{th}$  \\
 &  & window &  &  &  &  &  \\
\midrule
Pseudo-label & 8 &  & 49.2 & 70.0 & 34.6 & 36.0 & 36.9  \\
\midrule
Pseudo-label  & 2 & $\checkmark$ & 50.6 & 67.4 & \textbf{37.9} & 37.3 & \textbf{43.5}  \\
Pseudo-label  & 4 & $\checkmark$ & \textbf{51.4} & 69.5 & \textbf{37.9} & \textbf{38.1} & 42.4  \\
Pseudo-label  & 8 & $\checkmark$& 51.1 & \textbf{70.3} & 37.2 & 37.4 & 41.5 \\
Pseudo-label  & 16 & $\checkmark$& 50.5 & 69.6 & 36.7 & 38.0 & 39.5  \\
\bottomrule
\end{tabular}
\caption{\textbf{Pseudo-label ablations on temporal window size and cross-window association:} 
We ablate our approach on temporal window sizes of size $K=\{2,4,8, 16\}$ with stride $\frac{K}{2}$ on \textit{SemanticKITTI} validation set. We average CLIP features for each instance across time.
We observe association score ($S_{assoc}$) improve up to $8$ frames, while zero-shot recognition ($S_{cls}$) saturates at $4$ frames. Without the cross-window association (\cref{subsubsec:labeling-arbitrary-sequences}), the $LSTQ$ drops by 1.9 percentage points.
}
\label{tab:pseudolabels_window_size}
\end{table}

\subsection{Ablations}
\label{subsec:ablations}

We ablate design decisions behind our pseudo-label engine (\cref{subsubsection:pseudo-label-engine}) and model (\cref{subsubsection:model-training}). 
We focus this discussion on temporal window size for tracking, point cloud superposition strategies, and the impact of our cross-window association, and report additional ablations in the appendix. 

\subsubsection{Pseudo-label Engine}
\label{subsubsection:pseudo-label-engine}

\PAR{Labeling temporal windows \vs full sequences.}
Our \method model operates on superimposed point clouds, which only require temporal consistent 4D labels within temporal windows. 
This begs the question, is pseudo-labeling \textit{only} short sequences sufficient? 
We first generate pseudo-labels with consistent IDs only within fixed-size temporal windows (\cref{subsubsec:track-lift-splat}) and train our model by removing points that are not pseudo-labeled. 
However, this method does not fully leverage temporal and semantic information across the whole sequence and account for objects that appear after the first frame of the window. 
As can be seen in~\cref{tab:pseudolabels_window_size}, this version leads to $49.2$ LSTQ (1st entry). 
By additionally associating the fixed-size temporal window (\cref{subsubsec:labeling-arbitrary-sequences}), we observe an improvement of $+1.9$ and obtain $51.1$ LSTQ (4th entry). 
We observe improvements in association and, in particular, for zero-shot recognition ($37.2$ $S_{cls}$ \vs $34.6$, $+2.6$), as averaging CLIP features over longer temporal horizons (enabled by our cross-window association) provides a more consistent semantic signal.

\begin{table}[!t]
\centering
\footnotesize
\resizebox{\linewidth}{!}{
\begin{tabular}{lccccccc}
\toprule
\method & \# frame & Ego. & LSTQ & $S_{assoc}$ & $S_{cls}$ & $IoU_{st}$ & $IoU_{th}$ \\
 &  & Comp &  &  &  &  \\
\midrule
Labels & 8 &  & 51.1 & 70.3 & \textbf{37.2} & 37.4 & \textbf{41.5} \\
\midrule
\multicolumn{8}{c}{Ego-motion compensation} \\
\midrule
Model & 8 & None & 43.7 & 61.3 & 31.2 & 44.3 & 17.1 \\
Model & 8 & Rand & 50.7 & 74.2 & 34.7 & \textbf{48.5} & 19.9 \\
Model & 8 & Mix & \textbf{53.2} & \textbf{77.2} & \textbf{36.6} & 47.9 & \textbf{25.6} \\
\midrule
\multicolumn{8}{c}{Window size} \\
\midrule
Model & 2 & Mix & 52.3 & 74.8 & \textbf{36.6} & 47.7 & 21.3 \\
Model & 4 & Mix & 52.7 & 76.2 & 36.4 & 47.8 & 25.3 \\
Model & 8 & Mix & \textbf{53.2} & \textbf{77.2} & \textbf{36.6} & \textbf{47.9} & \textbf{25.6} \\
\bottomrule
\end{tabular}
}
\caption{
\textbf{\method training.} 
\textit{Top:} To distill our pseudo-labels into a stronger model, it is important to transform point clouds to a common coordinate frame during train- and test-time. Interestingly, our model benefits from randomly \textit{not} performing motion compensation during training by $10\%$. \textit{Bottom:} Processing larger temporal sequences directly benefits our model.
Overall, we distill our pseudo-labels ($51.1$ $LSTQ$) to a stronger model ($53.2$ $LSTQ$).
}
\label{tab:4dsal_model_ablation_Windowsize}
\end{table}

\begin{figure}[t]
\centering
\begin{subfigure}[t]{0.23\textwidth}
    \includegraphics[width=\textwidth]{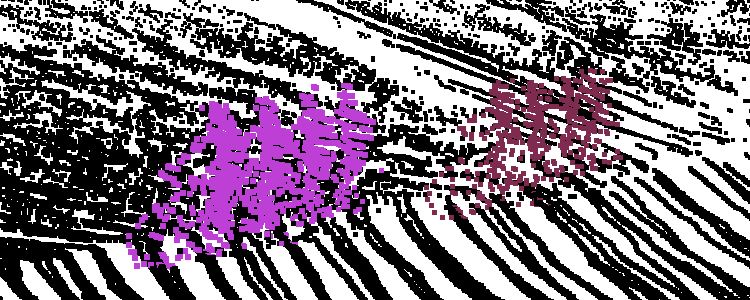}
    \caption{Ground Truth (GT).}
\end{subfigure}
\begin{subfigure}[t]{0.23\textwidth}
    \includegraphics[width=\textwidth]{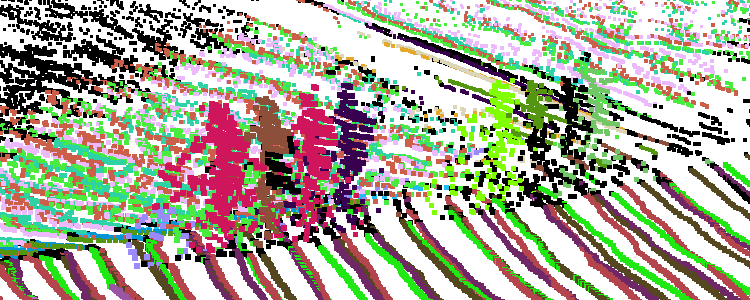}
    \caption{3D Pseudo-Labels.}
\end{subfigure}
\\
\begin{subfigure}[t]{0.23\textwidth}
    \includegraphics[width=\textwidth]{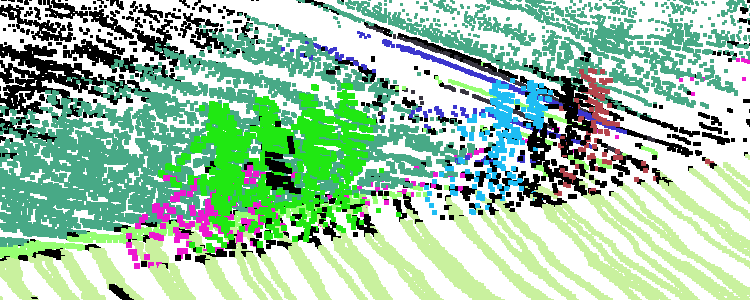}
    \caption{4D Pseudo-Labels ($2$ frames).}
\end{subfigure}
\begin{subfigure}[t]{0.23\textwidth}
    \includegraphics[width=\textwidth]{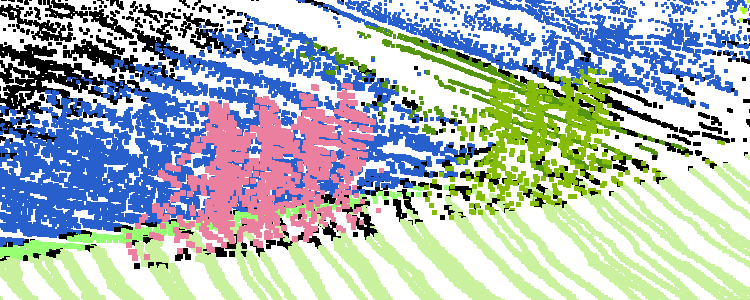}
    \caption{4D Pseudo-Labels ($8$ frames).}
\end{subfigure}
\caption{\textbf{Qualitative results.} We compare our 4D pseudo-labels (obtained over windows of $2\&8$ frames) to GT labels, and single-scan labels. By contrast to GT, our automatically-generated labels cover both \thing and \stuff classes. As can be seen, the temporal coherence of labels improves over larger window sizes.}
\label{fig:sal4d-pseudo-labels}
\end{figure}

\PAR{Temporal window size.}
As discussed in \cref{subsec:pseudo-label-engine}, we first label fixed-size temporal windows, followed by cross-window association. By labeling sequences of arbitrary length, we obtain temporally-stable semantic features and correctly handle outgoing/incoming objects. 
What is the optimal temporal window size? Intuitively, longer temporal windows should be preferable. However, errors that arise during video-instance propagation over larger horizons may degrade the performance.
Our analysis confirms this intuition: we generate pseudo-labels with varying window sizes ($K=\{2, 4, 8, 16\}$) with a fixed stride of $\frac{K}{2}$, and report our findings in \cref{tab:pseudolabels_window_size}.
Our results improve with increasing window size, but performance plateaus after $K=8$. We obtain the overall highest LSTQ with $K=4$ ($51.4$); however, with $K=8$, we observe larger gains in terms of segmentation and tracking ($S_{assoc}$: $70.3$ \vs $69.5$). 
In \cref{fig:sal4d-pseudo-labels}, we confirm this visually by contrasting ground-truth labels with single-scan labels, and our labels, obtained with $K=\{2,8\}$. 
Gains are most significant in terms of $S_{assoc}$, as these results are reported \textit{after} cross-window association. 
The appendix reports a similar analysis conducted before cross-window association. For the remainder, we fix $K=8$.  

\PAR{Comparison with single-scan pseudo-labels.} 
Do our spatio-temporal pseudo-labels improve quality on a single-scan basis?
In~\cref{tab:pseudolabels_single_scan}, we compare our \method pseudo-labels with single-scan labels (SAL~\cite{sal2024eccv}), and report zero-shot and class-agnostic segmentation results. As can be seen, our temporally consistent pseudo-labels perform better than our single-scan counterparts, especially in terms of semantics (a relative $15\%$ improvement \wrt $PQ$ and $20\%$ improvement \wrt $mIoU$).
Our spatio-temporal labels produce fewer instances per scan, which implies spatio-temporal labels improve precision due to temporal coherence. 
We conclude that our approach not only unlocks the training of models for ZS-4D-LPS but also substantially improves pseudo-labels for training ZS-LPS methods~\cite{sal2024eccv}.

\begin{table}[t]
\centering
\footnotesize
\begin{tabular}{lccccc}
\toprule
Method & PQ & SQ & PQ\textsubscript{th} & PQ\textsubscript{st} & mIoU \\
\midrule
\multicolumn{6}{c}{Class-agnostic (Semantic Oracle) LPS} \\
\midrule
SAL~\cite{sal2024eccv} labels & 55.3 & 79.9 & 66.0 & \textbf{47.5} & \textbf{62.1} \\
\method{} labels & \textbf{55.4} & \textbf{80.0} & \textbf{66.4} & 47.4 & 62.0 \\
\midrule
\multicolumn{6}{c}{Zero-Shot LPS} \\
\midrule
SAL~\cite{sal2024eccv} labels & 29.9 & \textbf{74.8} & 35.2 & 26.0 & 31.9 \\
\method{} labels & \textbf{34.5} & 70.5 & \textbf{40.7} & \textbf{29.9} & \textbf{39.1} \\
\bottomrule
\end{tabular}
\caption{\textbf{Single-scan pseudo-label evaluation:} 
We compare our \method{} pseudo-labels to its single-scan counterpart on \textit{SemanticKITTI} validation set. 
Following~\cite{sal2024eccv}, we also report both zero-shot and semantic-oracle \lps (LPS) results.
Our \method{} pseudo-label engine produces a smaller set of higher-quality labels when evaluated on a per-scan basis, with an improvement of over 15\% in recognition score (PQ) and over 20\% in segmentation quality (mIoU). 
}
\label{tab:pseudolabels_single_scan}
\end{table}

\subsubsection{Model and Training}
\label{subsubsection:model-training}

To train the 4D segmentation model, we superimpose point clouds within fixed-size temporal windows and train our model to directly segment superimposed point clouds within these short 4D volumes. For a comparison with our pseudo-labels, we ablate the model ``in-frustum'' and investigate two aspects of point cloud superposition. 
\PAR{Temporal window size:} 
Refers to the number of scans used to construct a superimposed point cloud.  As can be seen in \cref{tab:4dsal_model_ablation_Windowsize}, results are consistent with conclusions for a pseudo-label generation. We obtain the overall best results with a window size of $8$ ($53.2$ LSTQ). Larger temporal window sizes are especially beneficial in terms of segmentation. %
\PAR{Ego-motion:} In 4D space, we can utilize ego-pose to align point clouds to a common coordinate frame. We ablate three options:
(i) no ego-motion compensation (None), 
(ii) select a \emph{random} (Rand) scan as the reference scan, and (iii) a \emph{mixed} (Mix) version of 90\% random reference scan + 10\% no ego-motion compensation (\% determined via line search). 
Results reported in \cref{tab:4dsal_model_ablation_Windowsize} suggest that ego-motion compensation has a positive impact. We obtain $74.2$ $S_{assoc}$ when aligning point clouds, compared to $61.3$ $S_{assoc}$ without. 
Intuitively, this compensation simplifies tracking at inference, but this is not necessarily desirable during the training. 
To ensure that our model learns associations among non-aligned regions, we drop ego-compensation in 10\% of cases, yielding the best overall results ($77.2$ $S_{assoc}$). \textit{With this approach, we distill our pseudo-labels ($51.1$ $LSTQ$) to a \textbf{stronger} model ($53.2$ $LSTQ$) that segments point clouds in the absence of image features.} 

\subsection{Benchmarks}
\label{subsec:benchmarks}

\subsubsection{Lidar Panoptic Segmentation}
\label{subsubsec:lidar-panoptic-segmentation}

In \cref{tab:benchmark-lidar-panoptic-segmentation-semantic-kitti}, we compare our \method to several supervised methods~\cite{hong2021lidar, zhou2021panoptic,sirohi2021efficientlps,razani2021gp,marcuzzi2023ral}, and single-scan zero-shot baseline, SAL~\cite{sal2024eccv}.\footnote{Results we report for the baseline are slightly higher than those reported in~\cite{sal2024eccv}. We refer to the supplementary for further details.}
We compare two variants of our method: our top-performing model, trained on the temporal window of size $8$, and a variant of our model, trained on the temporal window of size $2$, with FrankenFrustum augmentation~\cite{sal2024eccv}, that helps our model, trained on pseudo-labels generated on $14\%$ of full point cloud, to generalize to the full $360^{\circ}$ point clouds. %
As can be seen in~\cref{tab:benchmark-lidar-panoptic-segmentation-semantic-kitti},
\method consistently outperforms SAL baseline: we obtain $38.2$ $PQ$ within-frustum ($+5.1$ \wrt. SAL), and $30.8$ $PQ$ on the full point cloud ($+5.5$ \wrt SAL), and overall reduces the gap to supervised baselines. Improvements are especially notable for \thing classes ($18.3$ \vs $25.5$ $PQ_{th}$). We attribute these gains to temporal coherence imposed during pseudo-labeling and model training.

\begin{table}[t]
\centering
\footnotesize
\resizebox{\linewidth}{!}{
\begin{tabular}{clHHHccccccH}
\toprule
& Method & \# fr. &\# stride. & Franken & frustum & \# inst & PQ & SQ & PQ\textsubscript{th} & PQ\textsubscript{st} & mIoU \\
& & & & Frustum & eval & total / mean & & & & & \\
\midrule
\multirow{4}{*}{\rotatebox[origin=c]{90}{Supervised}} 
& DS-Net~\cite{hong2021lidar} & - & - & $\times$ & $\times$ & - & 57.7 & 77.6 & 61.8 & 54.8  & 63.5 \\
& PolarSeg~\cite{zhou2021panoptic} & - & - & $\times$ & $\times$ & - & 59.1 & 78.3 & 65.7 & 54.3  & 64.5 \\
& GP-S3Net~\cite{razani2021gp} & - & - & $\times$ & $\times$ & - & \textbf{63.3} & \textbf{81.4} & \textbf{70.2} & \textbf{58.3}  & 73.0 \\
& MaskPLS~\cite{marcuzzi2023ral} & - & - & $\times$ & $\times$ & - & 59.8 & 76.3 & - & -  & - \\
\midrule
\multirow{4}{*}{\rotatebox[origin=c]{90}{Zero-shot}} 
& SAL~\cite{sal2024eccv} & - & - & $\times$ & $\checkmark$ & 62k / 15.2 & 33.1 & 71.3 & 21.5 & 41.5 & 34.1 \\
& \method & 8 & 4 & $\times$ & $\checkmark$ & 61k / 15.1 & \textbf{38.2} & \textbf{78.1} & \textbf{30.9} & \textbf{43.5} & \textbf{38.5} \\
& SAL~\cite{sal2024eccv} & - & - & $\checkmark$ & $\times$ & 25k / 49.0 & 25.3 & 63.8 & 18.3 & 30.3 & 29.2 \\
& \method & 2 & 1 & $\checkmark$ & $\times$ & 18k / 44.0 & \textbf{30.8} & \textbf{76.9} & \textbf{25.5} & \textbf{34.6} & \textbf{34.8} \\
\bottomrule
\end{tabular}
}
\caption{
    \textbf{3D-LPS evaluation}.
    Training our \method model on the temporal consistent 4D pseudo-labels yields superior 3D (single-scan) performance compared to 3D baselines.
    We evaluate on the SemanticKITTI validation set.
    \method evaluated not only in the frustum was trained with the FrankenFrustum~\cite{sal2024eccv} augmentation.
}
\label{tab:benchmark-lidar-panoptic-segmentation-semantic-kitti}
\end{table}

\begin{table}[t]
\centering
\footnotesize
\resizebox{\linewidth}{!}{
\begin{tabular}{c|clccccc}
\toprule
& & Method & LSTQ & $S_{assoc}$ & $S_{cls}$ & $IoU_{st}$ & $IoU_{th}$ \\
\midrule
\multirow{12}{*}{\rotatebox[origin=c]{90}{SemanticKITTI}} &
\multirow{8}{*}{\rotatebox[origin=c]{90}{Supervised}} &
4D-PLS~\cite{aygun21cvpr} & 62.7 & 65.1 & 60.5 & 65.4 & 61.3 \\
&& 4D-StOP~\cite{kreuzberg20224d} & 67.0 & 74.4 & 60.3 & 65.3 & 60.9 \\
&& 4D-DS-Net~\cite{hong2024unified} & 68.0 & 71.3 & 64.8 & 64.5 & 65.3 \\
&& Eq-4D-PLS~\cite{zhu20234d} & 65.0 & 67.7 & 62.3 & 66.4 & 64.6 \\
&& Eq-4D-StOP~\cite{zhu20234d} & 70.1 & \textbf{77.6} & 63.4 & 66.4 & 67.1 \\
&& Mask4Former~\cite{yilmaz2024mask4former} & 70.5 & 74.3 & 66.9 & \textbf{67.1} & 66.6 \\
&& Mask4D~\cite{marcuzzi2023mask4d} & \textbf{71.4} & 75.4 & 67.5 & 65.8 & 69.9 \\
&& \method & 69.1 & 70.1 & \textbf{68.0} & 65.7 & \textbf{71.2} \\
\cmidrule{2-8}
&
\multirow{4}{*}{\rotatebox[origin=c]{90}{Zero-shot}} &
SAL + MinVIS & 24.7 & 22.2 & 27.5 & 40.9 & 12.5 \\
&& SAL + MOT & 30.9 & 34.4 & 27.7 & 41.0 & 12.9 \\
&& SAL + SW & 32.7 & 38.5 & 27.7 & 41.0 & 12.9 \\
&& \method & \textbf{42.2} & \textbf{51.1} & \textbf{34.9} & \textbf{45.1} & \textbf{20.8} \\
\midrule
\midrule
\multirow{7}{*}{\rotatebox[origin=c]{90}{Paniptic nuScenes}} &
\multirow{3}{*}{\rotatebox[origin=c]{90}{Sup.}} &
4D-PLS~\cite{aygun21cvpr} & 56.1 & 51.4 & - & - & - \\
&& PanopticTrackNet~\cite{hurtado2020mopt} & 43.4 & 32.3 & - & - & - \\
&& EfficientLPS~\cite{sirohi2021efficientlps}+KF & \textbf{62.0} & \textbf{58.6} & - & - & - \\
\cmidrule{2-8}
&
\multirow{4}{*}{\rotatebox[origin=c]{90}{Zero-shot}} &
SAL + SW & 30.3 & 26.9 & 34.3 & 43.0 & 29.9 \\
&& SAL + MOT & 32.8 & 31.5 & 34.3 & 43.0 & 29.9 \\
&& SAL + MinVIS & 33.2 & 32.4 & 34.1 & 42.8 & 29.7 \\
&& \method & \textbf{45.0} & \textbf{48.8} & \textbf{41.5} & \textbf{45.9} & \textbf{37.0} \\
\bottomrule
\end{tabular}
}
\caption{\textbf{\task benchmark:} We compare \method to several supervised baselines for \fdpls and zero-shot baselines. While there is still a gap between supervised methods and zero-shot approaches, \method significantly narrows down this gap. On \textit{SemanticKITTI}, our model \method reaches $59\%$ of the top-performing supervised model, and on nuScenes, $72\%$, even though it is not trained using any labeled data.}
\label{tab:baselines_4d_pls}
\end{table}

\subsubsection{4D Lidar Panoptic Segmentation}
\label{subsubsec:comparison-zero-shot}
We compare \method to several zero-shot baselines and state-of-the-art 4D-LPS methods trained with ground-truth labels provided on \textit{SemanticKITTI} and \textit{Panoptic nuScenes} datasets.
In contrast, all zero-shot approaches rely only on single-scan 3D~\cite{sal2024eccv} or our 4D pseudo-labels. 
To compare \method to baselines that operate on full ($360^{\circ}$) point clouds, we train our model on temporal windows of size $2$, with FrankenFrustum augmentation~\cite{sal2024eccv}, which helps our model to generalize beyond view frustum.

\PAR{\taskabbrev baselines.}
We construct several baselines that associate single-scan 3D SAL~\cite{sal2024eccv} predictions in time (see \conf{Appendix B}\arxiv{\cref{sec:baselines-details}} for further details) and require no temporal GT supervision.
As SemanticKITTI~\cite{behley2019iccv} is dominated by static objects, we propose a minimal viable \emph{Stationary World} (SW) baseline that propagates single-scan masks solely via ego-motion.
Furthermore, we adopt a strong Lidar \emph{Multi-Object Tracking} (MOT) approach~\cite{Weng20iros}, which utilizes Kalman filters in conjunction with a linear assignment association.
As a data-driven and model-centric baseline, the \emph{Video instance segmentation} (VIS) baseline follows~\cite{huang2022minvis} and directly associates objects by matching decoder object queries of the 3D SAL~\cite{sal2024eccv} model in the embedding space.

\begin{figure}[ht]
    \centering
    \makebox[0.15\textwidth]{\footnotesize GT}
    \makebox[0.15\textwidth]{\footnotesize Pseudo-labels}
    \makebox[0.15\textwidth]{\footnotesize \method}
    \includegraphics[width=0.15\textwidth,trim=0 100 0 100,clip]{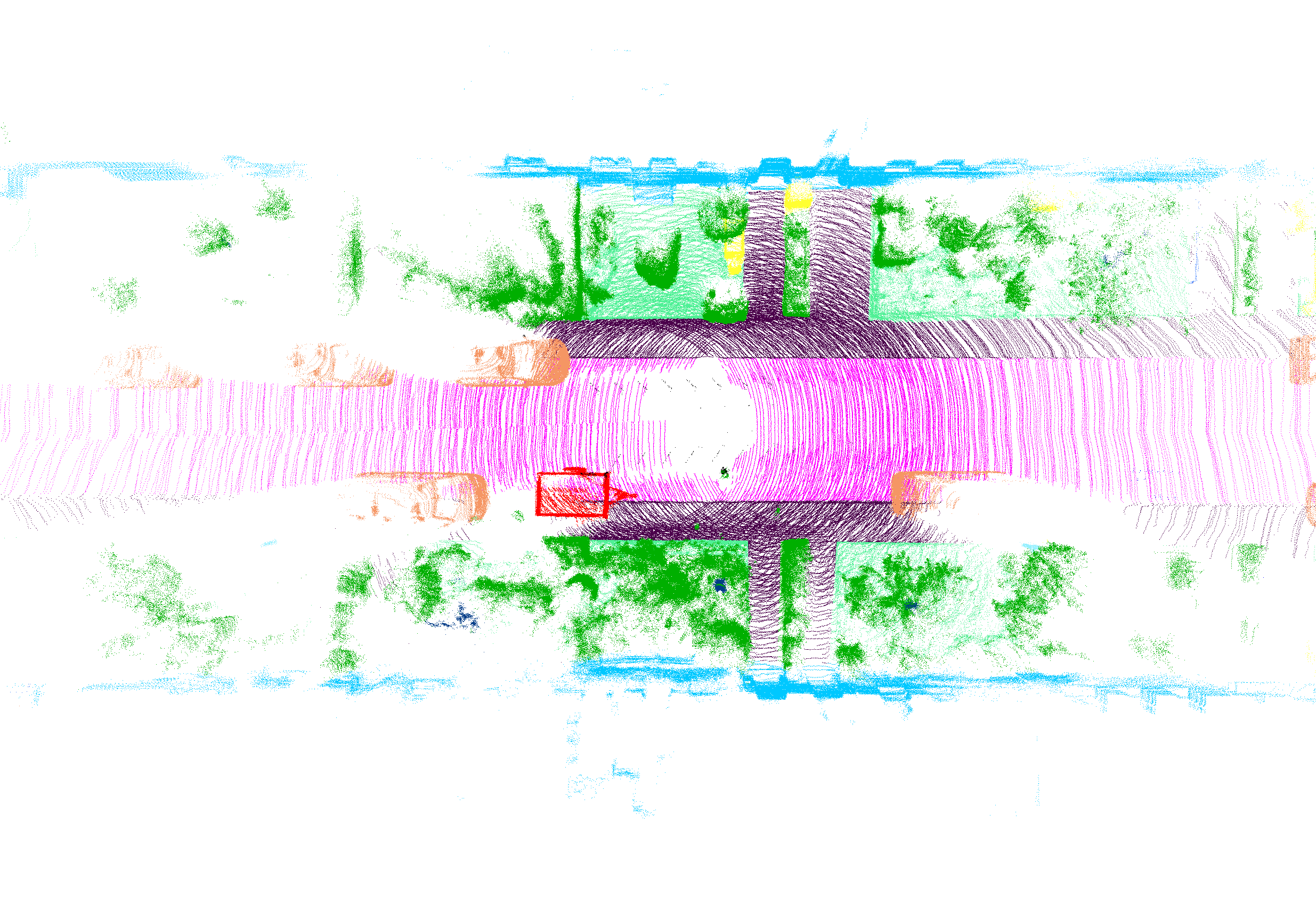}
    \includegraphics[width=0.15\textwidth,trim=0 100 0 100,clip]{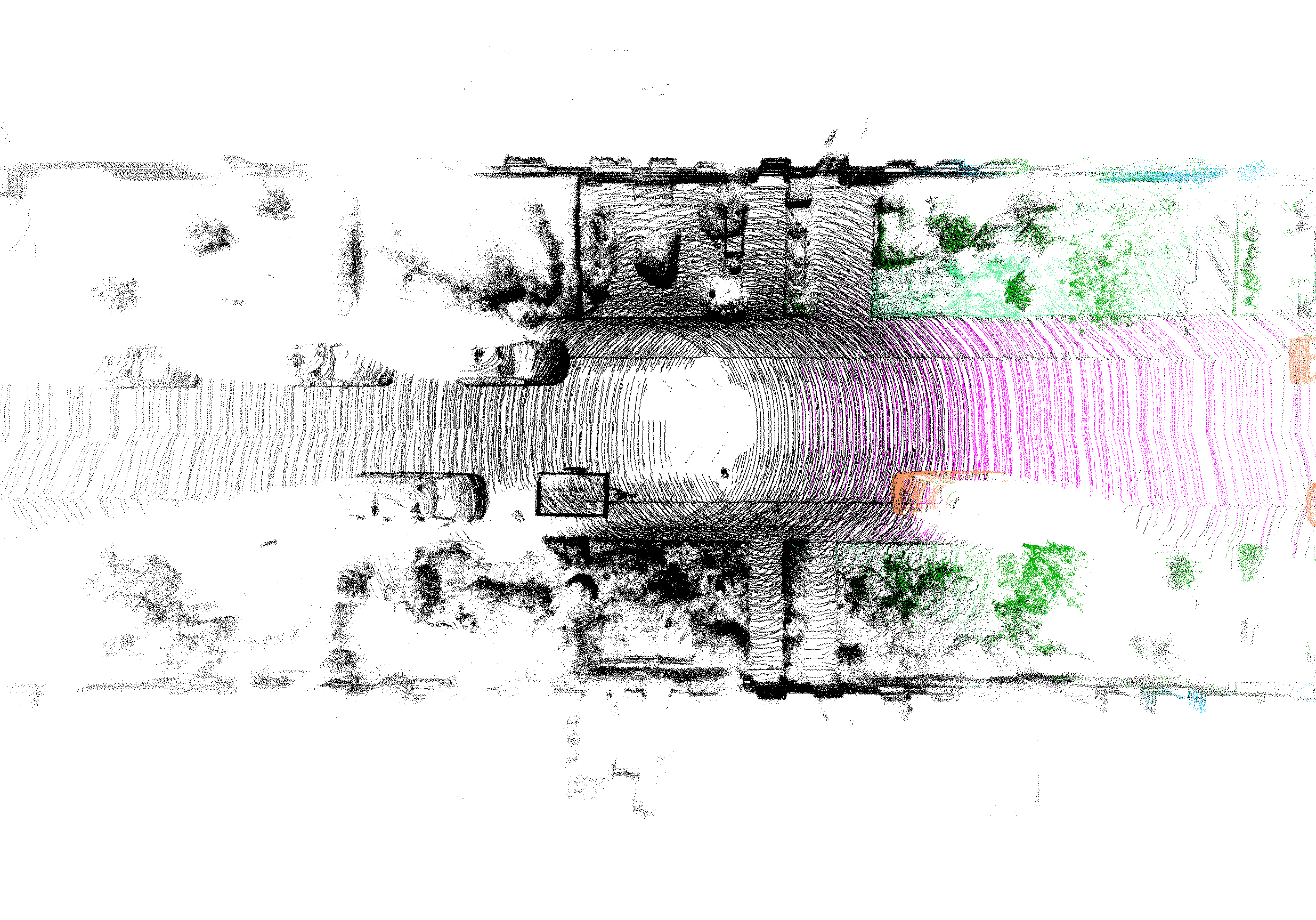}
    \includegraphics[width=0.15\textwidth,trim=0 100 0 100,clip]{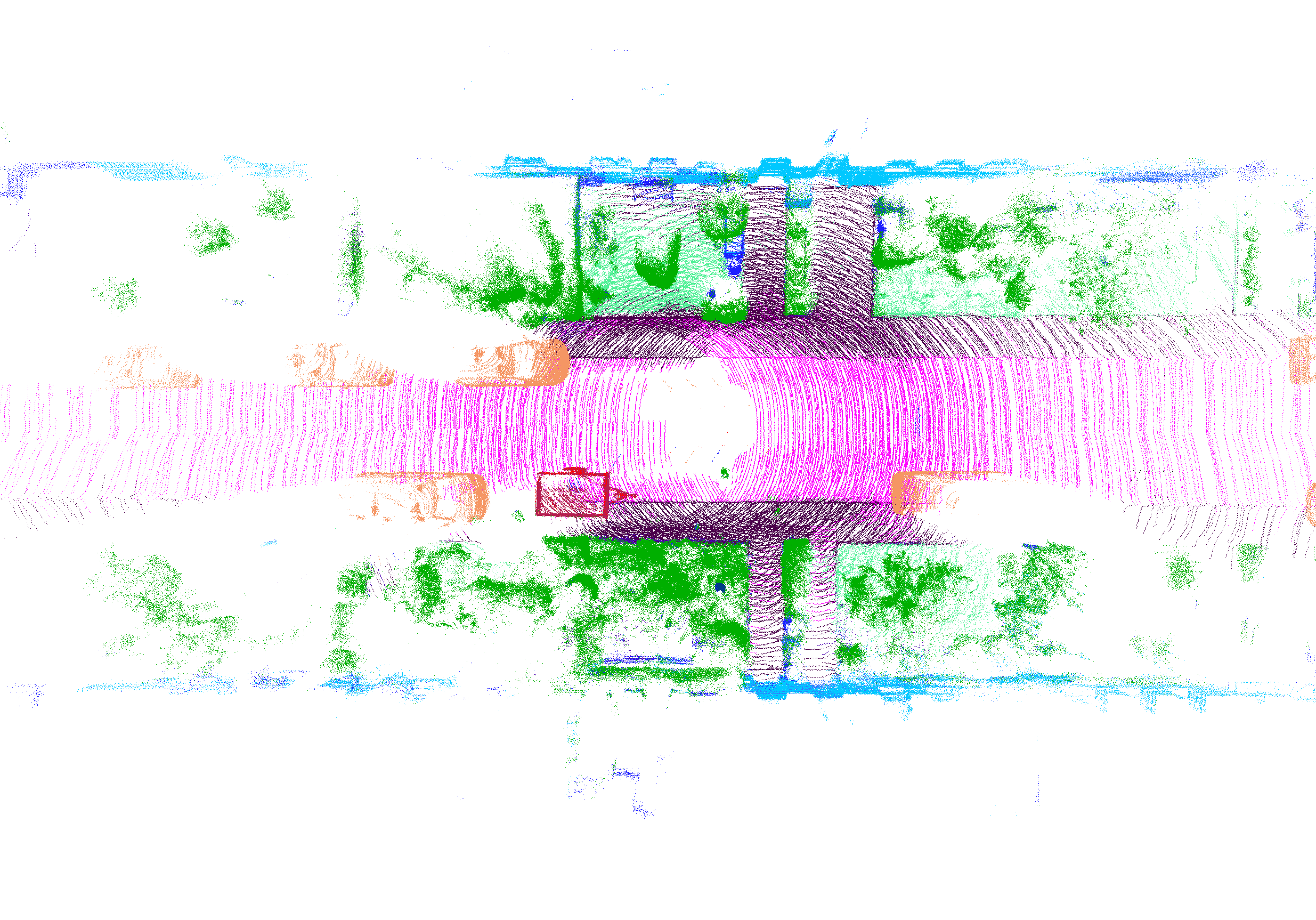}
    \\ \vspace{-10pt}
    \includegraphics[width=0.15\textwidth,trim=0 100 0 100,clip]{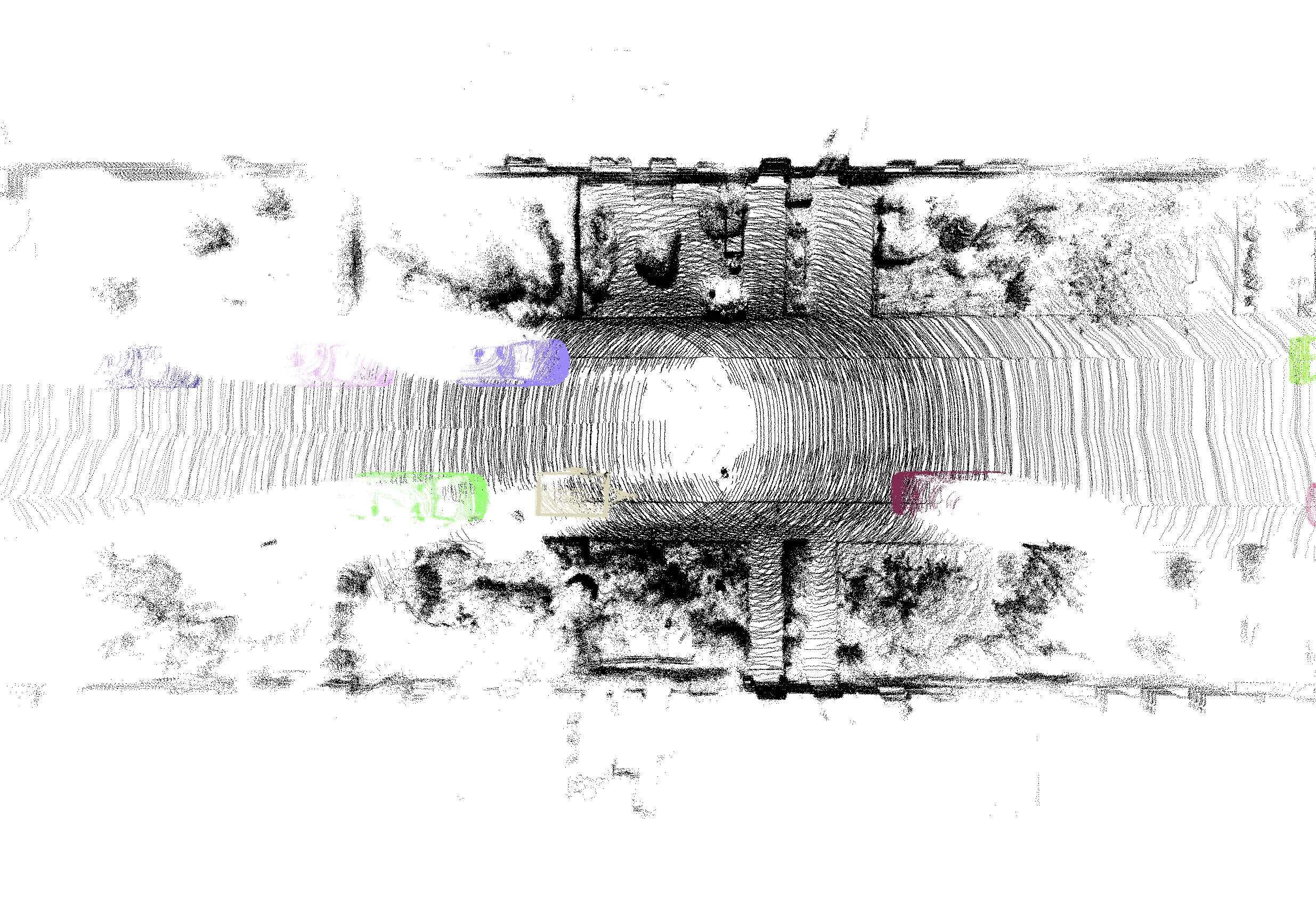}
    \includegraphics[width=0.15\textwidth,trim=0 100 0 100,clip]{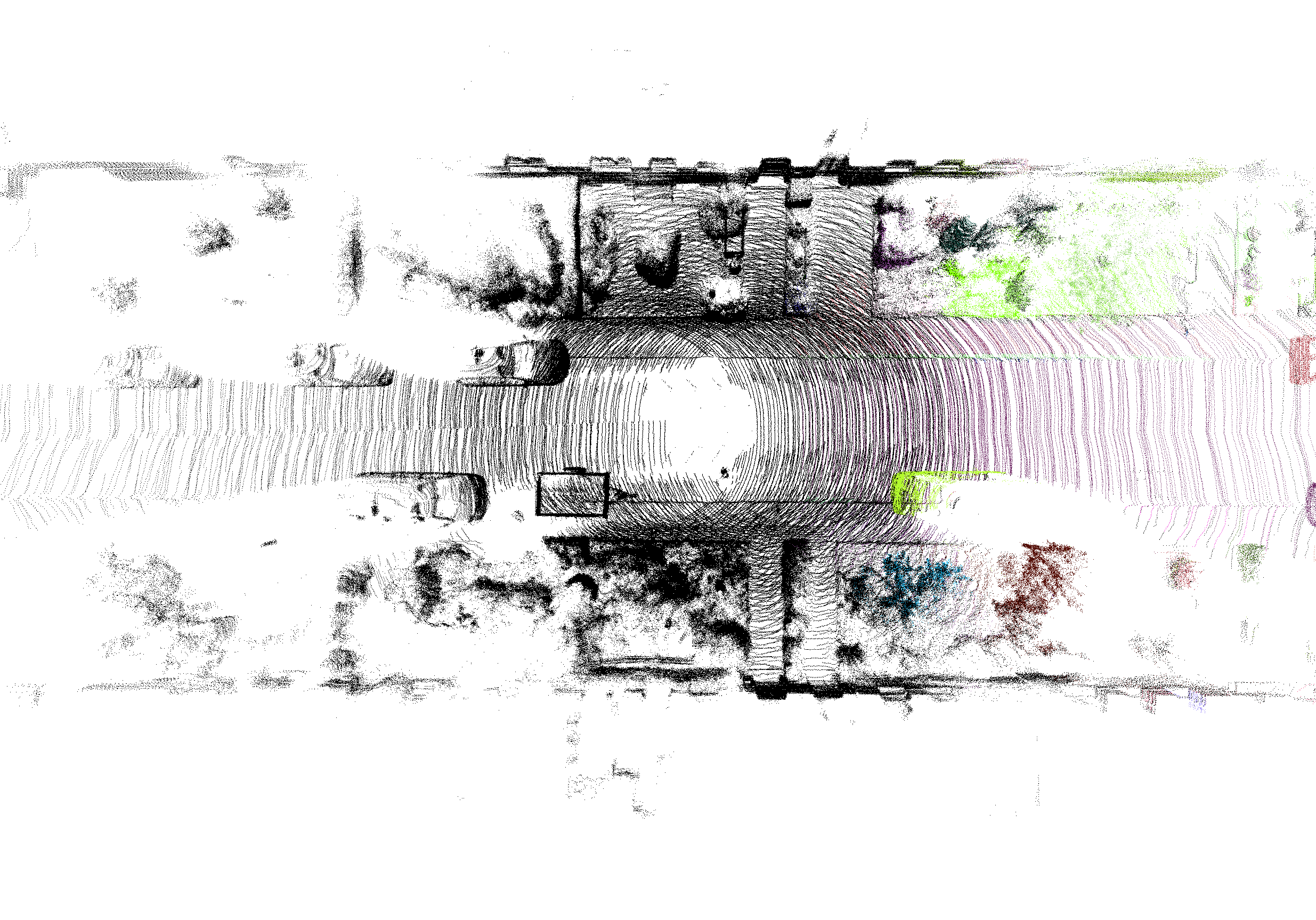}
    \includegraphics[width=0.15\textwidth,trim=0 100 0 100,clip]{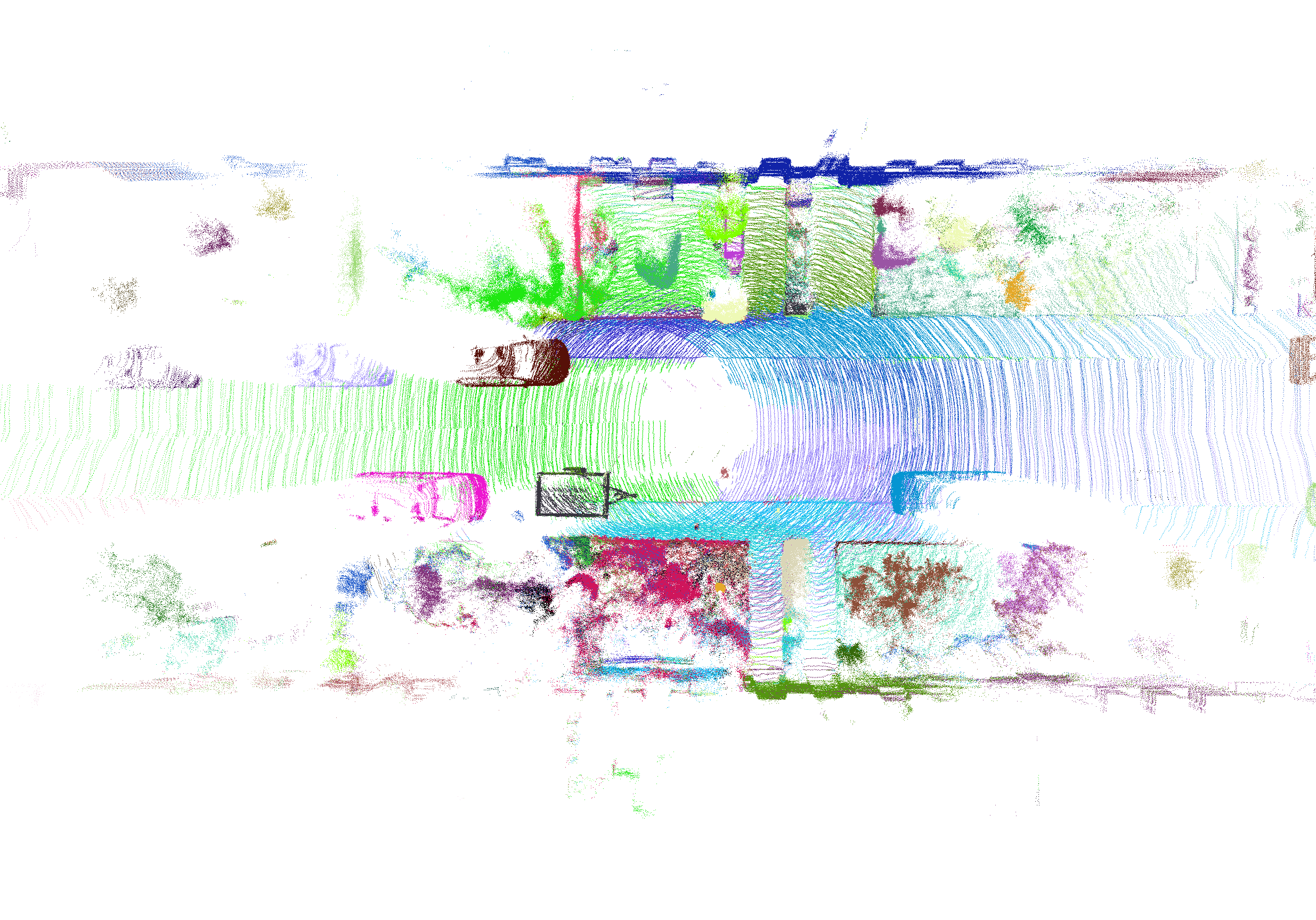}
    \\ \vspace{-10pt}
    \caption{\textbf{Qualitative results on SemanticKITTI.} 
    We show ground-truth (GT) labels (\textit{first column}), our pseudo-labels (\textit{middle column}), and \method results (\textit{right column}). We show semantic predictions (\textit{first row}) and instances (\textit{second row}). As can be seen, our pseudo-labels cover only the camera-visible portion of the sequence (\textit{middle}). By contrast to GT labels, our pseudo-label instances are not limited to a subset of \thing classes (GT, \textit{left column}). Our trained \method thus learns to \textit{densely} segment all classes in space and time (\textit{right column}). 
    \textbf{Importantly}, pseudo-labels do not provide semantic labels, only CLIP tokens. For visualization, we prompt individual instances with prompts that conform to the SemanticKITTI class vocabulary. \textit{Best seen zoomed.}
    }
    \label{fig:main_viz_kitti}
\end{figure}

\PAR{SemanticKITTI.} 
As can be seen in \cref{tab:baselines_4d_pls} (\textit{top}), supervised models are top-performers on this challenging benchmark, specifically, Mask4Former~\cite{yilmaz2024mask4former} ($70.5$ LSTQ) and Mask4D~\cite{marcuzzi2023mask4d} ($71.4$ LSTQ). 
Our \method ($42.2$ LSTQ) outperforms all zero-shot baselines and obtains $59.9\%$ of Mask4Former, similarly trained on temporal windows of size $2$. 
Interestingly, $2^{nd}$ among zero-shot methods is the SW baseline ($32.7$ LSTQ). We assume this baseline outperforms the MOT baseline as \textit{SemanticKITTI} is dominantly static. 
Both geometry-based baselines (SW, MOT) outperform the MinVIS baseline, which mainly relies on data-driven features for the association. We note that \method outperforms zero-shot baselines in terms of association ($S_{assoc}$: $51.1$  \method \vs $38.5$ SW), as well as zero-shot recognition ($S_{cls}$: $34.9$ \method \vs $27.7$ SW and MOT). We provide qualitative results in \cref{fig:main_viz_kitti} and the appendix. 

\PAR{Panoptic nuScenes.}
We report similar findings on \textit{Panoptic nuScenes} dataset in \cref{tab:baselines_4d_pls}. 
Our \method ($45.0$ LSTQ) consistently outperforms baselines and reaches $72.6\%$ of EfficientLPS+KF. 
Due to the different ratio between static and moving objects on nuScenes, MOT baseline ($32.8$ LSTQ) outperforms SW ($30.3$ LSTQ), as expected. MinVIS performs favorably compared to both and achieves $33.2$ LSTQ. This is likely because this data-driven method benefits from a larger \textit{Panoptic nuScenes} dataset. Improvements over baselines are most notable in terms of association ($S_{assoc}$: $48.8$ \method \vs $32.4$ MinVIS).

\section{Conclusions}

We introduced \method for zero-shot segmentation, tracking, and recognition of arbitrary objects in Lidar. Our core component, the pseudo-label engine, distills recent advancements in image-based video object segmentation to Lidar. This enables us to improve significantly over prior single-scan methods and unlock \task. However, as evidenced in \cref{tab:baselines_4d_pls}, a performance gap persists compared to fully-supervised methods. %

\PAR{Challenges.} We observe semantic recognition is the primary source of this gap, with zero-shot recognition $S_{cls}$ ($34.9$) trailing supervised methods ($68.0$). Second, segmentation consistency degrades over extended temporal horizons, reflecting challenges in maintaining coherence across superimposed point clouds. Third, segmentation quality is notably lower for \thing classes compared to \stuff classes, most likely due to the inherent imbalance, mitigated by augmentation strategies in supervised methods. %

\PAR{Future work.} To bridge these gaps, we will focus on (i) refining the data labeling engine to enhance temporal consistency, (ii) expanding the volume of pseudo-labeled data, and (iii) curating high-quality labels for fine-tuning. These steps aim to narrow the divide with supervised methods while preserving \method’s zero-shot scalability.

\clearpage

{\small
\bibliographystyle{ieeenat_fullname}
\bibliography{refs}

\begin{thebibliography}{116}
\providecommand{\natexlab}[1]{#1}
\providecommand{\url}[1]{\texttt{#1}}
\expandafter\ifx\csname urlstyle\endcsname\relax
  \providecommand{\doi}[1]{doi: #1}\else
  \providecommand{\doi}{doi: \begingroup \urlstyle{rm}\Url}\fi

\bibitem[Agarwalla et~al.(2023)Agarwalla, Huang, Ziglar, Ferroni, Leal-Taixe, Hays, Osep, and Ramanan]{Agarwalla23iros}
Abhinav Agarwalla, Xuhua Huang, Jason Ziglar, Francesco Ferroni, Laura Leal-Taixe, James Hays, Aljosa Osep, and Deva Ramanan.
\newblock Lidar panoptic segmentation and tracking without bells and whistles.
\newblock In \emph{Int. Conf. Intel. Rob. Sys.}, 2023.

\bibitem[Aksoy et~al.(2020)Aksoy, Baci, and Cavdar]{aksoy2020salsanet}
Eren~Erdal Aksoy, Saimir Baci, and Selcuk Cavdar.
\newblock Salsanet: Fast road and vehicle segmentation in lidar point clouds for autonomous driving.
\newblock In \emph{Intel. Veh. Symp.}, 2020.

\bibitem[Athar et~al.(2023)Athar, Li, Casas, and Urtasun]{athar20234d}
Ali Athar, Enxu Li, Sergio Casas, and Raquel Urtasun.
\newblock 4d-former: Multimodal 4d panoptic segmentation.
\newblock In \emph{Conf. Rob. Learn.}, 2023.

\bibitem[Ayg{\"u}n et~al.(2021)Ayg{\"u}n, O{\v{s}}ep, Weber, Maximov, Stachniss, Behley, and Leal-Taix{\'e}]{aygun21cvpr}
Mehmet Ayg{\"u}n, Aljo{\v{s}}a O{\v{s}}ep, Mark Weber, Maxim Maximov, Cyrill Stachniss, Jens Behley, and Laura Leal-Taix{\'e}.
\newblock 4d panoptic lidar segmentation.
\newblock In \emph{IEEE Conf. Comput. Vis. Pattern Recog.}, 2021.

\bibitem[Bansal et~al.(2018)Bansal, Sikka, Sharma, Chellappa, and Divakaran]{Bansal18ECCV}
Ankan Bansal, Karan Sikka, Gaurav Sharma, Rama Chellappa, and Ajay Divakaran.
\newblock Zero-shot object detection.
\newblock In \emph{Eur. Conf. Comput. Vis.}, 2018.

\bibitem[Behley and Stachniss(2018)]{behley18rss}
Jens Behley and Cyrill Stachniss.
\newblock {Efficient Surfel-Based SLAM using 3D Laser Range Data in Urban Environments}.
\newblock In \emph{Rob. Sci. Sys.}, 2018.

\bibitem[Behley et~al.(2019)Behley, Garbade, Milioto, Quenzel, Behnke, Stachniss, and Gall]{behley2019iccv}
Jens Behley, Martin Garbade, Andres Milioto, Jan Quenzel, Sven Behnke, Cyrill Stachniss, and Juergen Gall.
\newblock {SemanticKITTI: A Dataset for Semantic Scene Understanding of LiDAR Sequences}.
\newblock In \emph{ICCV}, 2019.

\bibitem[Behley et~al.(2021)Behley, Milioto, and Stachniss]{Behley21icra}
Jens Behley, Andres Milioto, and Cyrill Stachniss.
\newblock {A Benchmark for LiDAR-based Panoptic Segmentation based on KITTI}.
\newblock In \emph{Int. Conf. Rob. Automat.}, 2021.

\bibitem[Bendale and Boult(2015)]{bendale15CVPR}
Abhijit Bendale and Terrance Boult.
\newblock Towards open world recognition.
\newblock In \emph{IEEE Conf. Comput. Vis. Pattern Recog.}, 2015.

\bibitem[Bucher et~al.(2019)Bucher, Vu, Cord, and P{\'e}rez]{bucher2019zero}
Maxime Bucher, Tuan-Hung Vu, Matthieu Cord, and Patrick P{\'e}rez.
\newblock Zero-shot semantic segmentation.
\newblock \emph{Adv. Neural Inform. Process. Syst.}, 2019.

\bibitem[Caesar et~al.(2020)Caesar, Bankiti, Lang, Vora, Liong, Xu, Krishnan, Pan, Baldan, and Beijbom]{Caesar20CVPR}
Holger Caesar, Varun Bankiti, Alex~H. Lang, Sourabh Vora, Venice~Erin Liong, Qiang Xu, Anush Krishnan, Yu Pan, Giancarlo Baldan, and Oscar Beijbom.
\newblock {nuScenes}: A multimodal dataset for autonomous driving.
\newblock In \emph{IEEE Conf. Comput. Vis. Pattern Recog.}, 2020.

\bibitem[Carion et~al.(2020)Carion, Massa, Synnaeve, Usunier, Kirillov, and Zagoruyko]{carion2020end}
Nicolas Carion, Francisco Massa, Gabriel Synnaeve, Nicolas Usunier, Alexander Kirillov, and Sergey Zagoruyko.
\newblock End-to-end object detection with transformers.
\newblock In \emph{Eur. Conf. Comput. Vis.}, 2020.

\bibitem[Chen et~al.(2023)Chen, Xu, Zou, Cao, Yeung, and Fang]{chen2023svqnet}
Xuechao Chen, Shuangjie Xu, Xiaoyi Zou, Tongyi Cao, Dit-Yan Yeung, and Lu Fang.
\newblock Svqnet: Sparse voxel-adjacent query network for 4d spatio-temporal lidar semantic segmentation.
\newblock In \emph{Int. Conf. Comput. Vis.}, 2023.

\bibitem[Cheng et~al.(2022)Cheng, Misra, Schwing, Kirillov, and Girdhar]{cheng2022masked}
Bowen Cheng, Ishan Misra, Alexander~G Schwing, Alexander Kirillov, and Rohit Girdhar.
\newblock Masked-attention mask transformer for universal image segmentation.
\newblock In \emph{IEEE Conf. Comput. Vis. Pattern Recog.}, 2022.

\bibitem[Choi(2015)]{Choi15ICCV}
Wongun Choi.
\newblock Near-online multi-target tracking with aggregated local flow descriptor.
\newblock In \emph{Int. Conf. Comput. Vis.}, 2015.

\bibitem[Choy et~al.(2019)Choy, Gwak, and Savarese]{choy20194d}
Christopher Choy, JunYoung Gwak, and Silvio Savarese.
\newblock {4D} spatio-temporal convnets: Minkowski convolutional neural networks.
\newblock In \emph{IEEE Conf. Comput. Vis. Pattern Recog.}, 2019.

\bibitem[Chu et~al.(2024)Chu, Harley, Tokmakov, Dave, Guibas, and Fragkiadaki]{chu2024zero}
Wen-Hsuan Chu, Adam~W Harley, Pavel Tokmakov, Achal Dave, Leonidas Guibas, and Katerina Fragkiadaki.
\newblock Zero-shot open-vocabulary tracking with large pre-trained models.
\newblock In \emph{Int. Conf. Rob. Automat.}, 2024.

\bibitem[Dave et~al.(2019)Dave, Tokmakov, and Ramanan]{Dave19ICCVW}
Achal Dave, Pavel Tokmakov, and Deva Ramanan.
\newblock Towards segmenting anything that moves.
\newblock In \emph{ICCV Workshops}, 2019.

\bibitem[Dendorfer et~al.(2020)Dendorfer, O\v{s}ep, Milan, Schindler, Cremers, Reid, Roth, and Leal-Taix{\'e}]{dendorfer20ijcv}
Patrick Dendorfer, Aljo\v{s}a O\v{s}ep, Anton Milan, Konrad Schindler, Daniel Cremers, Ian Reid, Stefan Roth, and Laura Leal-Taix{\'e}.
\newblock Motchallenge: A benchmark for single-camera multiple target tracking.
\newblock \emph{Int. J. Comput. Vis.}, 2020.

\bibitem[Ding et~al.(2023{\natexlab{a}})Ding, Rehder, Schneider, Cordts, and Gall]{ding20233dmotformer}
Shuxiao Ding, Eike Rehder, Lukas Schneider, Marius Cordts, and Juergen Gall.
\newblock 3dmotformer: Graph transformer for online 3d multi-object tracking.
\newblock In \emph{Int. Conf. Comput. Vis.}, 2023{\natexlab{a}}.

\bibitem[Ding et~al.(2024)Ding, Qian, Dong, Zhang, Zang, Cao, Guo, Lin, and Wang]{ding2024sam2long}
Shuangrui Ding, Rui Qian, Xiaoyi Dong, Pan Zhang, Yuhang Zang, Yuhang Cao, Yuwei Guo, Dahua Lin, and Jiaqi Wang.
\newblock Sam2long: Enhancing sam 2 for long video segmentation with a training-free memory tree.
\newblock \emph{arXiv preprint arXiv:2410.16268}, 2024.

\bibitem[Ding et~al.(2023{\natexlab{b}})Ding, Wang, and Tu]{ding2023open}
Zheng Ding, Jieke Wang, and Zhuowen Tu.
\newblock Open-vocabulary universal image segmentation with maskclip.
\newblock In \emph{Int. Conf. Mach. Learn.}, 2023{\natexlab{b}}.

\bibitem[Ester et~al.(1996)Ester, Kriegel, Sander, Xu, et~al.]{Ester96KDD}
Martin Ester, Hans-Peter Kriegel, J{\"o}rg Sander, Xiaowei Xu, et~al.
\newblock A density-based algorithm for discovering clusters in large spatial databases with noise.
\newblock In \emph{Rob. Sci. Sys.}, 1996.

\bibitem[Fong et~al.(2021)Fong, Mohan, Hurtado, Zhou, Caesar, Beijbom, and Valada]{fong21ral}
Whye~Kit Fong, Rohit Mohan, Juana~Valeria Hurtado, Lubing Zhou, Holger Caesar, Oscar Beijbom, and Abhinav Valada.
\newblock Panoptic nuscenes: A large-scale benchmark for lidar panoptic segmentation and tracking.
\newblock \emph{RAL}, 2021.

\bibitem[Gasperini et~al.(2021)Gasperini, Mahani, Marcos-Ramiro, Navab, and Tombari]{gasperini2020panoster}
Stefano Gasperini, Mohammad-Ali~Nikouei Mahani, Alvaro Marcos-Ramiro, Nassir Navab, and Federico Tombari.
\newblock Panoster: End-to-end panoptic segmentation of lidar point clouds.
\newblock \emph{IEEE Rob. Automat. Letters}, 2021.

\bibitem[Ghiasi et~al.(2022)Ghiasi, Gu, Cui, and Lin]{ghiasi2022scaling}
Golnaz Ghiasi, Xiuye Gu, Yin Cui, and Tsung-Yi Lin.
\newblock Scaling open-vocabulary image segmentation with image-level labels.
\newblock In \emph{Eur. Conf. Comput. Vis.}, 2022.

\bibitem[Gu et~al.(2022)Gu, Lin, Kuo, and Cui]{gu2021open}
Xiuye Gu, Tsung-Yi Lin, Weicheng Kuo, and Yin Cui.
\newblock Open-vocabulary object detection via vision and language knowledge distillation.
\newblock \emph{Int. Conf. Learn. Represent.}, 2022.

\bibitem[Held et~al.(2016)Held, Guillory, Rebsamen, Thrun, and Savarese]{Held16RSS}
David Held, Devin Guillory, Brice Rebsamen, Sebastian Thrun, and Silvio Savarese.
\newblock A probabilistic framework for real-time 3d segmentation using spatial, temporal, and semantic cues.
\newblock In \emph{Rob. Sci. Sys.}, 2016.

\bibitem[Hong et~al.(2021)Hong, Zhou, Zhu, Li, and Liu]{hong2021lidar}
Fangzhou Hong, Hui Zhou, Xinge Zhu, Hongsheng Li, and Ziwei Liu.
\newblock Lidar-based panoptic segmentation via dynamic shifting network.
\newblock In \emph{IEEE Conf. Comput. Vis. Pattern Recog.}, 2021.

\bibitem[Hong et~al.(2024)Hong, Kong, Zhou, Zhu, Li, and Liu]{hong2024unified}
Fangzhou Hong, Lingdong Kong, Hui Zhou, Xinge Zhu, Hongsheng Li, and Ziwei Liu.
\newblock Unified 3d and 4d panoptic segmentation via dynamic shifting networks.
\newblock \emph{IEEE Trans. Pattern Anal. Mach. Intell.}, 2024.

\bibitem[Huang et~al.(2022)Huang, Yu, and Anandkumar]{huang2022minvis}
De-An Huang, Zhiding Yu, and Anima Anandkumar.
\newblock Minvis: A minimal video instance segmentation framework without video-based training.
\newblock In \emph{Adv. Neural Inform. Process. Syst.}, 2022.

\bibitem[Huang et~al.(2019)Huang, Zhao, and Huang]{huang2019got}
Lianghua Huang, Xin Zhao, and Kaiqi Huang.
\newblock Got-10k: A large high-diversity benchmark for generic object tracking in the wild.
\newblock \emph{IEEE Trans. Pattern Anal. Mach. Intell.}, 2019.

\bibitem[Huang et~al.(2024)Huang, Gu, Xu, and Kong]{huang2024semantics}
Yuming Huang, Yi Gu, Chengzhong Xu, and Hui Kong.
\newblock Why semantics matters: A deep study on semantic particle-filtering localization in a lidar semantic pole-map.
\newblock \emph{IEEE Transactions on Field Robotics}, 2024.

\bibitem[Hurtado et~al.(2020)Hurtado, Mohan, and Valada]{hurtado2020mopt}
Juana~Valeria Hurtado, Rohit Mohan, and Abhinav Valada.
\newblock Mopt: Multi-object panoptic tracking.
\newblock \emph{arXiv preprint arXiv:2004.08189}, 2020.

\bibitem[Kim et~al.(2022)Kim, Bras{\'o}, O{\v{s}}ep, and Leal-Taix{\'e}]{kim2022polarmot}
Aleksandr Kim, Guillem Bras{\'o}, Aljo{\v{s}}a O{\v{s}}ep, and Laura Leal-Taix{\'e}.
\newblock Polarmot: How far can geometric relations take us in 3d multi-object tracking?
\newblock In \emph{Eur. Conf. Comput. Vis.}, 2022.

\bibitem[Kirillov et~al.(2019)Kirillov, He, Girshick, Rother, and Doll{\'a}r]{Kirillov19CVPR}
Alexander Kirillov, Kaiming He, Ross Girshick, Carsten Rother, and Piotr Doll{\'a}r.
\newblock Panoptic segmentation.
\newblock In \emph{IEEE Conf. Comput. Vis. Pattern Recog.}, 2019.

\bibitem[Kirillov et~al.(2023)Kirillov, Mintun, Ravi, Mao, Rolland, Gustafson, Xiao, Whitehead, Berg, Lo, et~al.]{kirillov2023segment}
Alexander Kirillov, Eric Mintun, Nikhila Ravi, Hanzi Mao, Chloe Rolland, Laura Gustafson, Tete Xiao, Spencer Whitehead, Alexander~C Berg, Wan-Yen Lo, et~al.
\newblock Segment anything.
\newblock In \emph{Int. Conf. Comput. Vis.}, 2023.

\bibitem[Kochanov et~al.(2016)Kochanov, O\v{s}ep, St\"uckler, and Leibe]{Kochanov16IROS}
Deyvid Kochanov, Aljo\v{s}a O\v{s}ep, J\"org St\"uckler, and Bastian Leibe.
\newblock Scene flow propagation for semantic mapping and object discovery in dynamic street scenes.
\newblock In \emph{Int. Conf. Intel. Rob. Sys.}, 2016.

\bibitem[Kolmet et~al.(2022)Kolmet, Zhou, O{\v{s}}ep, and Leal-Taix{\'e}]{kolmet2022text2pos}
Manuel Kolmet, Qunjie Zhou, Aljo{\v{s}}a O{\v{s}}ep, and Laura Leal-Taix{\'e}.
\newblock Text2pos: Text-to-point-cloud cross-modal localization.
\newblock In \emph{IEEE Conf. Comput. Vis. Pattern Recog.}, 2022.

\bibitem[Kreuzberg et~al.(2022)Kreuzberg, Zulfikar, Mahadevan, Engelmann, and Leibe]{kreuzberg20224d}
Lars Kreuzberg, Idil~Esen Zulfikar, Sabarinath Mahadevan, Francis Engelmann, and Bastian Leibe.
\newblock 4d-stop: Panoptic segmentation of 4d lidar using spatio-temporal object proposal generation and aggregation.
\newblock In \emph{ECCV AVVision Workshop}, 2022.

\bibitem[Kristan et~al.(2015)Kristan, Matas, Leonardis, Felsberg, Cehovin, Fernandez, Vojir, Hager, Nebehay, Pflugfelder, et~al.]{kristan2015visual}
Matej Kristan, Jiri Matas, Ales Leonardis, Michael Felsberg, Luka Cehovin, Gustavo Fernandez, Tomas Vojir, Gustav Hager, Georg Nebehay, Roman Pflugfelder, et~al.
\newblock The visual object tracking vot2015 challenge results.
\newblock In \emph{Int. Conf. Comput. Vis. Workshops}, 2015.

\bibitem[Kristan et~al.(2016)Kristan, Matas, Leonardis, Voj{\'\i}{\v{r}}, Pflugfelder, Fernandez, Nebehay, Porikli, and {\v{C}}ehovin]{kristan2016novel}
Matej Kristan, Jiri Matas, Ale{\v{s}} Leonardis, Tom{\'a}{\v{s}} Voj{\'\i}{\v{r}}, Roman Pflugfelder, Gustavo Fernandez, Georg Nebehay, Fatih Porikli, and Luka {\v{C}}ehovin.
\newblock A novel performance evaluation methodology for single-target trackers.
\newblock \emph{IEEE Trans. Pattern Anal. Mach. Intell.}, 2016.

\bibitem[Lang et~al.(2019)Lang, Vora, Caesar, Zhou, Yang, and Beijbom]{Lang19CVPR}
Alex~H Lang, Sourabh Vora, Holger Caesar, Lubing Zhou, Jiong Yang, and Oscar Beijbom.
\newblock Pointpillars: Fast encoders for object detection from point clouds.
\newblock In \emph{IEEE Conf. Comput. Vis. Pattern Recog.}, 2019.

\bibitem[Le et~al.(2024)Le, Gou, Datta, Shi, Reid, Cai, and Rezatofighi]{le2024jrdb}
Duy~Tho Le, Chenhui Gou, Stavya Datta, Hengcan Shi, Ian Reid, Jianfei Cai, and Hamid Rezatofighi.
\newblock Jrdb-panotrack: An open-world panoptic segmentation and tracking robotic dataset in crowded human environments.
\newblock In \emph{IEEE Conf. Comput. Vis. Pattern Recog.}, 2024.

\bibitem[Leibe et~al.(2008)Leibe, Schindler, Cornelis, and Gool]{Leibe08TPAMI}
Bastian Leibe, Konrad Schindler, Nico Cornelis, and Luc~Van Gool.
\newblock Coupled object detection and tracking from static cameras and moving vehicles.
\newblock \emph{IEEE Trans. Pattern Anal. Mach. Intell.}, 2008.

\bibitem[Li et~al.(2022{\natexlab{a}})Li, Weinberger, Belongie, Koltun, and Ranftl]{li2022languagedriven}
Boyi Li, Kilian~Q Weinberger, Serge Belongie, Vladlen Koltun, and Rene Ranftl.
\newblock Language-driven semantic segmentation.
\newblock In \emph{Int. Conf. Learn. Represent.}, 2022{\natexlab{a}}.

\bibitem[Li et~al.(2022{\natexlab{b}})Li, Xiao~He, Gao, Cheng, and Zhang]{li2022panoptic}
Jinke Li, Yang~Wen Xiao~He, Yuan Gao, Xiaoqiang Cheng, and Dan Zhang.
\newblock Panoptic-phnet: Towards real-time and high-precision lidar panoptic segmentation via clustering pseudo heatmap.
\newblock In \emph{IEEE Conf. Comput. Vis. Pattern Recog.}, 2022{\natexlab{b}}.

\bibitem[Li et~al.(2021)Li, Chen, Liu, Dai, Stachniss, and Gall]{li2021multi}
Shijie Li, Xieyuanli Chen, Yun Liu, Dengxin Dai, Cyrill Stachniss, and Juergen Gall.
\newblock Multi-scale interaction for real-time lidar data segmentation on an embedded platform.
\newblock \emph{IEEE Rob. Automat. Letters}, 2021.

\bibitem[Li et~al.(2022{\natexlab{c}})Li, Danelljan, Ding, Huang, and Yu]{li2022tracking}
Siyuan Li, Martin Danelljan, Henghui Ding, Thomas~E Huang, and Fisher Yu.
\newblock Tracking every thing in the wild.
\newblock In \emph{Eur. Conf. Comput. Vis.}, 2022{\natexlab{c}}.

\bibitem[Li et~al.(2023)Li, Fischer, Ke, Ding, Danelljan, and Yu]{li2023ovtrack}
Siyuan Li, Tobias Fischer, Lei Ke, Henghui Ding, Martin Danelljan, and Fisher Yu.
\newblock Ovtrack: Open-vocabulary multiple object tracking.
\newblock In \emph{IEEE Conf. Comput. Vis. Pattern Recog.}, 2023.

\bibitem[Liang et~al.(2023)Liang, Wu, Dai, Li, Zhao, Zhang, Zhang, Vajda, and Marculescu]{liang2023open}
Feng Liang, Bichen Wu, Xiaoliang Dai, Kunpeng Li, Yinan Zhao, Hang Zhang, Peizhao Zhang, Peter Vajda, and Diana Marculescu.
\newblock Open-vocabulary semantic segmentation with mask-adapted clip.
\newblock In \emph{IEEE Conf. Comput. Vis. Pattern Recog.}, 2023.

\bibitem[Liu et~al.(2022)Liu, Zulfikar, Luiten, Dave, Ramanan, Leibe, O{\v{s}}ep, and Leal-Taix{\'e}]{liu2022opening}
Yang Liu, Idil~Esen Zulfikar, Jonathon Luiten, Achal Dave, Deva Ramanan, Bastian Leibe, Aljo{\v{s}}a O{\v{s}}ep, and Laura Leal-Taix{\'e}.
\newblock Opening up open world tracking.
\newblock In \emph{IEEE Conf. Comput. Vis. Pattern Recog.}, 2022.

\bibitem[Liu et~al.(2021)Liu, Zhang, Cao, Hu, and Tong]{liu2021iccv}
Ze Liu, Zheng Zhang, Yue Cao, Han Hu, and Xin Tong.
\newblock Group-free 3d object detection via transformers.
\newblock In \emph{Int. Conf. Comput. Vis.}, 2021.

\bibitem[Marcuzzi et~al.(2022)Marcuzzi, Nunes, Wiesmann, Vizzo, Behley, and Stachniss]{marcuzzi2022contrastive}
Rodrigo Marcuzzi, Lucas Nunes, Louis Wiesmann, Ignacio Vizzo, Jens Behley, and Cyrill Stachniss.
\newblock Contrastive instance association for 4d panoptic segmentation using sequences of 3d lidar scans.
\newblock \emph{IEEE Rob. Automat. Letters}, 2022.

\bibitem[Marcuzzi et~al.(2023{\natexlab{a}})Marcuzzi, Nunes, Wiesmann, Behley, and Stachniss]{marcuzzi2023ral}
Rodrigo Marcuzzi, Lucas Nunes, Louis Wiesmann, Jens Behley, and Cyrill Stachniss.
\newblock Mask-based panoptic lidar segmentation for autonomous driving.
\newblock \emph{IEEE Rob. Automat. Letters}, 2023{\natexlab{a}}.

\bibitem[Marcuzzi et~al.(2023{\natexlab{b}})Marcuzzi, Nunes, Wiesmann, Marks, Behley, and Stachniss]{marcuzzi2023mask4d}
Rodrigo Marcuzzi, Lucas Nunes, Louis Wiesmann, Elias Marks, Jens Behley, and Cyrill Stachniss.
\newblock Mask4d: End-to-end mask-based 4d panoptic segmentation for lidar sequences.
\newblock \emph{IEEE Rob. Automat. Letters}, 2023{\natexlab{b}}.

\bibitem[Milioto et~al.(2019)Milioto, Vizzo, Behley, and Stachniss]{Milioto19IROS}
Andres Milioto, Ignacio Vizzo, Jens Behley, and Cyrill Stachniss.
\newblock {RangeNet++: Fast and Accurate LiDAR Semantic Segmentation}.
\newblock In \emph{Int. Conf. Intel. Rob. Sys.}, 2019.

\bibitem[Miller et~al.(2018)Miller, Nicholson, Dayoub, and S{\"u}nderhauf]{miller18ICRA}
Dimity Miller, Lachlan Nicholson, Feras Dayoub, and Niko S{\"u}nderhauf.
\newblock Dropout sampling for robust object detection in open-set conditions.
\newblock In \emph{Int. Conf. Rob. Automat.}, 2018.

\bibitem[Mitzel and Leibe(2012)]{Mitzel12ECCV}
Dennis Mitzel and Bastian Leibe.
\newblock Taking mobile multi-object tracking to the next level: People, unknown objects, and carried items.
\newblock In \emph{Eur. Conf. Comput. Vis.}, 2012.

\bibitem[Moosmann and Stiller(2013)]{Moosmann13ICRA}
Frank Moosmann and Christoph Stiller.
\newblock Joint self-localization and tracking of generic objects in 3d range data.
\newblock In \emph{Int. Conf. Rob. Automat.}, 2013.

\bibitem[Najibi et~al.(2023)Najibi, Ji, Zhou, Qi, Yan, Ettinger, and Anguelov]{najibi2023unsupervised}
Mahyar Najibi, Jingwei Ji, Yin Zhou, Charles~R Qi, Xinchen Yan, Scott Ettinger, and Dragomir Anguelov.
\newblock Unsupervised 3d perception with 2d vision-language distillation for autonomous driving.
\newblock In \emph{Int. Conf. Comput. Vis.}, 2023.

\bibitem[Osep et~al.(2024)Osep, Meinhardt, Ferroni, Peri, Ramanan, and Leal-Taixe]{sal2024eccv}
Aljosa Osep, Tim Meinhardt, Francesco Ferroni, Neehar Peri, Deva Ramanan, and Laura Leal-Taixe.
\newblock Better call sal: Towards learning to segment anything in lidar.
\newblock In \emph{Eur. Conf. Comput. Vis.}, 2024.

\bibitem[Ost et~al.(2021)Ost, Mannan, Thuerey, Knodt, and Heide]{Ost_2021_CVPR}
Julian Ost, Fahim Mannan, Nils Thuerey, Julian Knodt, and Felix Heide.
\newblock Neural scene graphs for dynamic scenes.
\newblock In \emph{IEEE Conf. Comput. Vis. Pattern Recog.}, 2021.

\bibitem[O\v{s}ep et~al.(2016)O\v{s}ep, Hermans, Engelmann, Klostermann, Mathias, and Leibe]{Osep16ICRA}
Aljo\v{s}a O\v{s}ep, Alexander Hermans, Francis Engelmann, Dirk Klostermann, Markus Mathias, and Bastian Leibe.
\newblock Multi-scale object candidates for generic object tracking in street scenes.
\newblock In \emph{Int. Conf. Rob. Automat.}, 2016.

\bibitem[O\v{s}ep et~al.(2018)O\v{s}ep, Mehner, Voigtlaender, and Leibe]{Osep18ICRA}
Aljo\v{s}a O\v{s}ep, Wolfgang Mehner, Paul Voigtlaender, and Bastian Leibe.
\newblock Track, then decide: Category-agnostic vision-based multi-object tracking.
\newblock In \emph{Int. Conf. Rob. Automat.}, 2018.

\bibitem[O\v{s}ep et~al.(2020)O\v{s}ep, Voigtlaender, Weber, Luiten, and Leibe]{Osep20ICRA}
Aljo\v{s}a O\v{s}ep, Paul Voigtlaender, Mark Weber, Jonathon Luiten, and Bastian Leibe.
\newblock 4d generic video object proposals.
\newblock In \emph{Int. Conf. Rob. Automat.}, 2020.

\bibitem[Peng et~al.(2023)Peng, Genova, Jiang, Tagliasacchi, Pollefeys, and Funkhouser]{Peng2023OpenScene}
Songyou Peng, Kyle Genova, Chiyu Jiang, Andrea Tagliasacchi, Marc Pollefeys, and Thomas Funkhouser.
\newblock Openscene: 3d scene understanding with open vocabularies.
\newblock In \emph{IEEE Conf. Comput. Vis. Pattern Recog.}, 2023.

\bibitem[Petrovskaya and Thrun(2009)]{Petrovskaya09AR}
Anna Petrovskaya and Sebastian Thrun.
\newblock Model based vehicle detection and tracking for autonomous urban driving.
\newblock \emph{Aut. Rob.}, 2009.

\bibitem[Pont-Tuset et~al.(2016)Pont-Tuset, Perazzi, Caelles, Arbeláez, {Sorkine}-{Hornung}, and Gool]{Perazzi16CVPR}
Jordi Pont-Tuset, Federico Perazzi, Sergi Caelles, Pablo Arbeláez, Alex {Sorkine}-{Hornung}, and Luc~{Van} Gool.
\newblock A benchmark dataset and evaluation methodology for video object segmentation.
\newblock In \emph{IEEE Conf. Comput. Vis. Pattern Recog.}, 2016.

\bibitem[Pont-Tuset et~al.(2017)Pont-Tuset, Perazzi, Caelles, Arbeláez, Sorkine-Hornung, and Gool]{PontTuset17arxiv}
Jordi Pont-Tuset, Federico Perazzi, Sergi Caelles, Pablo Arbeláez, Alex Sorkine-Hornung, and Luc~Van Gool.
\newblock The 2017 {DAVIS} challenge on video object segmentation.
\newblock \emph{arXiv preprint arXiv:1704.00675}, 2017.

\bibitem[Puy et~al.(2023)Puy, Gidaris, Boulch, Sim{\'e}oni, Sautier, P{\'e}rez, Bursuc, and Marlet]{puy2023revisiting}
Gilles Puy, Spyros Gidaris, Alexandre Boulch, Oriane Sim{\'e}oni, Corentin Sautier, Patrick P{\'e}rez, Andrei Bursuc, and Renaud Marlet.
\newblock Revisiting the distillation of image representations into point clouds for autonomous driving.
\newblock \emph{arXiv preprint arXiv:2310.17504}, 2023.

\bibitem[Puy et~al.(2024)Puy, Gidaris, Boulch, Sim{\'e}oni, Sautier, P{\'e}rez, Bursuc, and Marlet]{puy2024three}
Gilles Puy, Spyros Gidaris, Alexandre Boulch, Oriane Sim{\'e}oni, Corentin Sautier, Patrick P{\'e}rez, Andrei Bursuc, and Renaud Marlet.
\newblock Three pillars improving vision foundation model distillation for lidar.
\newblock In \emph{IEEE Conf. Comput. Vis. Pattern Recog.}, 2024.

\bibitem[Qi et~al.(2017{\natexlab{a}})Qi, Su, Mo, and Guibas]{qi2017pointnet}
Charles~R. Qi, Hao Su, Kaichun Mo, and Leonidas~J. Guibas.
\newblock Pointnet: Deep learning on point sets for 3d classification and segmentation.
\newblock In \emph{IEEE Conf. Comput. Vis. Pattern Recog.}, 2017{\natexlab{a}}.

\bibitem[Qi et~al.(2017{\natexlab{b}})Qi, Yi, Su, and Guibas]{qi2017pointnet++}
Charles~R Qi, Li Yi, Hao Su, and Leonidas~J Guibas.
\newblock Pointnet++: Deep hierarchical feature learning on point sets in a metric space.
\newblock \emph{Adv. Neural Inform. Process. Syst.}, 2017{\natexlab{b}}.

\bibitem[Radford et~al.(2021)Radford, Kim, Hallacy, Ramesh, Goh, Agarwal, Sastry, Askell, Mishkin, Clark, et~al.]{radford2021learning}
Alec Radford, Jong~Wook Kim, Chris Hallacy, Aditya Ramesh, Gabriel Goh, Sandhini Agarwal, Girish Sastry, Amanda Askell, Pamela Mishkin, Jack Clark, et~al.
\newblock Learning transferable visual models from natural language supervision.
\newblock In \emph{Int. Conf. Mach. Learn.}, 2021.

\bibitem[Rahman et~al.(2018)Rahman, Khan, and Porikli]{Rahman18ACCV}
Shafin Rahman, Salman~Hameed Khan, and Fatih Porikli.
\newblock Zero-shot object detection: Learning to simultaneously recognize and localize novel concepts.
\newblock \emph{Asian Conf. Comput. Vis.}, 2018.

\bibitem[Rao et~al.(2022)Rao, Zhao, Chen, Tang, Zhu, Huang, Zhou, and Lu]{rao2022denseclip}
Yongming Rao, Wenliang Zhao, Guangyi Chen, Yansong Tang, Zheng Zhu, Guan Huang, Jie Zhou, and Jiwen Lu.
\newblock Denseclip: Language-guided dense prediction with context-aware prompting.
\newblock In \emph{IEEE Conf. Comput. Vis. Pattern Recog.}, 2022.

\bibitem[Ravi et~al.(2024)Ravi, Gabeur, Hu, Hu, Ryali, Ma, Khedr, R{\"a}dle, Rolland, Gustafson, et~al.]{ravi2024sam}
Nikhila Ravi, Valentin Gabeur, Yuan-Ting Hu, Ronghang Hu, Chaitanya Ryali, Tengyu Ma, Haitham Khedr, Roman R{\"a}dle, Chloe Rolland, Laura Gustafson, et~al.
\newblock Sam 2: Segment anything in images and videos.
\newblock \emph{arXiv preprint arXiv:2408.00714}, 2024.

\bibitem[Razani et~al.(2021{\natexlab{a}})Razani, Cheng, Li, Taghavi, Ren, and Bingbing]{razani2021gp}
Ryan Razani, Ran Cheng, Enxu Li, Ehsan Taghavi, Yuan Ren, and Liu Bingbing.
\newblock Gp-s3net: Graph-based panoptic sparse semantic segmentation network.
\newblock In \emph{IEEE Conf. Comput. Vis. Pattern Recog.}, 2021{\natexlab{a}}.

\bibitem[Razani et~al.(2021{\natexlab{b}})Razani, Cheng, Taghavi, and Bingbing]{razani2021lite}
Ryan Razani, Ran Cheng, Ehsan Taghavi, and Liu Bingbing.
\newblock Lite-hdseg: Lidar semantic segmentation using lite harmonic dense convolutions.
\newblock In \emph{Int. Conf. Rob. Automat.}, 2021{\natexlab{b}}.

\bibitem[Reid(1979)]{Reid79TAC}
Donald~B Reid.
\newblock An algorithm for tracking multiple targets.
\newblock \emph{Tran. Automat. Contr.}, 1979.

\bibitem[Sautier et~al.(2022)Sautier, Puy, Gidaris, Boulch, Bursuc, and Marlet]{sautier2022image}
Corentin Sautier, Gilles Puy, Spyros Gidaris, Alexandre Boulch, Andrei Bursuc, and Renaud Marlet.
\newblock Image-to-lidar self-supervised distillation for autonomous driving data.
\newblock In \emph{IEEE Conf. Comput. Vis. Pattern Recog.}, 2022.

\bibitem[Scheirer et~al.(2012)Scheirer, de~Rezende~Rocha, Sapkota, and Boult]{scheirer2012toward}
Walter~J Scheirer, Anderson de Rezende~Rocha, Archana Sapkota, and Terrance~E Boult.
\newblock Toward open set recognition.
\newblock \emph{IEEE transactions on pattern analysis and machine intelligence}, 35\penalty0 (7):\penalty0 1757--1772, 2012.

\bibitem[Seidenschwarz et~al.(2023)Seidenschwarz, Bras{\'o}, Serrano, Elezi, and Leal-Taix{\'e}]{seidenschwarz2023simple}
Jenny Seidenschwarz, Guillem Bras{\'o}, Victor~Castro Serrano, Ismail Elezi, and Laura Leal-Taix{\'e}.
\newblock Simple cues lead to a strong multi-object tracker.
\newblock In \emph{IEEE Conf. Comput. Vis. Pattern Recog.}, 2023.

\bibitem[Sirohi et~al.(2021)Sirohi, Mohan, B{\"u}scher, Burgard, and Valada]{sirohi2021efficientlps}
Kshitij Sirohi, Rohit Mohan, Daniel B{\"u}scher, Wolfram Burgard, and Abhinav Valada.
\newblock Efficientlps: Efficient lidar panoptic segmentation.
\newblock \emph{IEEE Transactions on Robotics}, 2021.

\bibitem[Sun et~al.(2020)Sun, Kretzschmar, Dotiwalla, Chouard, Patnaik, Tsui, Guo, Zhou, Chai, Caine, et~al.]{sun20CVPR}
Pei Sun, Henrik Kretzschmar, Xerxes Dotiwalla, Aurelien Chouard, Vijaysai Patnaik, Paul Tsui, James Guo, Yin Zhou, Yuning Chai, Benjamin Caine, et~al.
\newblock Scalability in perception for autonomous driving: Waymo open dataset.
\newblock In \emph{IEEE Conf. Comput. Vis. Pattern Recog.}, 2020.

\bibitem[Tancik et~al.(2020)Tancik, Srinivasan, Mildenhall, Fridovich-Keil, Raghavan, Singhal, Ramamoorthi, Barron, and Ng]{fourier2020neurips}
Matthew Tancik, Pratul Srinivasan, Ben Mildenhall, Sara Fridovich-Keil, Nithin Raghavan, Utkarsh Singhal, Ravi Ramamoorthi, Jonathan Barron, and Ren Ng.
\newblock Fourier features let networks learn high frequency functions in low dimensional domains.
\newblock In \emph{Adv. Neural Inform. Process. Syst.}, 2020.

\bibitem[Tang et~al.(2020)Tang, Liu, Zhao, Lin, Lin, Wang, and Han]{tang2020spvnas}
Haotian Tang, Zhijian Liu, Shengyu Zhao, Yujun Lin, Ji Lin, Hanrui Wang, and Song Han.
\newblock Searching efficient 3d architectures with sparse point-voxel convolution.
\newblock In \emph{Eur. Conf. Comput. Vis.}, 2020.

\bibitem[Teichman et~al.(2011)Teichman, Levinson, and Thrun]{Teichman11ICRA}
Alex Teichman, Jesse Levinson, and Sebastian Thrun.
\newblock Towards {3D} object recognition via classification of arbitrary object tracks.
\newblock In \emph{Int. Conf. Rob. Automat.}, 2011.

\bibitem[Thomas et~al.(2019)Thomas, Qi, Deschaud, Marcotegui, Goulette, and Guibas]{Thomas19ICCV}
Hugues Thomas, Charles~R. Qi, Jean-Emmanuel Deschaud, Beatriz Marcotegui, Fran{\c{c}}ois Goulette, and Leonidas~J. Guibas.
\newblock Kpconv: Flexible and deformable convolution for point clouds.
\newblock In \emph{Int. Conf. Comput. Vis.}, 2019.

\bibitem[Thrun et~al.(2006)Thrun, Montemerlo, Dahlkamp, Stavens, Aron, Diebel, et~al.]{thrun2006stanley}
Sebastian Thrun, Mike Montemerlo, Hendrik Dahlkamp, David Stavens, Andrei Aron, James Diebel, et~al.
\newblock Stanley: The robot that won the darpa grand challenge.
\newblock \emph{Journal of field Robotics}, 2006.

\bibitem[Wang et~al.(2023)Wang, Li, Liu, Liu, and Zhu]{Wang_2023_CVPR}
Shaoyu Wang, Wanji Li, Wenwei Liu, Xin Liu, and Jianguo Zhu.
\newblock Lidar2map: In defense of lidar-based semantic map construction using online camera-to-lidar distillation.
\newblock In \emph{IEEE Conf. Comput. Vis. Pattern Recog.}, 2023.

\bibitem[Weng et~al.(2020)Weng, Wang, Held, and Kitani]{Weng20iros}
Xinshuo Weng, Jianren Wang, David Held, and Kris Kitani.
\newblock {3D Multi-Object Tracking: A Baseline and New Evaluation Metrics}.
\newblock In \emph{Int. Conf. Intel. Rob. Sys.}, 2020.

\bibitem[Wu et~al.(2018)Wu, Wan, Yue, and Keutzer]{Wu18ICRA}
Bichen Wu, Alvin Wan, Xiangyu Yue, and Kurt Keutzer.
\newblock Squeezeseg: Convolutional neural nets with recurrent crf for real-time road-object segmentation from 3d lidar point cloud.
\newblock In \emph{Int. Conf. Rob. Automat.}, 2018.

\bibitem[Wu et~al.(2019)Wu, Zhou, Zhao, Yue, and Keutzer]{Wu19ICRA}
Bichen Wu, Xuanyu Zhou, Sicheng Zhao, Xiangyu Yue, and Kurt Keutzer.
\newblock Squeezesegv2: Improved model structure and unsupervised domain adaptation for road-object segmentation from a lidar point cloud.
\newblock In \emph{Int. Conf. Rob. Automat.}, 2019.

\bibitem[Wu et~al.(2024)Wu, Hou, Huang, Lin, He, Zhu, et~al.]{wu2024taseg}
Xiaopei Wu, Yuenan Hou, Xiaoshui Huang, Binbin Lin, Tong He, Xinge Zhu, et~al.
\newblock Taseg: Temporal aggregation network for lidar semantic segmentation.
\newblock In \emph{IEEE Conf. Comput. Vis. Pattern Recog.}, 2024.

\bibitem[Wu et~al.(2013)Wu, Lim, and Yang]{wu2013online}
Yi Wu, Jongwoo Lim, and Ming-Hsuan Yang.
\newblock Online object tracking: A benchmark.
\newblock In \emph{IEEE Conf. Comput. Vis. Pattern Recog.}, 2013.

\bibitem[Xian et~al.(2018)Xian, Lampert, Schiele, and Akata]{Xian18TPAMI}
Yongqin Xian, Christoph~H. Lampert, Bernt Schiele, and Zeynep Akata.
\newblock Zero-shot learning - a comprehensive evaluation of the good, the bad and the ugly.
\newblock \emph{IEEE Trans. Pattern Anal. Mach. Intell.}, 2018.

\bibitem[Xiao et~al.(2024)Xiao, Jing, Wu, Zhu, Ji, Jiang, et~al.]{xiao20243d}
Zihao Xiao, Longlong Jing, Shangxuan Wu, Alex~Zihao Zhu, Jingwei Ji, Chiyu~Max Jiang, et~al.
\newblock 3d open-vocabulary panoptic segmentation with 2d-3d vision-language distillation.
\newblock In \emph{Eur. Conf. Comput. Vis.}, 2024.

\bibitem[Xiong et~al.(2011)Xiong, Munoz, Bagnell, and Hebert]{xiong11icra}
Xuehan Xiong, Daniel Munoz, J.~Andrew Bagnell, and Martial Hebert.
\newblock {3-D Scene Analysis via Sequenced Predictions over Points and Regions}.
\newblock In \emph{Int. Conf. Rob. Automat.}, 2011.

\bibitem[Xu et~al.(2023{\natexlab{a}})Xu, Liu, Vahdat, Byeon, Wang, and De~Mello]{xu2023open}
Jiarui Xu, Sifei Liu, Arash Vahdat, Wonmin Byeon, Xiaolong Wang, and Shalini De~Mello.
\newblock Open-vocabulary panoptic segmentation with text-to-image diffusion models.
\newblock In \emph{IEEE Conf. Comput. Vis. Pattern Recog.}, 2023{\natexlab{a}}.

\bibitem[Xu et~al.(2023{\natexlab{b}})Xu, Zhang, Wei, Hu, and Bai]{xu2023side}
Mengde Xu, Zheng Zhang, Fangyun Wei, Han Hu, and Xiang Bai.
\newblock Side adapter network for open-vocabulary semantic segmentation.
\newblock In \emph{IEEE Conf. Comput. Vis. Pattern Recog.}, 2023{\natexlab{b}}.

\bibitem[Xu et~al.(2018)Xu, Yang, Fan, Yang, Yue, Liang, Price, Cohen, and Huang]{Xu18ECCV}
Ning Xu, Linjie Yang, Yuchen Fan, Jianchao Yang, Dingcheng Yue, Yuchen Liang, Brian Price, Scott Cohen, and Thomas Huang.
\newblock {YouTube-VOS}: Sequence-to-sequence video object segmentation.
\newblock In \emph{Eur. Conf. Comput. Vis.}, 2018.

\bibitem[Yan et~al.(2018)Yan, Mao, and Li]{yan2018second}
Yan Yan, Yuxing Mao, and Bo Li.
\newblock Second: Sparsely embedded convolutional detection.
\newblock \emph{Sensors}, 2018.

\bibitem[Yilmaz et~al.(2024)Yilmaz, Schult, Nekrasov, and Leibe]{yilmaz2024mask4former}
Kadir Yilmaz, Jonas Schult, Alexey Nekrasov, and Bastian Leibe.
\newblock Mask4former: Mask transformer for 4d panoptic segmentation.
\newblock In \emph{Int. Conf. Rob. Automat.}, 2024.

\bibitem[Yin et~al.(2021)Yin, Zhou, and Kr{\"a}henb{\"u}hl]{yin2021center}
Tianwei Yin, Xingyi Zhou, and Philipp Kr{\"a}henb{\"u}hl.
\newblock Center-based 3d object detection and tracking.
\newblock In \emph{IEEE Conf. Comput. Vis. Pattern Recog.}, 2021.

\bibitem[Yuan et~al.(2024)Yuan, Li, Zhou, Li, Chen, and Loy]{yuan2024open}
Haobo Yuan, Xiangtai Li, Chong Zhou, Yining Li, Kai Chen, and Chen~Change Loy.
\newblock Open-vocabulary sam: Segment and recognize twenty-thousand classes interactively.
\newblock In \emph{Eur. Conf. Comput. Vis.}, 2024.

\bibitem[Zareian et~al.(2021)Zareian, Rosa, Hu, and Chang]{zareian2021open}
Alireza Zareian, Kevin~Dela Rosa, Derek~Hao Hu, and Shih-Fu Chang.
\newblock Open-vocabulary object detection using captions.
\newblock In \emph{IEEE Conf. Comput. Vis. Pattern Recog.}, 2021.

\bibitem[Zhang et~al.(2008)Zhang, Yuan, and Nevatia]{Zhang08CVPR}
Li Zhang, Li Yuan, and Ramakant Nevatia.
\newblock Global data association for multi-object tracking using network flows.
\newblock In \emph{IEEE Conf. Comput. Vis. Pattern Recog.}, 2008.

\bibitem[Zhang et~al.(2024)Zhang, Zhou, and Pei]{zhang2024evaluation}
Tiantian Zhang, Zhangjun Zhou, and Jialun Pei.
\newblock Evaluation study on sam 2 for class-agnostic instance-level segmentation.
\newblock \emph{arXiv preprint arXiv:2409.02567}, 2024.

\bibitem[Zhong et~al.(2022)Zhong, Yang, Zhang, Li, Codella, Li, Zhou, Dai, Yuan, Li, et~al.]{zhong2022regionclip}
Yiwu Zhong, Jianwei Yang, Pengchuan Zhang, Chunyuan Li, Noel Codella, Liunian~Harold Li, Luowei Zhou, Xiyang Dai, Lu Yuan, Yin Li, et~al.
\newblock Regionclip: Region-based language-image pretraining.
\newblock In \emph{IEEE Conf. Comput. Vis. Pattern Recog.}, 2022.

\bibitem[Zhou et~al.(2022)Zhou, Loy, and Dai]{zhou2022maskclip}
Chong Zhou, Chen~Change Loy, and Bo Dai.
\newblock Extract free dense labels from clip.
\newblock In \emph{Eur. Conf. Comput. Vis.}, 2022.

\bibitem[Zhou and Tuzel(2018)]{zhou2018voxelnet}
Yin Zhou and Oncel Tuzel.
\newblock Voxelnet: End-to-end learning for point cloud based 3d object detection.
\newblock In \emph{IEEE Conf. Comput. Vis. Pattern Recog.}, 2018.

\bibitem[Zhou et~al.(2021)Zhou, Zhang, and Foroosh]{zhou2021panoptic}
Zixiang Zhou, Yang Zhang, and Hassan Foroosh.
\newblock Panoptic-polarnet: Proposal-free lidar point cloud panoptic segmentation.
\newblock In \emph{IEEE Conf. Comput. Vis. Pattern Recog.}, 2021.

\bibitem[Zhu et~al.(2023)Zhu, Han, Cai, Borse, Ghaffari, and Porikli]{zhu20234d}
Minghan Zhu, Shizhong Han, Hong Cai, Shubhankar Borse, Maani Ghaffari, and Fatih Porikli.
\newblock 4d panoptic segmentation as invariant and equivariant field prediction.
\newblock In \emph{Int. Conf. Comput. Vis.}, 2023.

\bibitem[Zhu et~al.(2021)Zhu, Zhou, Wang, Hong, Ma, Li, Li, and Lin]{zhu2020cylindrical}
Xinge Zhu, Hui Zhou, Tai Wang, Fangzhou Hong, Yuexin Ma, Wei Li, Hongsheng Li, and Dahua Lin.
\newblock Cylindrical and asymmetrical 3d convolution networks for lidar segmentation.
\newblock In \emph{IEEE Conf. Comput. Vis. Pattern Recog.}, 2021.

\end{thebibliography}
}

\clearpage

\clearpage
\appendix
\setcounter{section}{0}

\maketitlesupplementary

\begin{abstract}
In this appendix, we provide:
\begin{itemize}
    \item A more detailed description of our core methodology, \method pseudo-label engine and model in (\cref{sec:implementation-details});
    \item In \cref{sec:baselines-details}, we provide more detailed discussion of our baselines;
    \item Additional evaluation, including pseudo-label and model ablations and per-class results (\cref{sec:additional-experimental-evaluation}), and, finally, 
    \item Additional qualitative results (\cref{sec:additional-qualitative-results}).     
\end{itemize}
\end{abstract}

\section{Implementation Details}
\label{sec:implementation-details}

\subsection{Pseudo-label engine}
\label{appendix:pseudo_label_engine}
This section expands the description (\conf{Sec. 3.2}\arxiv{\cref{subsec:pseudo-label-engine}}) of our pseudo-label engine with a higher level of detail, including pseudo-code detailing core components of our pseudo-engine (\textbf{Track--Lift--Flatten, \cref{alg:track_lift_flatten}}, \textbf{Cross-Window Association, \cref{alg:cross-window-association}}). To ensure this section is self-contained, we start with a high-level overview.

\PAR{Inputs\&Notation.} 
Our pseudo-label engine operates with a multi-modal sensory setup. We assume an input Lidar sequence $\mathcal{P} = \{ P_t \}_{t=1}^T$
along with $C$ unlabeled videos $\mathcal{V} = \{ \mathcal{V}^c \}_{c=1}^C$, where each video $\mathcal{V}^c = \{ I_t^c \}_{t=1}^T$ consists of images $I_t^c \in \mathbb{R}^{H \times W \times 3}$ of spatial dimensions $H \times W$, captured by camera $c$ at time $t$.
For each point cloud $P_t$, we produce pseudo-labels, comprising of tuples $\{ \tilde{m}_{i,t}, \text{id}_i, f_i \}_{i=1}^{M_t}$, where $\tilde{m}_{i,t} \in \{0, 1\}^{N_t}$ represents the binary segmentation mask for instance $i$ at time $t$ in the point cloud $P_t$, and $\text{id}_i \in \mathbb{N}$ is the unique object identifier for spatio-temporal instance $i$.  %
Finally, $f_i \in \mathbb{R}^d$ represents instance semantic features aggregated over time.

\PAR{Hyperparameters.} We list relevant hyperparameters in \cref{tab:hyperparameters}.

\subsubsection{Track--Lift--Flatten}

\PAR{Overview.} 
In a nutshell, for each temporal window, we track objects in video (\textit{track}), lift masks to 4D Lidar sequences (\textit{lift}), and, finally, ``\textit{flatten}'' overlapping masklets in the 4D volume. 

\PAR{Sliding windows.} 
We proceed by sliding a temporal window of size $K$ with a stride $S$ over the sequence of length $T$. We first pseudo-label each temporal window, and then perform cross-window association to obtain pseudo-labels for sequences of arbitrary length. 
Our temporal windows \( w_k = \{ (P_t, I_t) \mid t \in T_k \} \) consist of Lidar point clouds and images over specific time frames. Here, \( T_k = \{ t_k, t_k + 1, \ldots, t_k + K - 1 \} \) is the set of time indices for window \( w_k \).
For simplicity, we drop the camera index \( c \) unless explicitly needed.
We explain our approach assuming a single-camera setup (\( C = 1 \)) and discuss the generalization to a multi-camera setup as necessary.

\PAR{Track.}
For each video, we use a segmentation foundation model (SAM~\cite{kirillov2023segment}) to perform grid-prompting in the first video frame of the window \( I_{t_k} \) to localize objects as masks \( \{ m_{i,t_k} \}_{i=1}^{M_{t_k}} \), \( m_{i,t_k} \in \{0,1\}^{H \times W} \), where $M_{t_k}$ denotes the number of discovered instances in $I_{t_k}$. 
We then propagate  masks through the entire window \( \{ I_t \mid t \in T_k \} \) using SAMv2~\cite{ravi2024sam} to obtain masklets \( \{ m_{i,t} \mid t \in T_k \}_{i=1}^{M_{t_k}} \) for all instances discovered in ${I_{t_k}}$. 
This results in $M_{t_k}$ overlapping masklets in a 3D video volume of dimensions \( H \times W \times K \), representing objects visible in \( I_{t_k} \) across the window \( w_k \).

Given masklets  \(\{ m_{i,t} \mid t \in T_k \}_{i=1}^{M_{t_k}} \) and corresponding images \( \{ I_t \mid t \in T_k \} \), we compute semantic features \( f_{i,t} \) for each mask \( m_{i,t} \) using relative mask attention in the CLIP~\cite{radford2021learning} 
feature space and obtain masklets paired with their CLIP features \( \{ ( m_{i,t}, \text{id}_{i,k}, f_{i,t} ) \mid t \in T_k \} \) for each instance \( i \), where \( \text{id}_{i,k} \) is a local instance identifier within window \( w_k \). Detailed parameters of SAM and SAMv2 can be found in \cref{tab:hyperparameters}.

\PAR{Lift.}
We associate 3D points \( \{ P_t \mid t \in T_k \} \) with image masks \( m_{i,t} \) via Lidar-to-camera transformation and projection. 
We refine our lifted Lidar masklets to address sensor misalignment errors using density-based clustering~\cite{Ester96KDD}. We create an ensemble of DBSCAN clusters by varying the density parameter and replacing all lifted masks with DBSCAN masks with sufficient intersection-over-union (IoU) overlap (0.5)~\cite{sal2024eccv}. 
Due to the presence of moving objects, which makes the DBSCAN cluster prone to error, we perform this procedure separately for individual scans. 
Detailed ablations can be found in \cref{sec:additional-experimental-evaluation}.

We obtain sets \( \{ ( \tilde{m}_{i,t}^c, \text{id}_{i,k}^c, f_{i,t}^c ) \mid t \in T_k \} \) independently for each camera \( c \), and fuse instances with sufficient IoU overlap (0.5) across cameras. We fuse their semantic features \( f_{i,t} \) via mask-area-based weighted average to obtain a set of tuples \( \{ ( \tilde{m}_{i,t}, \text{id}_{i,k}, f_{i,t} ) \mid t \in T_k \} \), that represent spatio-temporal instances localized in window \( w_k \).

\PAR{Flatten.}
The resulting set \( \{ ( \tilde{m}_{i,t}, \text{id}_{i,k}, f_{i,t} ) \mid t \in T_k \} \)  contains overlapping masklets in 4D space-time volume, leading to ambiguities in point assignments. 
To ensure each point is assigned to at most one instance, we perform spatio-temporal flattening as follows.
We compute the spatio-temporal volume \( V_i \) of each masklet \( \tilde{M}_i = \{ \tilde{m}_{i,t} \mid t \in T_k \} \) by summing the number of points across all frames: $V_i = \sum_{t \in T_k} \left| \tilde{m}_{i,t} \right|$,
where \( \left| \tilde{m}_{i,t} \right| \) denotes the number of points in mask \( \tilde{m}_{i,t} \). 
We sort the masklets in descending order based on their volumes \( V_i \), and incrementally suppress masklets with intersection-over-minimum larger than empirically determined threshold.
For each masklet \( \tilde{M}_i \) in the sorted list, we compute the Intersection-over-Minimum (IoM) with all remaining masklets \( \tilde{M}_j \):
\begin{equation}
\text{IoM}_{ij} = \frac{ \sum_{t \in T_k} \left| \tilde{m}_{i,t} \cap \tilde{m}_{j,t} \right| }{ \min( V_i, V_j ) }.
\end{equation}
If \( \text{IoM}_{ij} > \theta \) (a predefined threshold we set it as 0.5), we suppress \( \tilde{M}_j \) by removing it from the list.
The value of \( \theta \) controls the aggressiveness of suppression (set to a high value to prevent overlapping masklets).
With this flattening operation, we favor larger and temporally consistent instances (\ie, prefer larger volumes), and ensure unique point-to-instance assignments (via IoM-based suppression) in the 4D space-time volume. However, we obtain pseudo-labels \textit{only} for objects visible in the first video frame \( I_{t_k} \) of each window \( w_k \). 
Objects appearing after $ t_k $ are not captured in this label set.

\subsubsection{Labeling Arbitrary-Length Sequences}

After labeling each temporal window, we obtain pseudo-labels for point clouds within overlapping windows of size $K$, \( \{ ( \tilde{m}_{i,t}, \text{id}_{i,k}, f_{i,t} ) \mid t \in T_k \} \), with local instance identifiers \( \text{id}_{i,k} \).
As mentioned before, the pseudo-label only covers objects found in the first frame of each window.
To produce pseudo-labels for the full sequence of length $T$ and account for new objects entering the scene, as detailed in \cref{alg:cross-window-association}, we associate instances across windows in a near-online fashion (with stride $S$), resulting in our final pseudo-labels \( \{ ( \tilde{m}_{i,t}, \text{id}_{i}, f_{i} ) \mid t \in T \} \), where \( \text{id}_i \) is consistent across the sequence and $f_{i}$ is averaged CLIP feature of the same instance across the sequence.

For each pair of overlapping windows $(w_{k-1}, w_k)$, we perform association by solving a linear assignment problem:
\begin{equation}
\mathbf{A}^* = \arg \min_{\mathbf{A}} \sum_{i=1}^{M_{k-1}} \sum_{j=1}^{M_k} c_{ij} A_{ij}
\end{equation}
Subject to:
\[
\sum_{j=1}^{M_k} A_{ij} \leq 1, \quad \forall i = 1, \ldots, M_{k-1}
\]
\[
\sum_{i=1}^{M_{k-1}} A_{ij} \leq 1, \quad \forall j = 1, \ldots, M_k
\]
\[
A_{ij} \in \{0,1\}.
\]
Here, $A_{ij}$ indicates whether instance \( \text{id}_{i,k-1} \) in $w_{k-1}$ is assigned to instance \( \text{id}_{j,k} \) in $w_k$.
 We derive association costs from temporal instance overlaps (measured by 3D-IoU) in the overlapping frames \( T_{k-1} \cap T_k \), defined as:
\begin{equation}
c_{ij} = 1 - \text{IoU}_{\text{3D}}(\tilde{m}_{i,k-1}, \tilde{m}_{j,k}),
\end{equation}
where $\tilde{m}_{i,k-1}$ and $\tilde{m}_{j,k}$ are the aggregated Lidar masks of instances \( i \) and \( j \) over the overlapping frames.
This linear assignment problem can be efficiently solved using the Hungarian algorithm.
After association, we update the global instance identifiers \( \text{id}_i \) for matched instances and aggregate their semantic features \( f_i \) over time.
As a final post-processing step, we remove instances that are shorter than a specified temporal threshold $\tau$ (\ie, instances appearing in fewer than $\tau$ frames, $\tau$ is set to 1 in our experiments).

\begin{algorithm}[t]
\caption{Track-Lift-Flatten (Per-Window Processing)}
\begin{algorithmic}[1]
\Require Window index $k$, time indices $T_k$, Lidar point clouds $\{ P_t \mid t \in T_k \}$, images $\{ I_t^c \mid t \in T_k, c=1,\ldots,C \}$
\Ensure Pseudo-labels for window $w_k$: $\{ (\tilde{m}_{i,t}, \text{id}_{i,k}, f_{i,t}) \mid t \in T_k \}$

\State // \textbf{Track}
\For{each camera $c$}
    \State $I_{t_k}^c \leftarrow$ image at time $t_k$ from camera $c$
    \State $\{ m_{i,t_k}^c \} \leftarrow$ SAM($I_{t_k}^c$) \Comment{Generate initial masks}
    \State $\{ m_{i,t}^c \}_{t \in T_k} \leftarrow$ SAMv2($\{ I_t^c \}_{t \in T_k}, \{ m_{i,t_k}^c \}$) \Comment{Propagate masks}
    \State $\{ f_{i,t}^c \}_{t \in T_k} \leftarrow$ MaskCLIP($\{ I_t^c \}_{t \in T_k}, \{ m_{i,t}^c \}_{t \in T_k}$) \Comment{Compute semantic features}
\EndFor

\State // \textbf{Lift}
\For{each time $t \in T_k$}
    \State $P_t \leftarrow$ Lidar point cloud at time $t$
    \For{each instance $i$}
        \For{each camera $c$}
            \State $\tilde{m}_{i,t}^c \leftarrow$ project\_mask($P_t$, $m_{i,t}^c$) \Comment{Project image masks onto Lidar}
        \EndFor
        \State $\tilde{m}_{i,t} \leftarrow$ merge\_masks($\{ \tilde{m}_{i,t}^c \}_{c=1}^C$) \Comment{Merge masks from all cameras}
        \State $\tilde{m}_{i,t} \leftarrow$ refine\_with\_DBSCAN($\tilde{m}_{i,t}$, $P_t$) \Comment{Refine using DBSCAN}
    \EndFor
\EndFor

\State // \textbf{Flatten}
\State Compute volumes $V_i \leftarrow \sum_{t \in T_k} |\tilde{m}_{i,t}|$ for each instance $i$
\State Sort instances $\{ i \}$ in descending order of $V_i$
\For{each instance $i$ in sorted order}
    \For{each instance $j \ne i$ not yet suppressed}
        \State Compute $\text{IoM}_{ij} \leftarrow \frac{\sum_{t \in T_k} |\tilde{m}_{i,t} \cap \tilde{m}_{j,t}|}{\min(V_i, V_j)}$
        \If{$\text{IoM}_{ij} > \theta$}
            \State Suppress instance $j$
        \EndIf
    \EndFor
\EndFor

\State Assign local instance IDs $\text{id}_{i,k}$ within window $w_k$
\State \Return $\{ (\tilde{m}_{i,t}, \text{id}_{i,k}, f_{i,t}) \mid t \in T_k \}$
\end{algorithmic}
\label{alg:track_lift_flatten}
\end{algorithm}

\begin{algorithm}[t]
\caption{Pseudo-label Engine with Cross-Window Association}
\begin{algorithmic}[1]
\Require Lidar sequence $\mathcal{P} = \{ P_t \}_{t=1}^T$, unlabeled videos $\mathcal{V} = \{ \mathcal{V}^c \}_{c=1}^C$, window size $K$, stride $S$
\Ensure Pseudo-labels $\{ (\tilde{m}_{i,t}, \text{id}_i, f_i) \}$ for $t = 1$ to $T$

\State Initialize global instance ID counter: $\text{id} \leftarrow 0$
\State Initialize empty global instance set: $\mathcal{I} \leftarrow \emptyset$
\For{$k = 0$ to $\left\lceil \frac{T}{S} \right\rceil$} \Comment{Slide temporal window}
    \State $t_k \leftarrow kS$
    \State $T_k \leftarrow \{ t_k, t_k + 1, \ldots, \min(t_k + K-1, T) \}$ \Comment{Time indices for window $w_k$}
    \State // \textbf{Per-Window Processing}
    \State $\{ (\tilde{m}_{i,t}, \text{id}_{i,k}, f_{i,t}) \mid t \in T_k \} \leftarrow$ \textbf{Track-Lift-Flatten}($k$, $T_k$, $\{ P_t \}_{t \in T_k}$, $\{ I_t^c \}_{t \in T_k}$)
    \State // \textbf{Cross-Window Association}
    \If{$k > 0$}
        \State $O_k \leftarrow T_k \cap T_{k-1}$ \Comment{Overlapping time frames}
        \State For instances in $w_{k-1}$ and $w_k$, compute costs:
        \For{each instance $i$ in $w_{k-1}$}
            \For{each instance $j$ in $w_k$}
                \State $\tilde{m}_{i,O} \leftarrow$ aggregate\_masks($\{ \tilde{m}_{i,t} \}_{t \in O_k}$)
                \State $\tilde{m}_{j,O} \leftarrow$ aggregate\_masks($\{ \tilde{m}_{j,t} \}_{t \in O_k}$)
                \State $c_{ij} \leftarrow 1 - \text{IoU}_{\text{3D}}(\tilde{m}_{i,O}, \tilde{m}_{j,O})$
            \EndFor
        \EndFor
        \State Solve linear assignment problem with costs $c_{ij}$
        \For{each instance $i$ is matched}
            \State Update global instance IDs $\text{id}_i$ for matched instances
        \EndFor
        \For{each instance $i$ is not matched}
            \State Assign new global instance IDs $\text{id}_i$ for the unmatched new instances
        \EndFor
    \EndIf
    \State Add instances from window $w_k$ to global set $\mathcal{I}$
    \For{each instance $i$ in $\mathcal{I}$}
        \State Aggregate semantic features $f_i$ over time
    \EndFor
\EndFor
\State // \textbf{Post-processing}
\For{each instance $i$ in $\mathcal{I}$}
    \If{number of frames where instance $i$ appears $< \tau$}
        \State Remove instance $i$ from $\mathcal{I}$ \Comment{Discard short-lived instances}
    \EndIf
\EndFor
\State \Return $\{ (\tilde{m}_{i,t}, \text{id}_i, f_i) \}$ for all $i$ and $t$
\end{algorithmic}
\label{alg:cross-window-association}
\end{algorithm}

\subsection{Model}
\label{appendix:model}
\begin{table}[t]
    \centering
    \scriptsize
    \begin{tabular}{l|l}
    \toprule
    Parameter & Value \\
    \midrule
    \multicolumn{2}{c}{SAM~\cite{kirillov2023segment}} \\
    \midrule
    Model & sam\_vit\_h\_4b8939 \\
    Inference POINTS\_PER\_SIDE & $32$ \\
    Inference POINTS\_PER\_BATCH & $64$ \\
    Inference PRED\_IOU\_THRESH & $0.84$ \\
    Inference STABILITY\_SCORE\_THRESH & $0.86$ \\
    Inference STABILITY\_SCORE\_OFFSET & $1.0$ \\
    CROP\_N\_LAYERS & $1$ \\
    Inference BOX\_NMS\_THRESH & $0.7$ \\
    Inference CROP\_NMS\_THRESH & $0.7$ \\
    Inference MIN\_MASK\_REGION\_AREA & $100$ \\    
    \midrule
    \multicolumn{2}{c}{SAM2~\cite{ravi2024sam}} \\
    \midrule
    Model & sam2\_hiera\_large.pt \\
    Config & sam2\_hiera\_l.yaml \\
    \midrule
    \multicolumn{2}{c}{Pseudo-label engine} \\
    \midrule
    NMS IoU threshold & $0.5$ \\
    Multi-view IoU threshold & $0.5$ \\
    DBSCAN IoU overlap threshold & $0.5$\\
    DBSCAN density thresholds & ($1.2488, 0.8136, 0.6952,$ \\
                               & $0.594, 0.4353, 0.3221$) \\
       
    \midrule
    \multicolumn{2}{c}{Zero-shot model} \\
    \midrule
    GPUs & 8 $\times$ 80GB (A100) \\ 
    Batch size & 24 (3 per GPU) \\
    Learning rate (LR) & 0.0002 \\
    Number of iterations & 40000 \\
    LR scheduler & OneCycleLR (pct\_start=0.1) \\
    Number of queries & 300 \\
    Overlap threshold & 0.0 \\
    Loss weights & 2.0, 5.0, 5.0, 10.0, 2.0 \\

    \bottomrule
    \end{tabular}
    \caption{
        \textbf{\method hyperparameters.} We list hyperparameters, including (i) segmentation foundation model parameters (SAM model~\cite{kirillov2023segment}, which we use to generate segmentation masks in images, and SAMv2~\cite{ravi2024sam} that we use for the temporal mask propagation), (ii) the pseudo-label engine and (iii) 4D zero-shot segmentation model parameters.
    }
    \label{tab:hyperparameters}
\end{table}

This section extends \conf{Sec. 3.3}\arxiv{\cref{subsec:model}}, and provides a more detailed description of our model. 
Our model operates on point clouds $\mathcal{P}_{super} \in \mathbb{R}^{N \times 4}$, $N = N_{t_k}+\ldots+N_{t_k+K-1}$, superimposed over fixed-size temporal windows $w_k$.
Within these, our model directly estimates a set of spatio-temporal instances as (binary) segmentation masks, $\mathcal{M} \in \mathbb{R}^{M \times N}$. %
Instead of estimating a posterior over a (fixed) set of semantic classes (as in prior work~\cite{aygun21cvpr, yilmaz2024mask4former,marcuzzi2023ral}), we regress objectness scores $\mathcal{O} \in \mathbb{R}^{M \times 2}$ that indicate how likely an instance represents an actual object. 
Following \cite{sal2024eccv}, we additionally regress for each instance a semantic (CLIP~\cite{radford2021learning}) feature token $\mathcal{F} \in \mathbb{R}^{M \times d}$ that can be used for zero-shot recognition at the test-time.

\PAR{Hyperparameters.} We list relevant hyperparameters in~\cref{tab:hyperparameters}. 

\PAR{Model.} 
Our model operates on point clouds $\mathcal{P}_{super} \in \mathbb{R}^{N \times 4}$, $N = N_{t_k}+\ldots+N_{t_k+K-1}$, superimposed over fixed-size temporal windows $w_k$. %
As in~\cite{sal2024eccv}, we encode superimposed sequences using Minkowski U-Net~\cite{choy20194d} backbone to learn a multi-resolution representation of our input using sparse 3D convolutions. 
resulting in voxel features $F_v \in \mathbb{R}^{C_v \times N_v}$ and point feature $F_p \in \mathbb{R}^{C_p \times N}$. 
For spatio-temporal reasoning, we augment voxel features with Fourier positional embeddings~\cite{fourier2020neurips,yilmaz2024mask4former} that encode 3D spatial and temporal coordinates.

Our segmentation decoder follows the design of~\cite{carion2020end,cheng2022masked,marcuzzi2023ral}. 
Inputs to the decoder are a set of $M$ learnable queries
that interact with voxel features, \ie, our (4D) spatio-temporal representation of the input sequence.  
For each query, we estimate a spatio-temporal mask
, an objectness score indicating how likely a query represents an object and a $d$-dimensional CLIP token capturing object semantics. %

\subsubsection{Backbone} 
As in~\cite{sal2024eccv}, we encode superimposed sequences using Minkowski U-Net~\cite{choy20194d} backbone to learn a multi-resolution representation of our input using sparse 3D convolutions. 
resulting in voxel features $F_v \in \mathbb{R}^{C_v \times N_v}$ and point feature $F_p \in \mathbb{R}^{C_p \times N}$. 
For spatio-temporal reasoning, we augment voxel features with Fourier positional embeddings~\cite{fourier2020neurips,yilmaz2024mask4former} that encode 3D spatial and temporal coordinates.
\subsubsection{Superimposing Point Clouds}
At \textit{test-time}, we transform point clouds to a common coordinate frame using known ego-poses, concatenate points, and voxelize them. Due to the voxelization of point clouds, such concatenation has a minor memory overhead (by contrast to point-based backbones that require more careful superposition strategies~\cite{aygun21cvpr}, which utilizes point-based backbones and performs sub-sampling). 
However, at the \textit{train time}, we leave 10\% of batches un-aligned to expose the network to a larger variety of non-aligned spatio-temporal instances to reduce the imbalance between spatially aligned (static) and non-aligned (dynamic) instances. This imbalance is especially visible in our zero-shot scenario, as opposed to prior works that specialize to \thing classes, among which we observe a larger percentage of moving objects.

\subsubsection{Segmentation Decoder} 
Our segmentation decoder follows the design of~\cite{carion2020end,cheng2022masked,marcuzzi2023ral}. 
Inputs to the decoder are a set of $M$ learnable queries
that interact with voxel features, \ie, our (4D) spatio-temporal representation of the input sequence.  
For each query, we estimate a spatio-temporal mask
$\mathcal{M} \in \mathbb{R}^{M \times N}$, an objectness score $\mathcal{O} \in \mathbb{R}^{M \times 2}$ indicating how likely a query represents an object and a $d$-dimensional CLIP token $\mathcal{F} \in \mathbb{R}^{M \times d}$ capturing object semantics. %

\subsubsection{Training}
Our network predicts a set of spatio-temporal instances, parametrized via segmentation masks over the superimposed point cloud: $\hat{m}_j \in \{0,1\}^{N}$, $j=1, \ldots, M$, obtained by sigmoid activating and thresholding the spatio-temporal mask $\mathcal{M}$. 
To train our network, we first establish correspondences between our set of predictions \( \{ ( \hat{m}_{i,t}, \text{id}_{i,k}, f_{i,t} ) \mid t \in T_k \} \) and pseudo-labels \( \{ ( \tilde{m}_{i,t}, \text{id}_{i,k}, f_{i,t} ) \mid t \in T_k \} \) based on the mask intersection-over-union within temporal window (we perform bipartite matching using Hungarian algorithm, 
as commonly done by Mask transformer-based methods~\cite{marcuzzi2023ral, yilmaz2024mask4former}). 
Once matches are established, we evaluate the following loss: 
\begin{equation}
    \mathcal{L}_{SAL-4D} = \mathcal{L}_{obj} + \mathcal{L}_{mask} + \mathcal{L}_{dice} + \mathcal{L}_{token} + \mathcal{L}_{token\_aux},
\end{equation}
with a cross-entropy loss $\mathcal{L}_{obj}$ indicating whether a mask localizes an object, 
a segmentation loss consists of a binary cross-entropy $\mathcal{L}_{mask}$ and a dice loss $\mathcal{L}_{dice}$ following~\cite{marcuzzi2023ral},
and cosine distanc CLIP token losses $\mathcal{L}_{token}$ and $\mathcal{L}_{token\_aux}$ following~\cite{marcuzzi2023ral}. 
As all three terms are evaluated on a sequence rather than individual frame level, our network implicitly learns to segment and associate instances over time, encouraging temporal semantic coherence. %

As we are training with noisy pseudo-labels that label only a portion of the full Lidar point cloud, we use standard data augmentations (translation, scaling, rotations, \cf, \cite{marcuzzi2023ral}), as well as FrankenFrustum~\cite{sal2024eccv} to train a model that can segment full Lidar point clouds. We also follow the recommendation by \cite{sal2024eccv} and remove all unlabeled points (\ie, those not covered by our pseudo-labels) from our training instances.

\subsubsection{Inference} 
The mask inference is done by first multiplying the objectness score with the spatio-temporal mask $\mathcal{M} \in \mathbb{R}^{M \times N}$ and then performing argmax over each point:
\begin{equation}
    \text{score} = \max(\mathcal{O} \in \mathbb{R}^{M \times 2}, \text{dim=-1}),
\end{equation}
\begin{equation}
    \text{mask} = \argmax(\text{sigmoid}(\mathcal{M} \in \mathbb{R}^{M \times N}) \cdot \text{score}, \text{dim=0}).
\end{equation}
As our model directly processes superimposed point clouds within windows of size $K$, we perform \textit{near-online} inference~\cite{Choi15ICCV} by associating Lidar masklets across time based on 3D-IoU overlap via bi-partite matching (as described in \conf{Sec. 3.2.2}\arxiv{\cref{subsubsec:labeling-arbitrary-sequences}}).
For zero-shot prompting, 
we follow \cite{sal2024eccv} and first encode prompts specified in the semantic class vocabulary using a CLIP language encoder. Then, we perform argmax over scores, computed as a dot product between encoded queries and predicted CLIP features.

\section{Baselines Details}
\label{sec:baselines-details}

We evaluate several alternative approaches for \taskabbrev, inspired by multi-object tracking~\cite{Weng20iros} and video-instance segmentation~\cite{huang2022minvis} communities. In this section, we provide implementation details for these baselines.

\PAR{Stationary world (SW).} 
As SemanticKITTI~\cite{behley2019iccv} is dominated by static objects, the minimal viable baseline utilizes ego-motion to propagate masks, estimated by a single-scan network~\cite{sal2024eccv}. 
To this end, we first process each point cloud individually using SAL~\cite{sal2024eccv}. 
To associate masks from $P_{t-1}$ $\to$ $P_t$, we perform the following: 
we transform all point clouds in the sequence to a common coordinate frame at time $t$. 
Then, we compute for each point $p_i \in P_t$ a nearest-neighbor $p_j \in P_{t-1}$. 
Then for each instance $\text{id}_i$ that appears in the current frame $P_t$, find all the points $p_i \in P_t$, where $\text{id}(p_{i}) \in \{\text{id}_i\}$.
The corresponding nearest points in the previous frame $p_j \in P_{t-1}$ have $\text{id}(p_j) \in \{\text{id}_{j_1}, \text{id}_{j_2}, \text{id}_{j_3} ...\}$. 
We determine for each instance $\text{id}(p_{i})$ a track ID via majority voting of $\text{id}(p_j)$. The threshold of majority voting is set to 0.5.

\PAR{Multi-object tracking (MOT).} Model-free approaches that utilize Kalman filters in conjunction with linear or greedy association of single-scan object detections are strong baselines for Lidar-based tracking~\cite{Weng20iros}. To this end, we adapt \cite{Weng20iros} to associate masks from SAL~\cite{sal2024eccv}. %
Approach by \cite{Weng20iros} parametrizes object tracks via object-oriented 3D bounding boxes (parametrized via center, bounding box size, and yaw-angle). 
Tracks are propagated from past point clouds to the current state via a constant-velocity Kalman filter, and associations are determined based on 3D intersection-over-union (IoU) between track predictions and detected objects (also parametrized as object-oriented bounding boxes).
We adapt \cite{Weng20iros} in our work by first predicting segmentation masks for each point cloud and then fitting bounding box to each segment (the box boundary are set as the minimum and maximum 3D coordinates of the segmentation masks, $\text{Bbox} = \{ x_{min}, y_{min}, z_{min}, x_{max}, y_{max}, z_{max} \}$).
We report our configuration for~\cite{Weng20iros} in \cref{tab:ab3dmot_parameters}.

\begin{table}[t]
    \centering
    \footnotesize
    \begin{tabular}{l|l}
    \toprule
    Parameter & Value \\
    \midrule
    \multicolumn{2}{c}{AB3DMOT Parameters~\cite{Weng20iros}} \\
    \midrule
    algm & "greedy" \\
    metric & "giou\_3d" \\
    thres & -0.4 \\
    min\_hits & 1 \\
    max\_age & 2 \\
    Ego-motion compensation & Yes \\
    \bottomrule
    \end{tabular}
    \caption{
        \textbf{Multi-object tracking (MOT) baseline}. We report the key hyperparameters used in our adaptation of~\cite{Weng20iros}.
    }
    \label{tab:ab3dmot_parameters}
\end{table}

\PAR{Video Instance Segmentation (VIS)}
This baseline associates objects in 3D without explicit sequence-level training. 
Specifically, we adapt a video instance segmentation approach MinVIS~\cite{huang2022minvis}, that utilizes object queries for associating objects at test time within Lidar data. 
The algorithm operates as follows.
We first generate \( N \) object queries per frame using SAL~\cite{sal2024eccv}.
Then we match queries from frame \( t \) to frame \( t+1 \) using cosine distance as the metric.
Finally, the IDs are transferred based on established matches.
As we only have a limited number of queries, which makes long-term tracking challenging. 
To solve this, we first do MinVIS within a temporal window of size 2 and then employ the same cross-window association as our \method model prediction for post-processing.

\section{Additional Experimental Evaluation}
\label{sec:additional-experimental-evaluation}
\vspace{-2mm}
\begin{algorithm}[h]
\begin{algorithmic}[1]
\Require Lidar point clouds $P_t$, $C$ camera views $\mathcal{I}_t^c$, $C$ camera calibrations $K_c$, timestamps $t \in 1, \ldots, T$
\Ensure $\{\tilde{m}_t, f_t \}, t \in 1, \ldots, T$ 
\For{each timestamp $t$}
    \State $P_t$ $\leftarrow$ load\_lidar($t$)
    \State $\tilde{m}_t = \emptyset$, $f_t = \emptyset$
    \State $\tilde{m}_t^{DBSCAN}$ $\leftarrow$ DBSCAN\_ensamble($P_t$)   
    \For{each camera $c$}
        \State $\mathcal{I}_t^c$ $\leftarrow$ load\_image($t, K_c$)
        \State $m_t^c$ $\leftarrow$ SAM($\mathcal{I}_t^c$) 
        \State $f_t^c \leftarrow$ MaskCLIP~\cite{ding2023open}($\mathcal{I}_t^c$, $m_t^c$)
        \State $\tilde{m}_t^c \leftarrow$ lift\_to\_3D ($P_t^c, m_t^c$, $K_c$) 
        \State $\tilde{m}_t^c \leftarrow$ DBSCAN\_refine($\tilde{m}_t^c, \tilde{m}_t^{DBSCAN}$)
        \State $\tilde{m}_t^c$ $\leftarrow$ flatten\_in\_3D($\tilde{m}_t^c$) 
        \State \{$\tilde{m}_t, f_t \} \leftarrow$ insert\_or\_merge($\tilde{m}_t, f_t, \tilde{m}_t^c$, $f_t^c$)
    \EndFor
\EndFor
\end{algorithmic}
\caption{Single-scan 3D SAL~\cite{sal2024eccv} pseudo-label engine}
\label{alg:sal3d_pseudo_label_engine}
\end{algorithm}
\vspace{-2mm}

\subsection{Pseudo-label Engine Ablations}
\label{appendix:subsection:pseudo_label_engine_ablation}
\noindent\textbf{DBSCAN.} We investigate how to use DBSCAN for segmentation refinement during pseudo-label generation. \cref{tab:pseudolabels_dbscan} shows the effect of doing DBSCAN on per scan separately or on all the scans within the temporal window all together. The temporal window size is set to 2. The best pseudo-label is obtained by only enabling DBSCAN per scan separately. Possibly because doing DBSCAN on all the scans will harm the segmentation performance on dynamic objects, which results in a significant drop in association score ($S_{assoc}$) when enabled. 

\begin{table*}
\centering
\footnotesize
\begin{tabular}{cccccccccc}
\toprule
\# frames & per-frame & all-frame & Frust. Eval. & LSTQ & $S_{assoc}$ & $S_{cls}$ & $IoU_{st}$ & $IoU_{th}$ & \\
\midrule
\multicolumn{9}{c}{Class-agnostic (Semantic Oracle)} \\
\midrule
2 & $\checkmark$ &  & $\checkmark$ & 63.5 & 68.0 & 59.3 & 56.1 & 71.0 & \\
2 &  & $\checkmark$ & $\checkmark$ & 60.8 & 63.9 & 57.8 & 54.9 & 68.9 & \\
2 & $\checkmark$ & $\checkmark$ & $\checkmark$ & 60.5 & 63.5 & 57.7 & 54.4 & 69.4 & \\
\midrule
\multicolumn{9}{c}{Zero-Shot} \\
\midrule
2 & $\checkmark$ &  & $\checkmark$ & 46.3 & 66.4 & 32.3 & 34.1 & 33.9 & \\
2 &  & $\checkmark$ & $\checkmark$ & 44.6 & 62.6 & 31.8 & 33.6 & 33.3 & \\
2 & $\checkmark$ & $\checkmark$ & $\checkmark$ & 44.2 & 62.0 & 31.5 & 33.2 & 33.1 & \\
\bottomrule
\end{tabular}
\caption{\textbf{Pseudo-label ablations on DBSCAN settings, per-frame or all-frame:} We show the effect of doing DBSCAN per scan separately or on all the scans within the temporal window together on the KITTI validation set. The temporal window size is set to 2. The results show that doing DBSCAN per-frame gives the best result.
}
\label{tab:pseudolabels_dbscan}
\end{table*}

\subsection{Single-scan SAL pseudo-label improvements}
In the process of developing \method, we also re-think and improve the single-scan 3D pseudo-labels proposed in~\cite{sal2024eccv}.
Training on these labels yields the 3D SAL model results reported in the main paper (we report improved results, compared to those reported in~\cite{sal2024eccv}).
We formalize our novel single-scan label engine in~\cref{alg:sal3d_pseudo_label_engine} and ablate the performance boosts for class-agnostic and zero-shot \lps (LPS) of the following improvements in~\cref{tab:pseudolabels_single_scan_improvements}.

\begin{table}[t]
\centering
\footnotesize
\resizebox{\linewidth}{!}{
\begin{tabular}{lccccH}
\toprule
Single-scan 3D pseudo-labels & PQ & SQ & PQ\textsubscript{th} & PQ\textsubscript{st} & mIoU \\
\midrule
\multicolumn{6}{c}{Class-agnostic (Semantic Oracle) LPS} \\
\midrule
Original                        & 48.7 & 73.7 & 53.1 & 45.4 & XX.X \\
+ Flatten in 3D                 & 51.8 & 78.3 & 62.1 & 44.4 & XX.X \\
+ DBSCAN refine per instance    & 53.6 & 80.1 & 65.2 & 45.2 & XX.X \\
+ Flatten via coverage          & 55.3 & 79.9 & 66.0 & 47.5 & 62.1 \\
\midrule
\multicolumn{6}{c}{Zero-Shot LPS} \\
\midrule
Original                        & 27.5 & 71.5 & 31.7 & 24.5 & XX.X \\
+ Flatten in 3D                 & 28.6 & 73.4 & 34.0 & 24.7 & XX.X \\
+ DBSCAN refine per instance    & 29.7 & 75.1 & 36.0 & 25.1 & XX.X \\
+ Flatten via coverage          & 29.9 & 74.8 & 35.2 & 26.0 & 31.9 \\
\bottomrule
\end{tabular}
}
\caption{
    \textbf{Single-scan 3D pseudo-label improvements:} 
    We report class-agnostic and zero-shot single-scan \lps (LPS) results with several improvements added to the original~\cite{sal2024eccv} pseudo-labels. 
    Evaluation is performed in the camera frustum of the \textit{SemanticKITTI} validation set.
}
\label{tab:pseudolabels_single_scan_improvements}
\end{table}

\begin{table}[t]
\centering
\footnotesize
\begin{tabular}{HcHccccccc}
\toprule
Method & \# frames & \# stride & LSTQ & $S_{assoc}$ & $S_{cls}$ & $IoU_{st}$ & $IoU_{th}$ & \\
\midrule
Pseudo-label & 2 & 1 & 30.0 & 31.1 & 28.9 & 31.9 & 29.5 & \\
Pseudo-label & 4 & 2 & 27.6 & 26.9 & 28.4 & 31.6 & 28.7 & \\
\bottomrule
\end{tabular}
\caption{\textbf{Pseudo-label ablations on nuScenes dataset on temporal window size:} We ablate on temporal window sizes $2-4$ frames. The quality of pseudo labels with 4 frame temporal window drops significantly. The stride is set as half the window size.
}
\label{tab:nuscenes_pseudolabels_window_size}
\end{table}

\paragraph{Flatten in 3D.}
In contrast to~\cite{sal2024eccv}, we switch the order of Flatten--Lift to Lift--Flatten, \ie, perform the flattening of overlapping SAM~\cite{kirillov2023segment} masks after and not before their unprojection to 3D.
To this end, we apply a non-maximum suppression (NMS) in 3D for the $flatten\_in\_3D$ step in line 11 of~\cref{alg:sal3d_pseudo_label_engine}.
Lift--Flatten has the advantage of resolving potentially ambiguous or edge-case overlaps in the 2D image after their unprojection to the actual 3D geometry.
Furthermore, we can run our DBSCAN refinement before the flattening.
The performance boost of $+3.1$ PQ is particularly noticeable for class-agnostic segmentation.

\paragraph{DBSCAN refine per instance.}
The original DBSCAN refinement step in~\cite{sal2024eccv} creates an ensemble of DBSCAN segments (line 4 in~\cref{alg:sal3d_pseudo_label_engine}) by first removing the ground plane and then collecting the segments of a set of epsilon density parameters.
Afterward, each SAM-based 3D mask (line 9 in~\cref{alg:sal3d_pseudo_label_engine}) with a sufficiently large IoU is replaced with a DBSCAN segment.
This step refines the image-based segments and removes false positives or adds false negatives caused by wrong SAM predictions or unprojection/parallax errors.
Since DBSCAN can only make statements on non-ground plane points, any ground point is added back to its original 3D instance.

Our improved DBSCAN refinement mitigates this issue and removes potential false positives even in the ground plane.
To this end, we run an additional DBSCAN segmentation on each previously replaced instance.
We remove all points that do not belong to the instance, keep potential ground points, and use the same epsilon density value that produced the original DBSCAN replacement mask.
Using the same epsilon, we introduce an expected density prior that allows us to remove all ground points following a different distribution.
The additional per-instance refinement improves class-agnostic and zero-shot \lps performance by $+1.8$ and $+1.1$ PQ, respectively.

\paragraph{Flatten via coverage.}
Our final improvement of the single-scan label engine changes the matching metric for the 3D NMS applied during flattening (line 11 in~\cref{alg:sal3d_pseudo_label_engine}).
Instead of IoU, we compute coverage (intersection-over-minimum) which removes any mask significantly covered by another mask independently of the relative mask sizes.
Flattening via coverage removes many small noisy segments, for example, on large road segments.
In particular, class-agnostic segmentation performance improves by $+1.7$ PQ points.

\begin{table}[t]
\centering
\footnotesize
\begin{tabular}{HcHccccc}
\toprule
Method & \# frames & Cross & LSTQ & $S_{assoc}$ & $S_{cls}$ & $IoU_{st}$ & $IoU_{th}$  \\
 &  & window &  &  &  &  &  \\
\midrule
Pseudo-label  & 2 &  & 46.3 & 66.4 & 32.3 & 34.1 & 33.9 \\
Pseudo-label  & 4 &  & 48.0 & 68.9 & 33.5 & 35.3 & 35.1 \\
Pseudo-label  & 8 &  & 49.2 & 70.0 & 34.6 & 36.0 & 36.9  \\
Pseudo-label  & 16 &  & 49.9 & 70.0 & 35.6 & 36.4 & 39.0\\
\bottomrule
\end{tabular}
\caption{\textbf{Pseudo-label ablations on temporal window size without cross window association:} 
We ablate our approach on temporal window sizes of size $K=\{2,4,8, 16\}$ with stride $\frac{K}{2}$ on \textit{SemanticKITTI} validation set. 
We got a similar observation as the ablation study on the cross-associated version of pseudo-labels that the association score ($S_{assoc}$) improves up to $8$ frames, while zero-shot recognition does not saturate and continues to improve as the temporal window size increases.
}
\label{tab:pseudolabels_window_size_nonstitched}
\end{table}

\begin{table}[t]
\centering
\footnotesize
\begin{tabular}{lcccccc}
\toprule
Method  & label & LSTQ & $S_{assoc}$ & $S_{cls}$ & $IoU_{st}$ & $IoU_{th}$  \\
\midrule
\method  & $v_1$ & 50.7 & 67.2 & 38.3 & 48.7 & 28.8 \\
\method  & $v_2$ & 53.2 & 77.2 & 36.6 & 47.9 & 25.6 \\
\bottomrule
\end{tabular}
\caption{\textbf{4DSAL ablations on training on different version of labels:} 
We ablate our model on training on different versions of labels on \textit{SemanticKITTI}. The temporal window size is set to $K=8$ with stride $4$. 
Pseudo-label $v_1$: the pseudo-labels are not associated cross window (\ie, the semantic features are aggregated per window). 
Pseudo-label $v_2$: the pseudo-labels are associated cross window (\ie, the semantic features are aggregated over the whole sequence). 
}
\label{tab:4dsal_model_ablation_w_o_stitch}
\end{table}

\subsection{Per-Class Results}

We report per-class results for Zero-Shot Lidar Panoptic Segmentation (PQ) in~\cref{tab:kitti_nuscenes-baselines-per_class}. 
Remarkably, not only we consistently outperform SAL~\cite{sal2024eccv} on (almost) all classes on both, \textit{SemanticKITTI} and \textit{Panoptic nuScenes} -- we show we can localize and recognize even instances that the single-scan model by~\cite{sal2024eccv} (\texttt{motorcyclist}, \texttt{cyclist}, \texttt{barrier}) is unable to segment. 

\begin{table}[t]
\centering
\footnotesize
\resizebox{\linewidth}{!}{
\begin{tabular}{clcccccccc}
\toprule
& Method & \# frames & Franken &  LSTQ & $S_{assoc}$ & $S_{cls}$ & $IoU_{st}$ & $IoU_{th}$ \\
&  &  & Frustum &  &  &  &  &  \\
\midrule
& pseudo-labels &  & $\times$ & 5.8 & 4.0 & 8.4 & 6.9 & 11.7 \\
& \method & 2 & $\times$ & 8.3 & 5.4 & 12.7 & 20.2 & 2.3 \\
& \method & 2 & $\checkmark$ & 42.2 & 51.1 & 34.9 & 45.1 & 20.8 \\
\bottomrule
\end{tabular}
}
\caption{\textbf{\method on SemanticKITTI validation set, full ($360^{\circ}$) point cloud evaluation.} On SemanticKITTI, only $14\%$ of all Lidar points are seen in the left RGB camera, used for pseudo-labeling. Due to low coverage, when we evaluate pseudo-labels, we obtain $LSTQ$ of $5.8$ (low recall). It is critical to train the model using \textit{FrankenFrustum} augmentation to obtain a good generalization to the whole point cloud ($42.2$ $LSTQ$) -- only employing standard data augmentations (rotation, translation, scaling) is not sufficient ($8.3$ $LSTQ$).}
\label{tab:4dsal_model_ablation_outsidefrustum}
\end{table}

\subsection{Per-Window \vs. Per-Sequence Labels}

\cref{tab:pseudolabels_window_size_nonstitched} evaluates pseudo-labels $v_1$ \wrt window size, without cross-window association (\ie, the semantic features are aggregated per window). In the main paper, we report $v_2$ labels that additionally apply cross-window association (\ie, the semantic features are aggregated over the whole sequence). 
We observe similar trends, that association performance ($S_{assoc}$) improvements saturate at window sizes of 8, while zero-shot recognition ($S_{cls}$) benefits from a larger temporal span. However, overall, we obtain better results with $v_2$ labels, as reported in the main paper. 
    
This is also reflected in \cref{tab:4dsal_model_ablation_w_o_stitch}, where we train our model with $v_1$ and $v_2$ pseudo-labels. 
With $v_2$, we obtain overall higher $LSTQ$ ($53.2$), compared to $v_1$ ($50.7$). 
We observe that training on the cross window associated version of the pseudo-label improves significantly on association score $S_{assoc}$ by about 15\%, which demonstrates that our cross window associated pseudo label, accounting for objects entering the scene, provides precisely the supervisory signal for 4D Lidar segmentation. 
We note that while cross-window association significantly improves the association aspect, we observe a less severe drop in terms of zero-shot recognition ($-1.7$ $S_{cls}$). 

\subsection{Franken Frustum}

\cref{tab:4dsal_model_ablation_outsidefrustum} shows the generalization ability of our model and the importance of applying Franken Frustum data augmentation. 
The results show that if we only train on 14\% of the labeled data, the model doesn't generate well when evaluated on the full point cloud (8.3 LSTQ) even with standard data augmentation. By additionally employing Franken Frustum augmentation, the model generates well outside of the camera Frustum and achieves 42.2 LSTQ. 

\section{Qualitative Results}
\label{sec:additional-qualitative-results}
\begin{figure*}[t] \centering
    \makebox[0.280\textwidth]{\footnotesize GT}
    \makebox[0.280\textwidth]{\footnotesize Pseudo-labels}
    \makebox[0.280\textwidth]{\footnotesize \method}
    \\
    \includegraphics[width=0.280\textwidth]{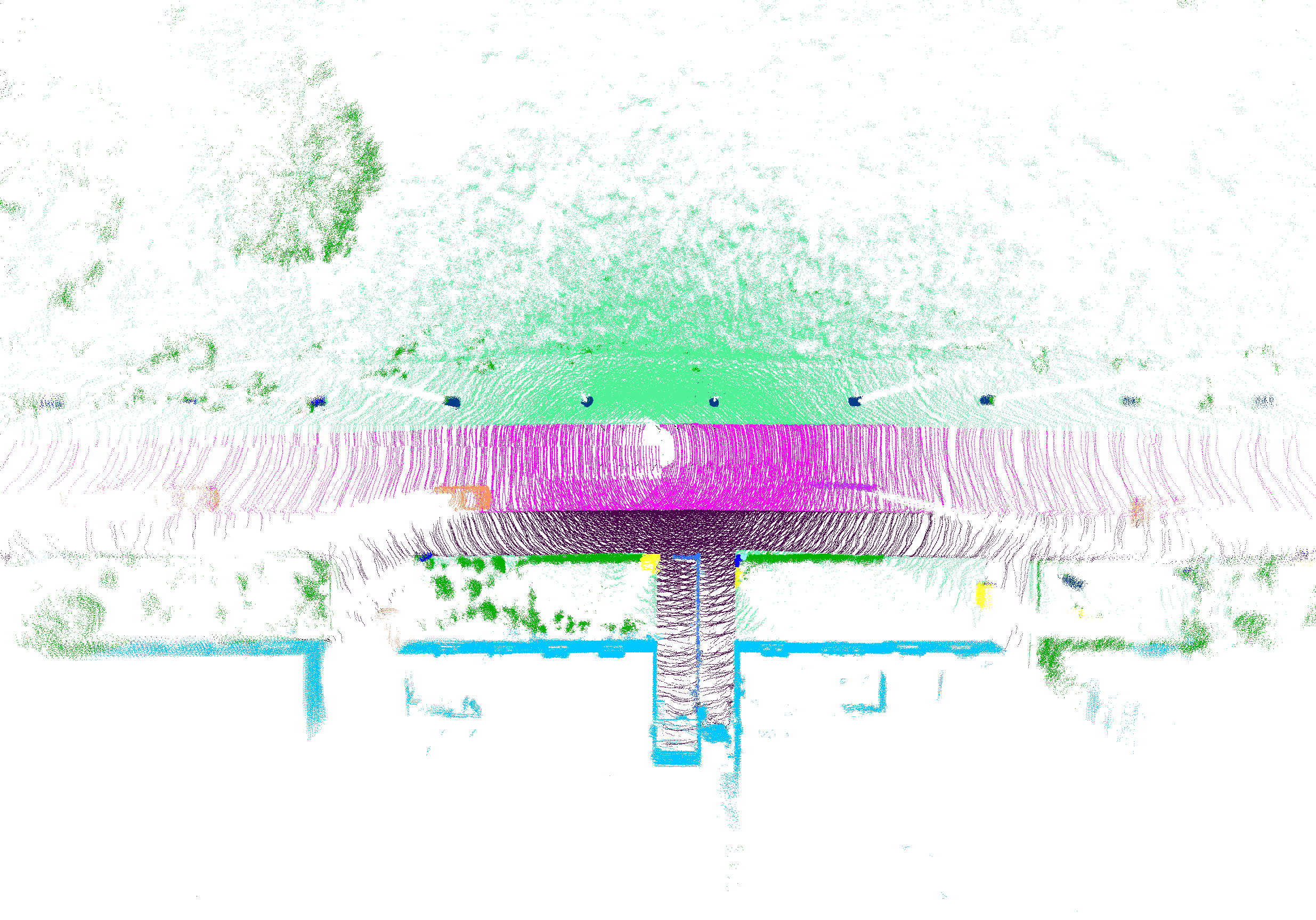}
    \includegraphics[width=0.280\textwidth]{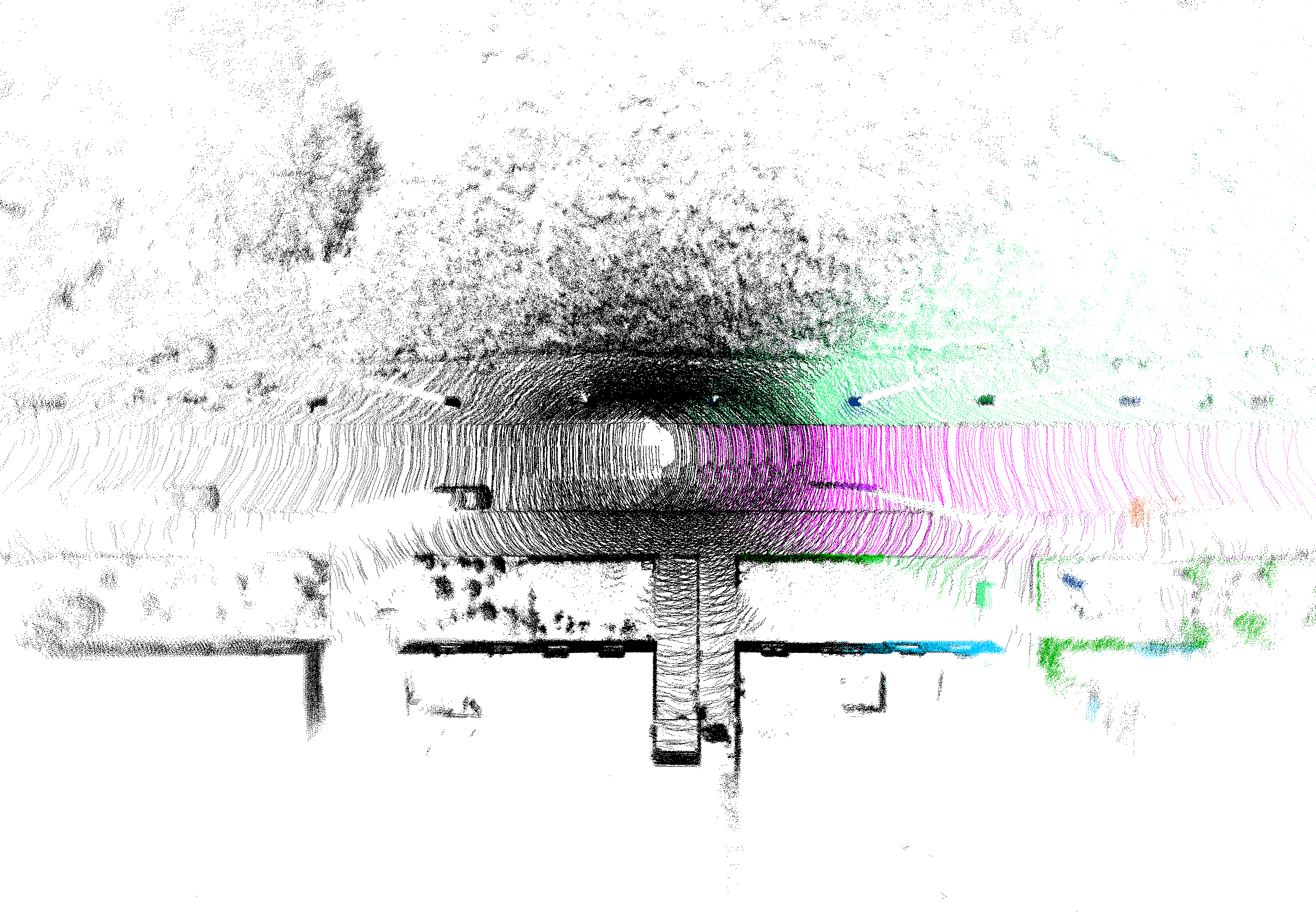}
    \includegraphics[width=0.280\textwidth]{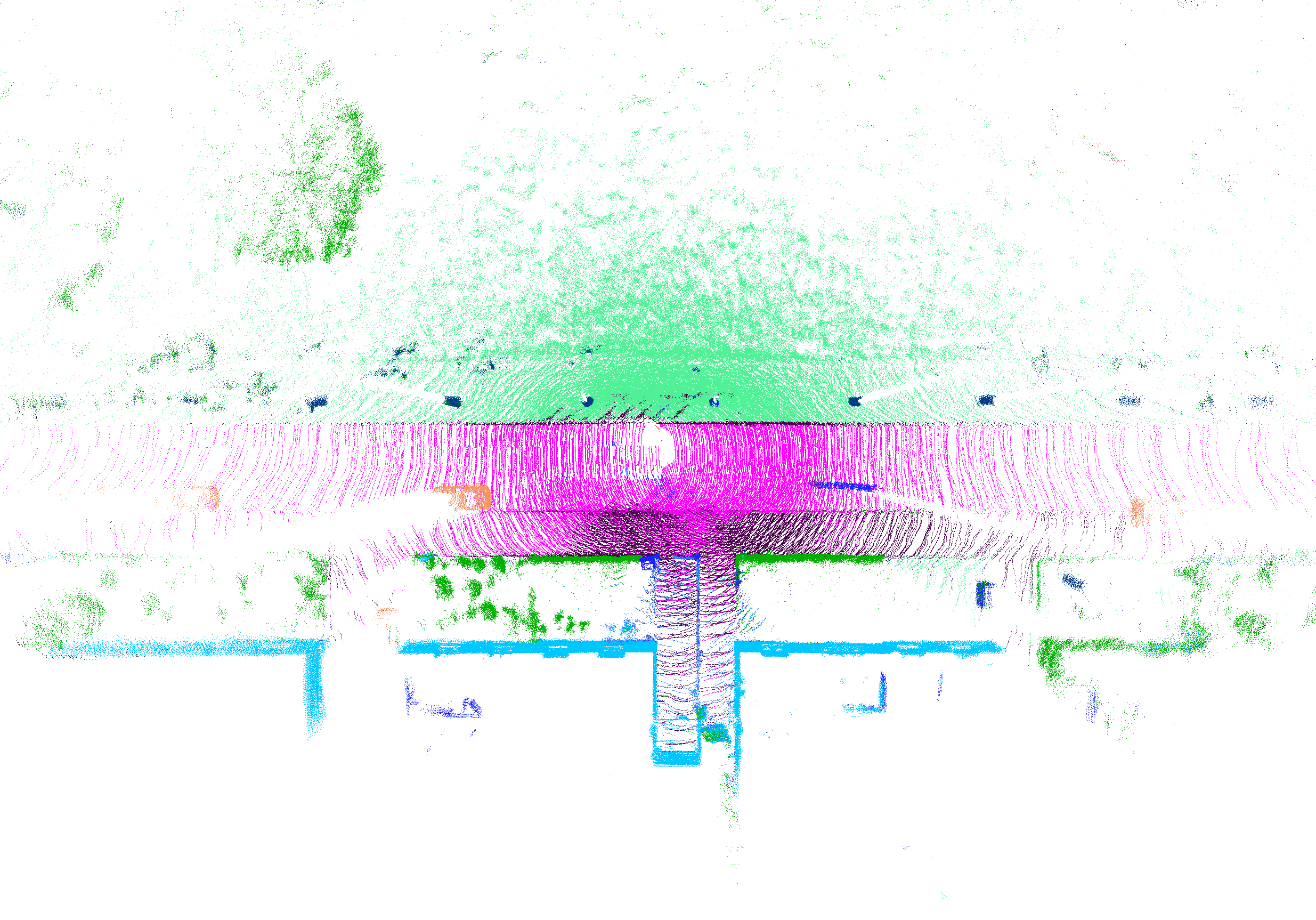}
    \\
    \includegraphics[width=0.280\textwidth]{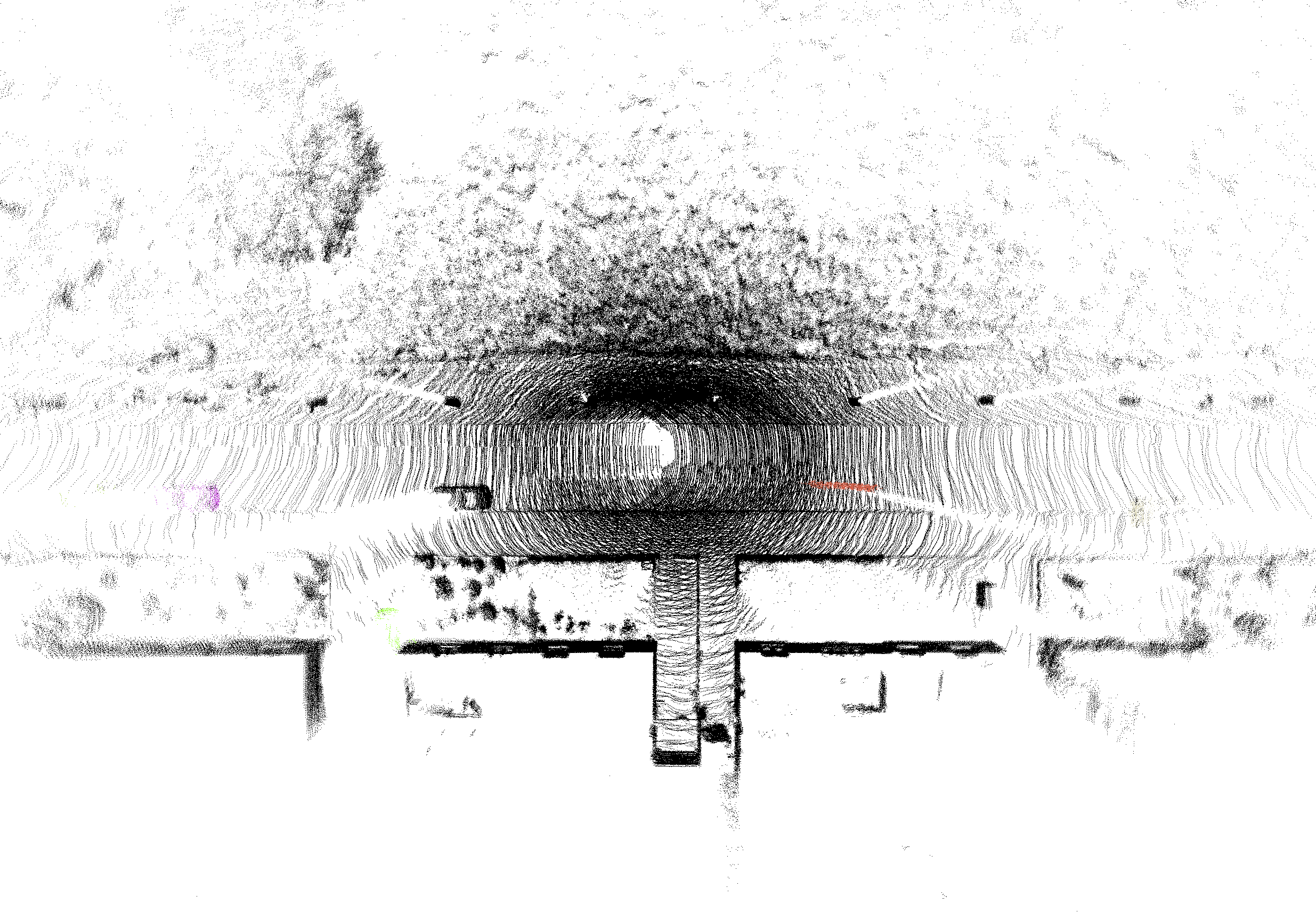}
    \includegraphics[width=0.280\textwidth]{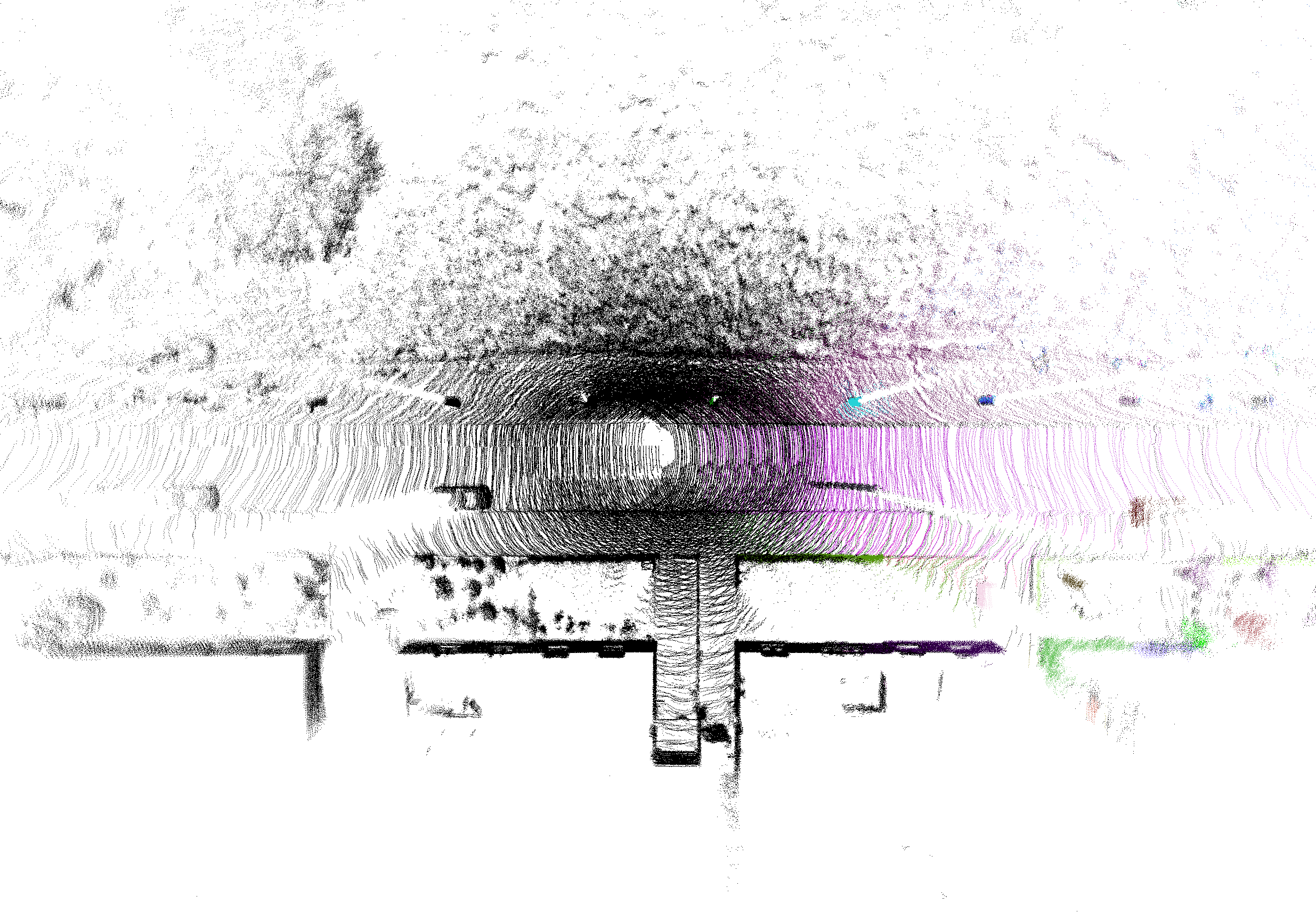}
    \includegraphics[width=0.280\textwidth]{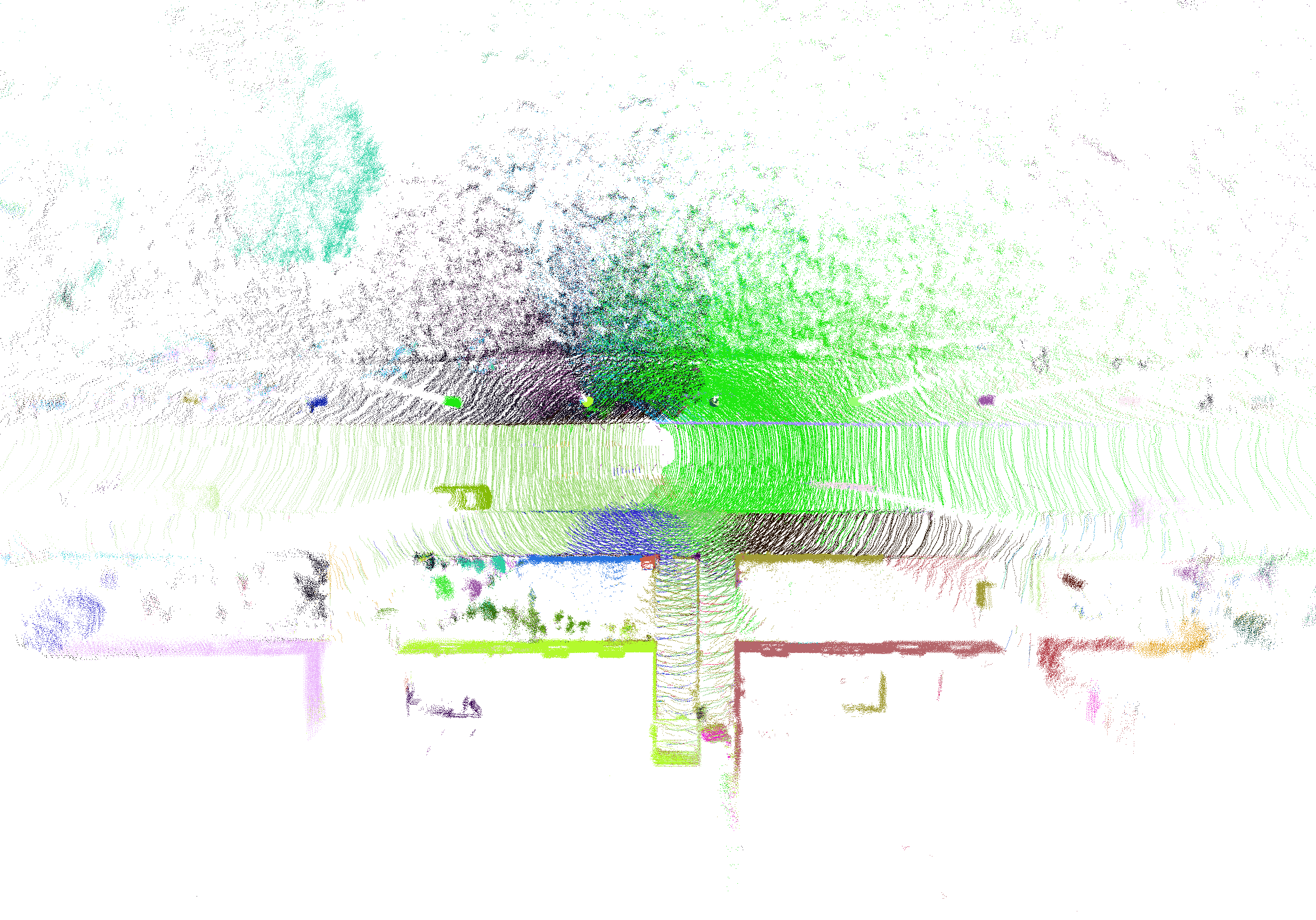}
    \\
    \makebox[0.280\textwidth]{\footnotesize GT}
    \makebox[0.280\textwidth]{\footnotesize Pseudo-labels}
    \makebox[0.280\textwidth]{\footnotesize \method}
    \\
    \includegraphics[width=0.280\textwidth]{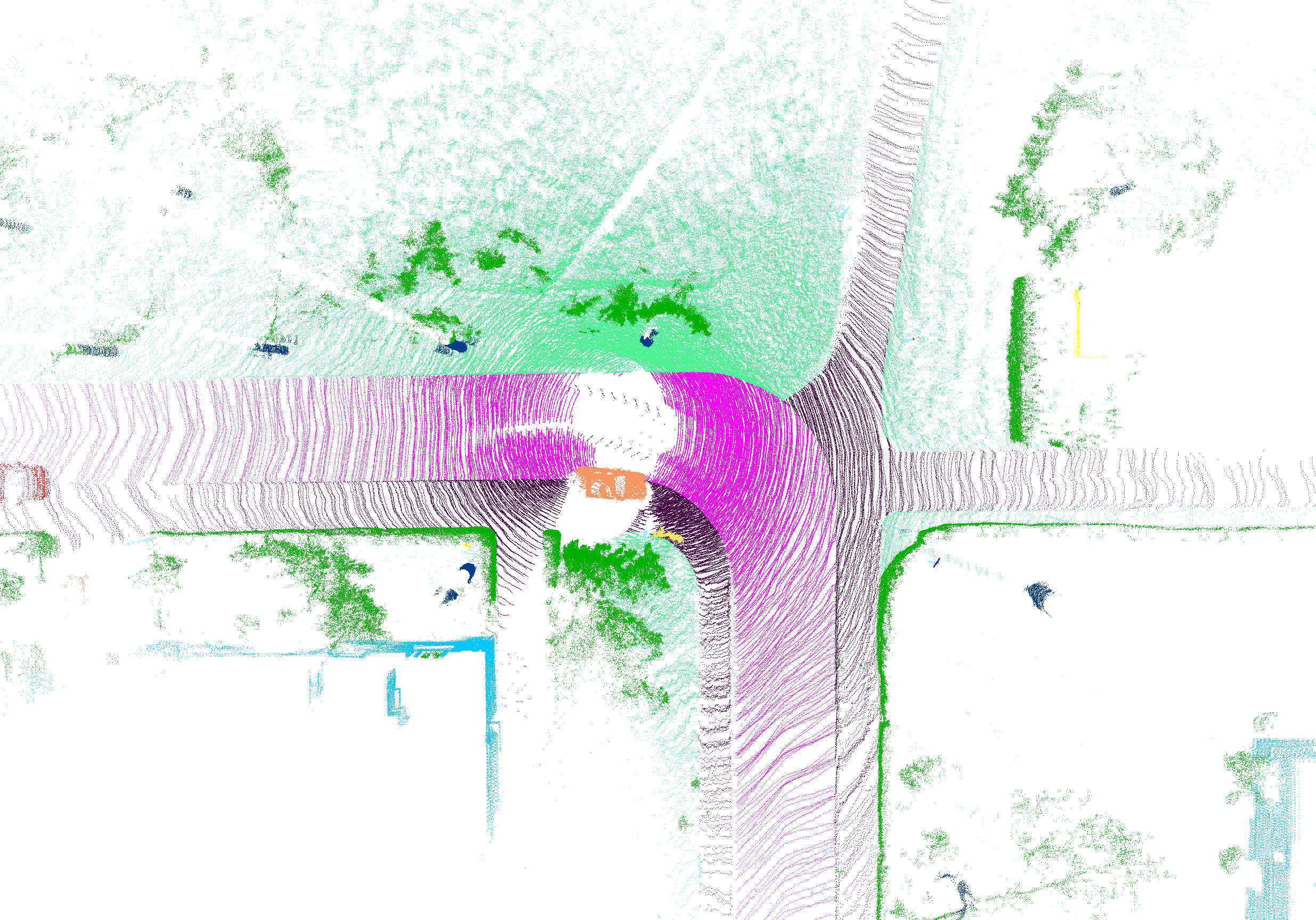}
    \includegraphics[width=0.280\textwidth]{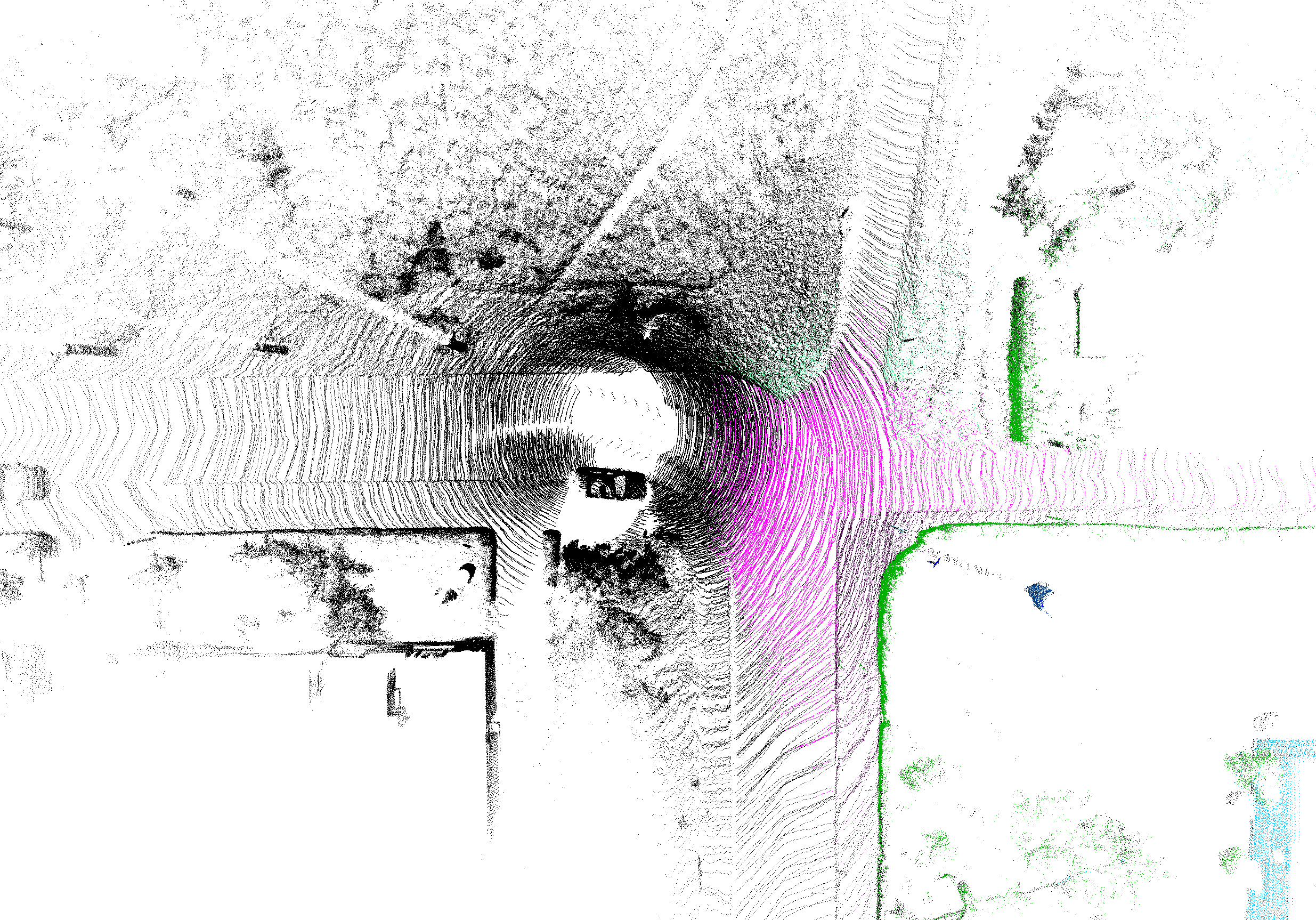}
    \includegraphics[width=0.280\textwidth]{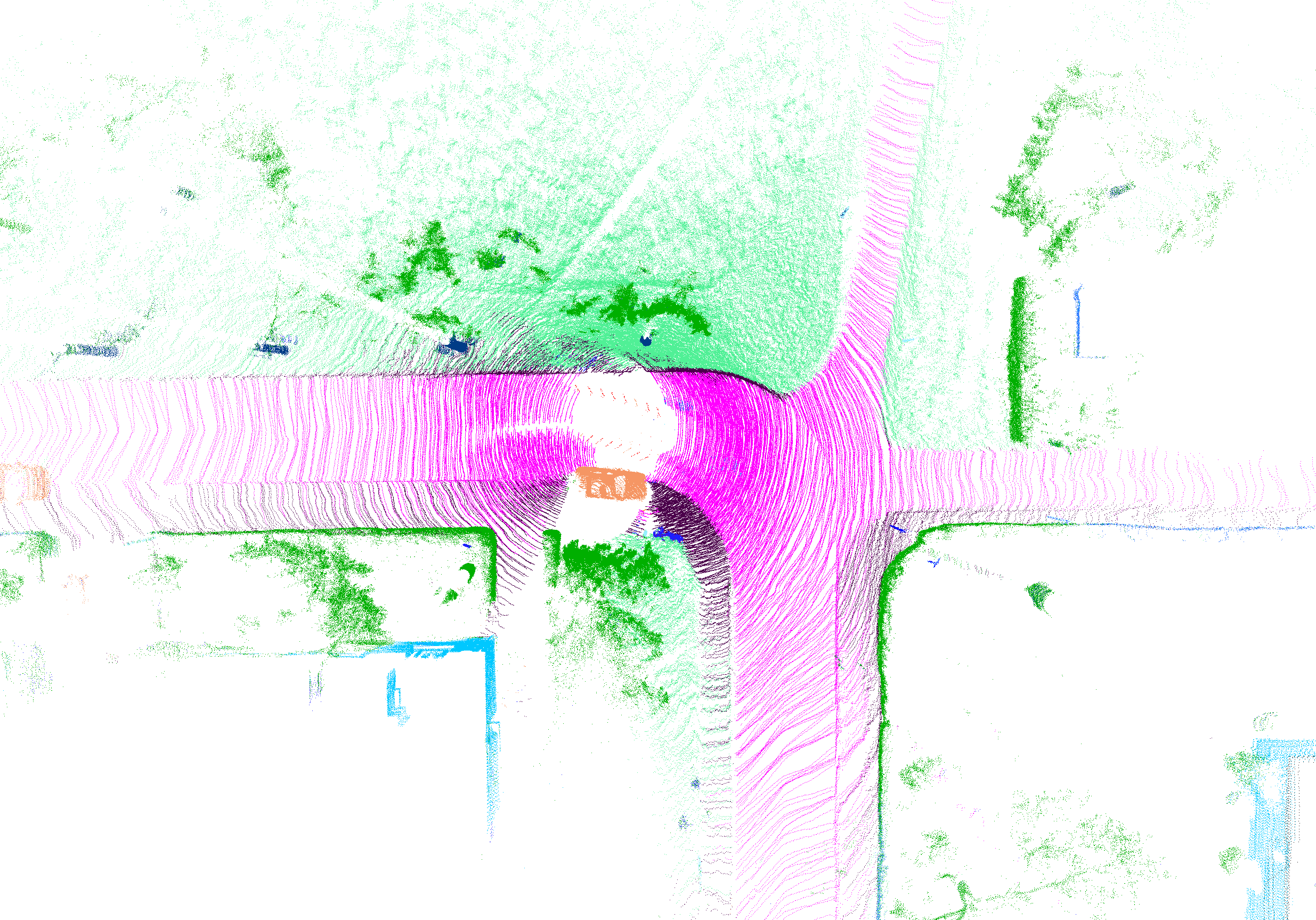}
    \\
    \includegraphics[width=0.280\textwidth]{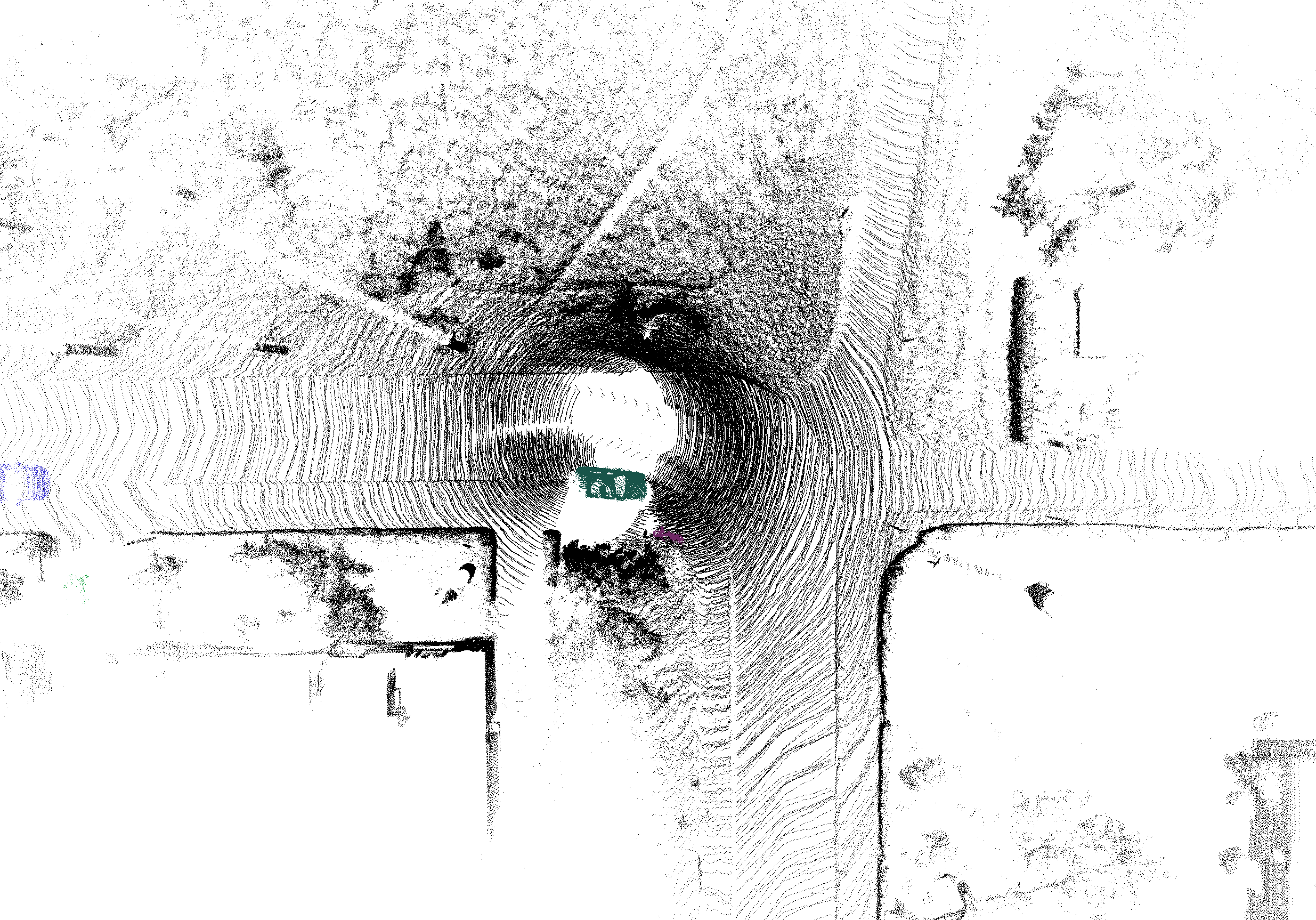}
    \includegraphics[width=0.280\textwidth]{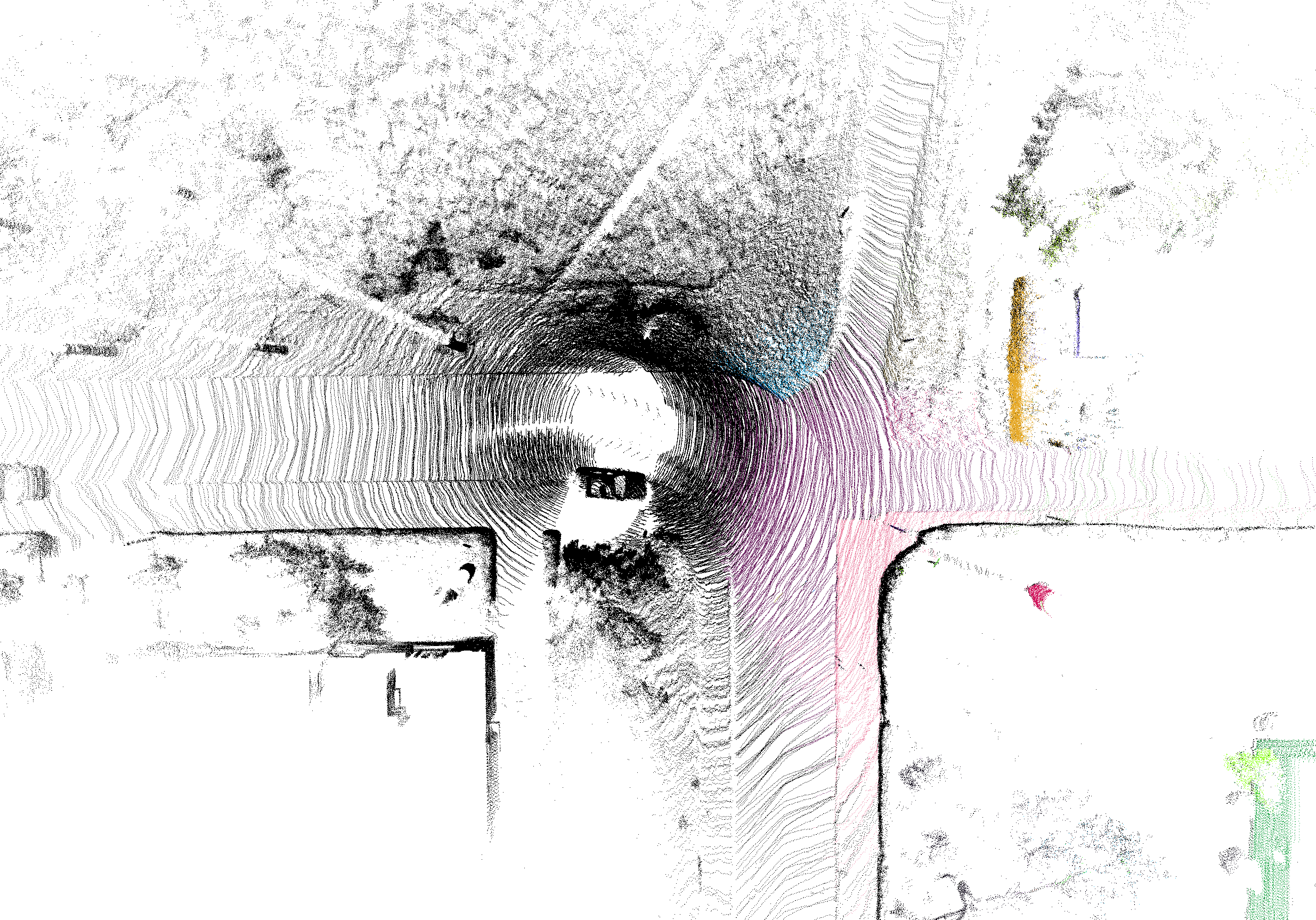}
    \includegraphics[width=0.280\textwidth]{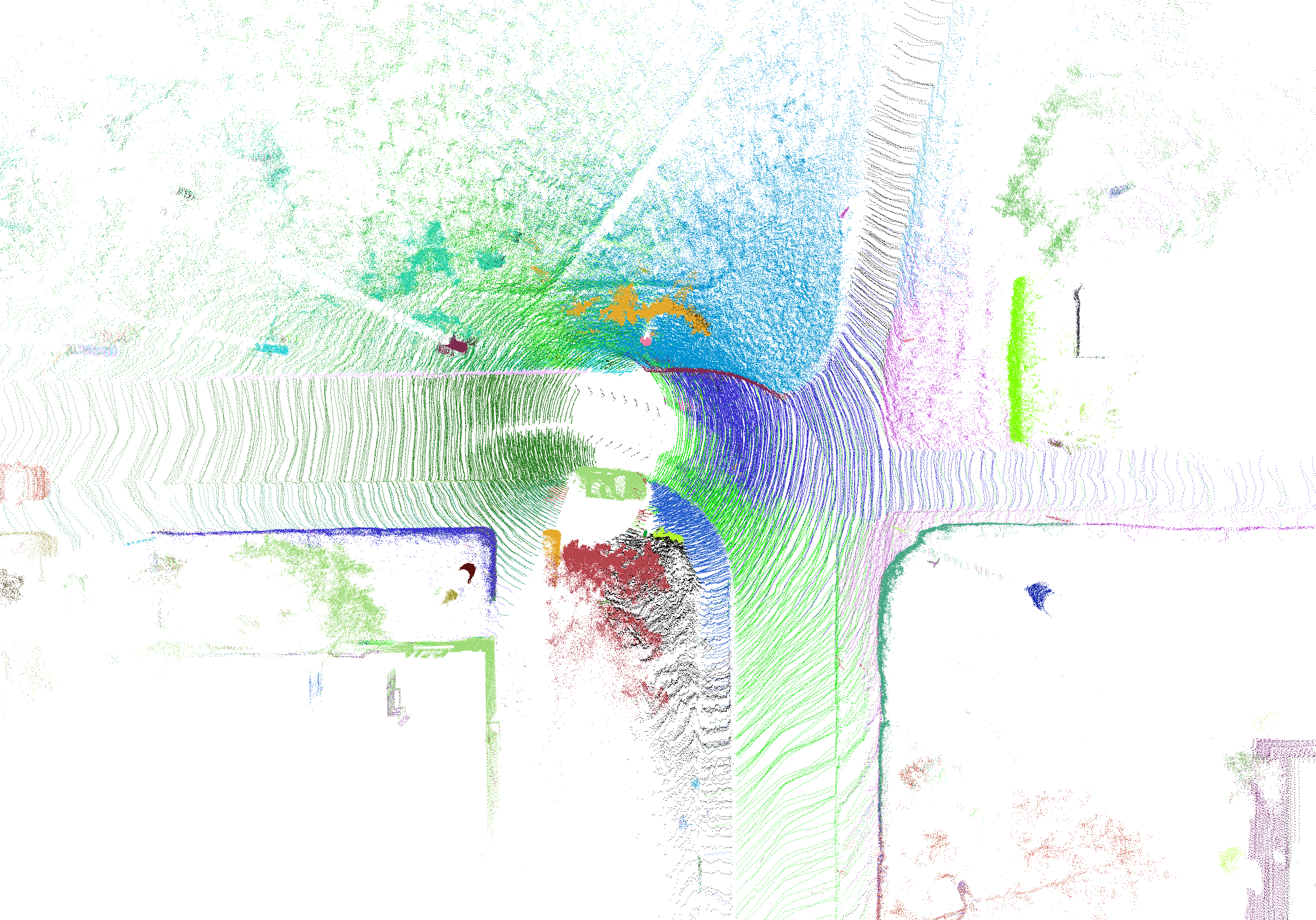}
    \\
    \makebox[0.280\textwidth]{\footnotesize GT}
    \makebox[0.280\textwidth]{\footnotesize Pseudo-labels}
    \makebox[0.280\textwidth]{\footnotesize \method}
    \\
    \includegraphics[width=0.280\textwidth]{img/viz_kitti/scene_08_000600/000600_semantics_ground_truth.ply02.png}
    \includegraphics[width=0.280\textwidth]{img/viz_kitti/scene_08_000600/000600_semantics_pseudo_label_prompted.ply02.png}
    \includegraphics[width=0.280\textwidth]{img/viz_kitti/scene_08_000600/000600_semantics_sal4d.ply02.png}
    \\
    \includegraphics[width=0.280\textwidth]{img/viz_kitti/scene_08_000600/000600_instances_ground_truth.ply02.png}
    \includegraphics[width=0.280\textwidth]{img/viz_kitti/scene_08_000600/000600_instances_pseudo_label.ply02.png}
    \includegraphics[width=0.280\textwidth]{img/viz_kitti/scene_08_000600/000600_instances_sal4d.ply02.png}
    \\
    \vspace{-15pt}
    \caption{\textbf{Qualitative results on SemanticKITTI}. We show ground-truth (GT) labels (\textit{first column}), our pseudo-labels (\textit{middle column}), and \method results (\textit{right column}). We show three scenes (we superimpose point clouds). For each, we show semantic predictions in the \textit{first row} and instances predictions in the \textit{second row}. \textbf{Importantly, we visualize semantics for pseudo-labels via zero-shot prompting whereas pseudo-labels do not provide explicit semantic labels, only CLIP tokens.}} 
    \label{fig:viz_kitti}
\end{figure*}

\begin{figure*}[t] \centering
    \makebox[0.250\textwidth]{\footnotesize GT}
    \makebox[0.250\textwidth]{\footnotesize Pseudo-labels}
    \makebox[0.250\textwidth]{\footnotesize \method}
    \\
    \includegraphics[width=0.250\textwidth]{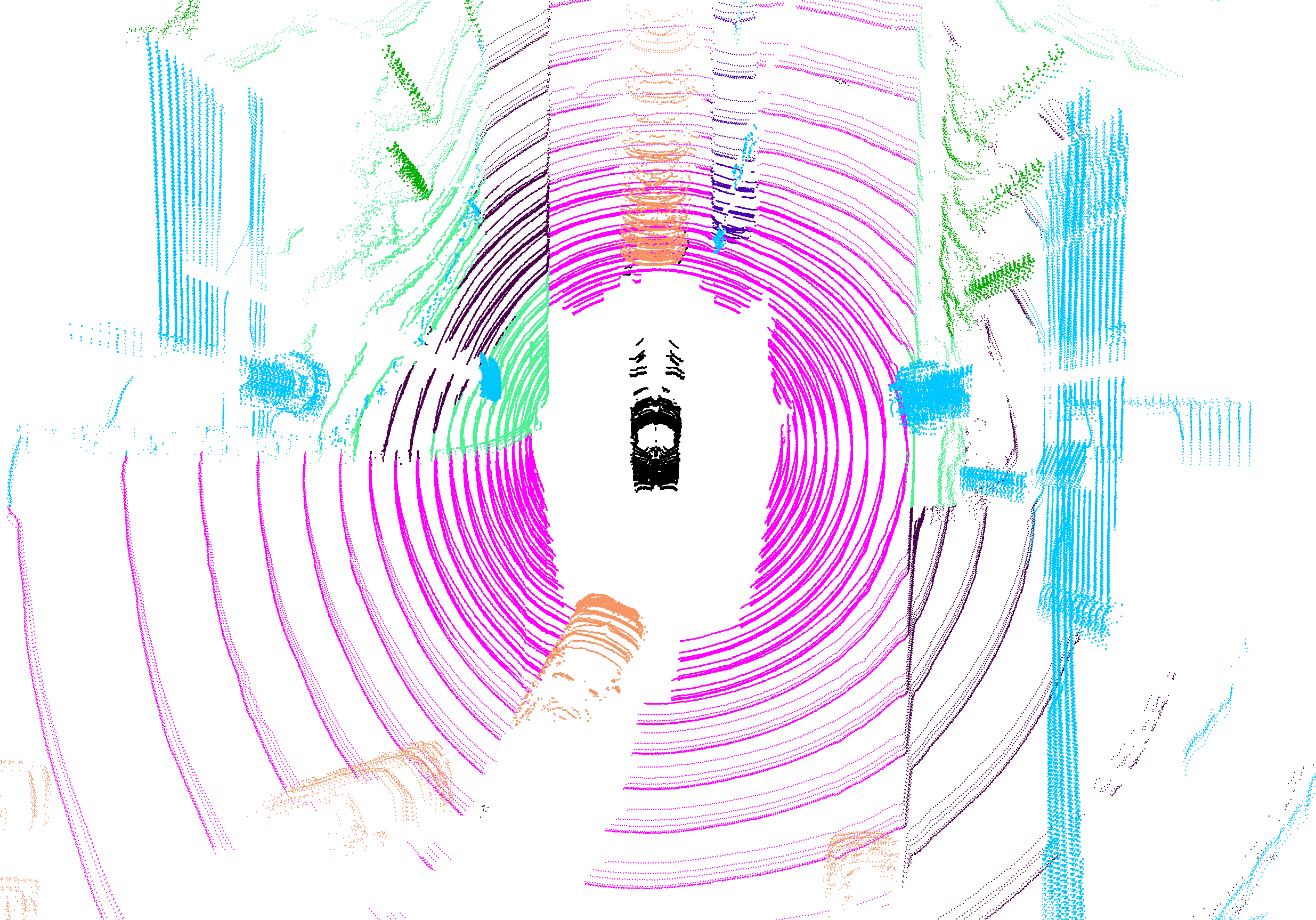}
    \includegraphics[width=0.250\textwidth]{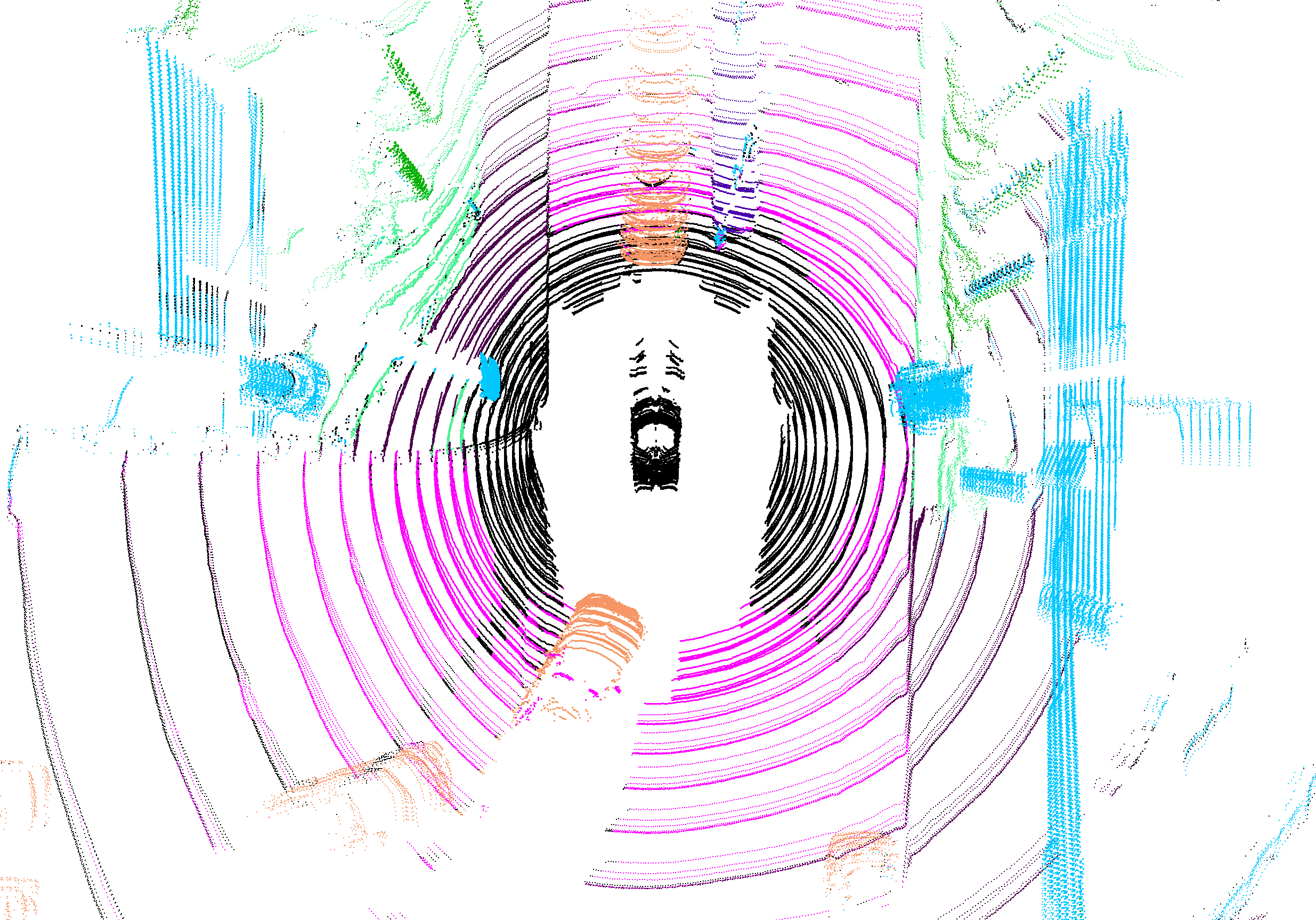}
    \includegraphics[width=0.250\textwidth]{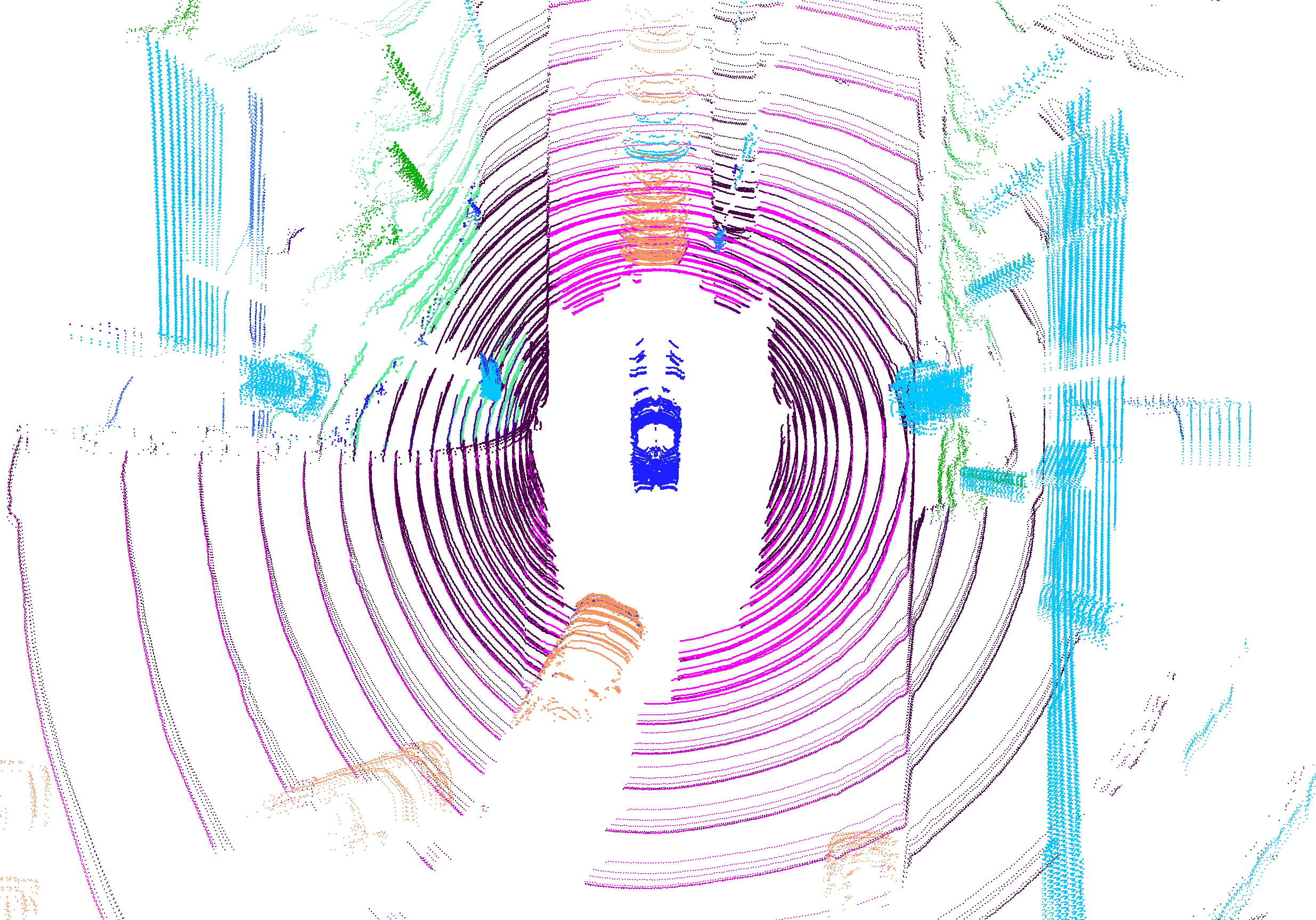}
    \\
    \includegraphics[width=0.250\textwidth]{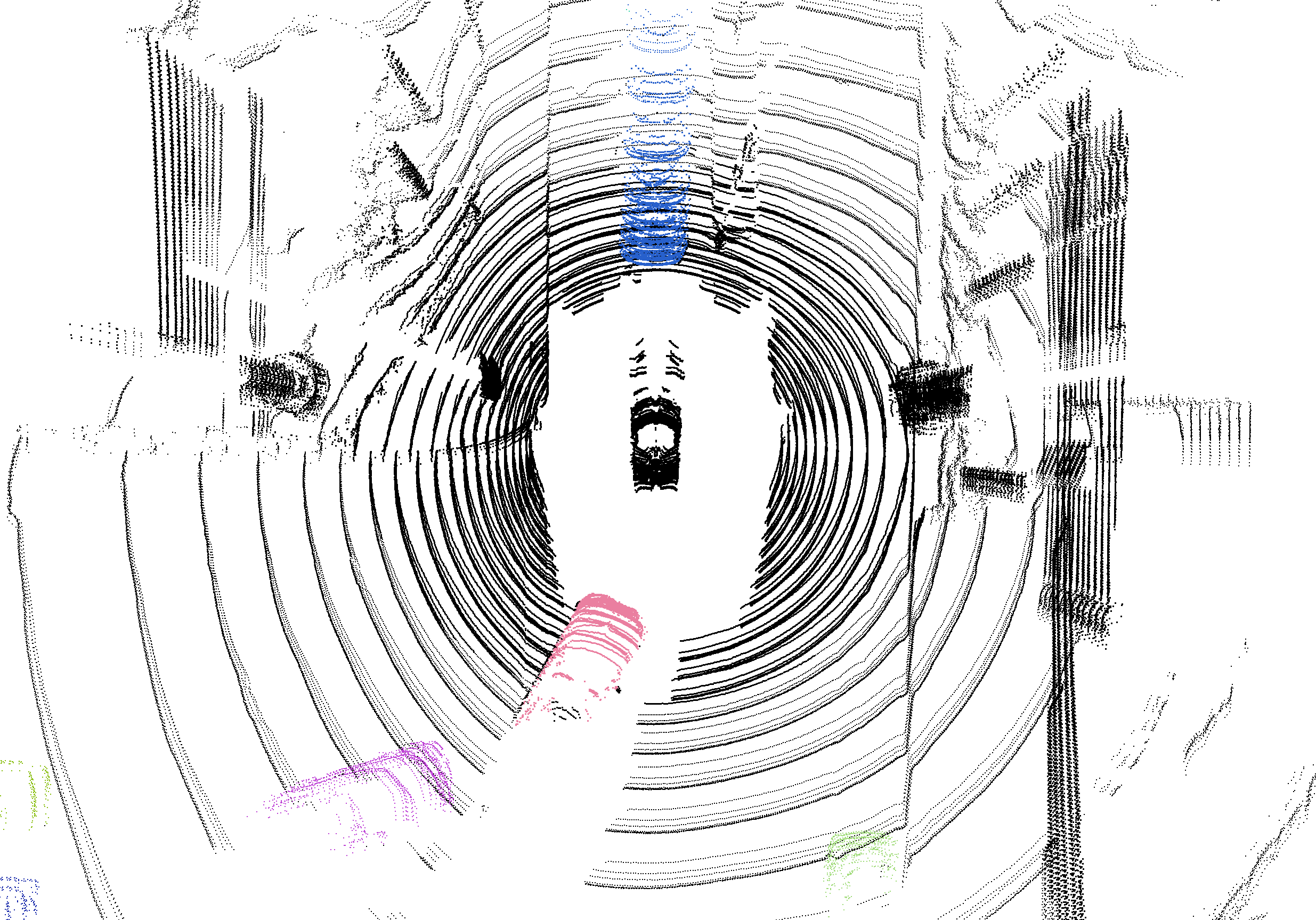}
    \includegraphics[width=0.250\textwidth]{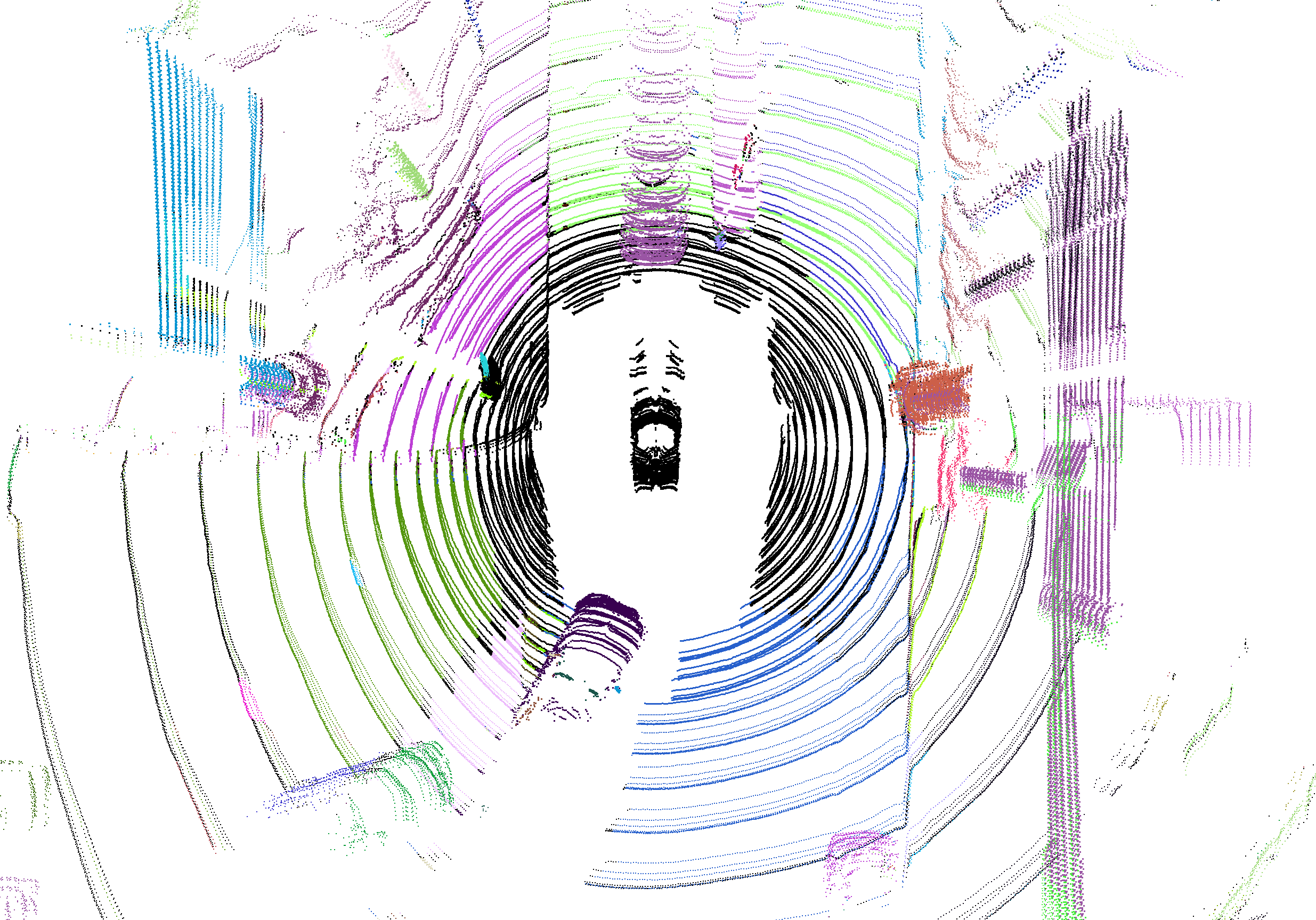}
    \includegraphics[width=0.250\textwidth]{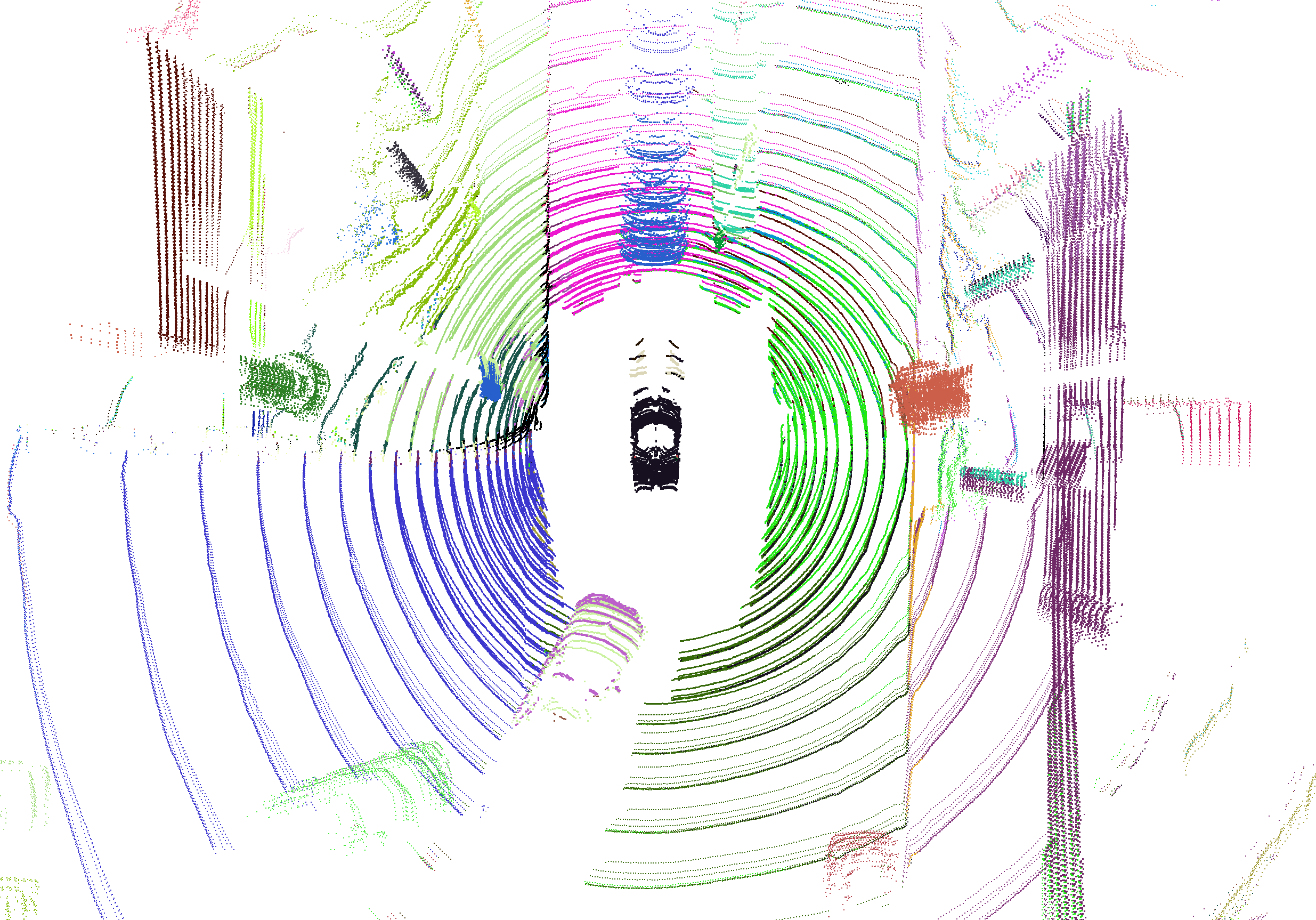}
    \\
    \makebox[0.250\textwidth]{\footnotesize GT}
    \makebox[0.250\textwidth]{\footnotesize Pseudo-labels}
    \makebox[0.250\textwidth]{\footnotesize \method}
    \\
    \includegraphics[width=0.250\textwidth]{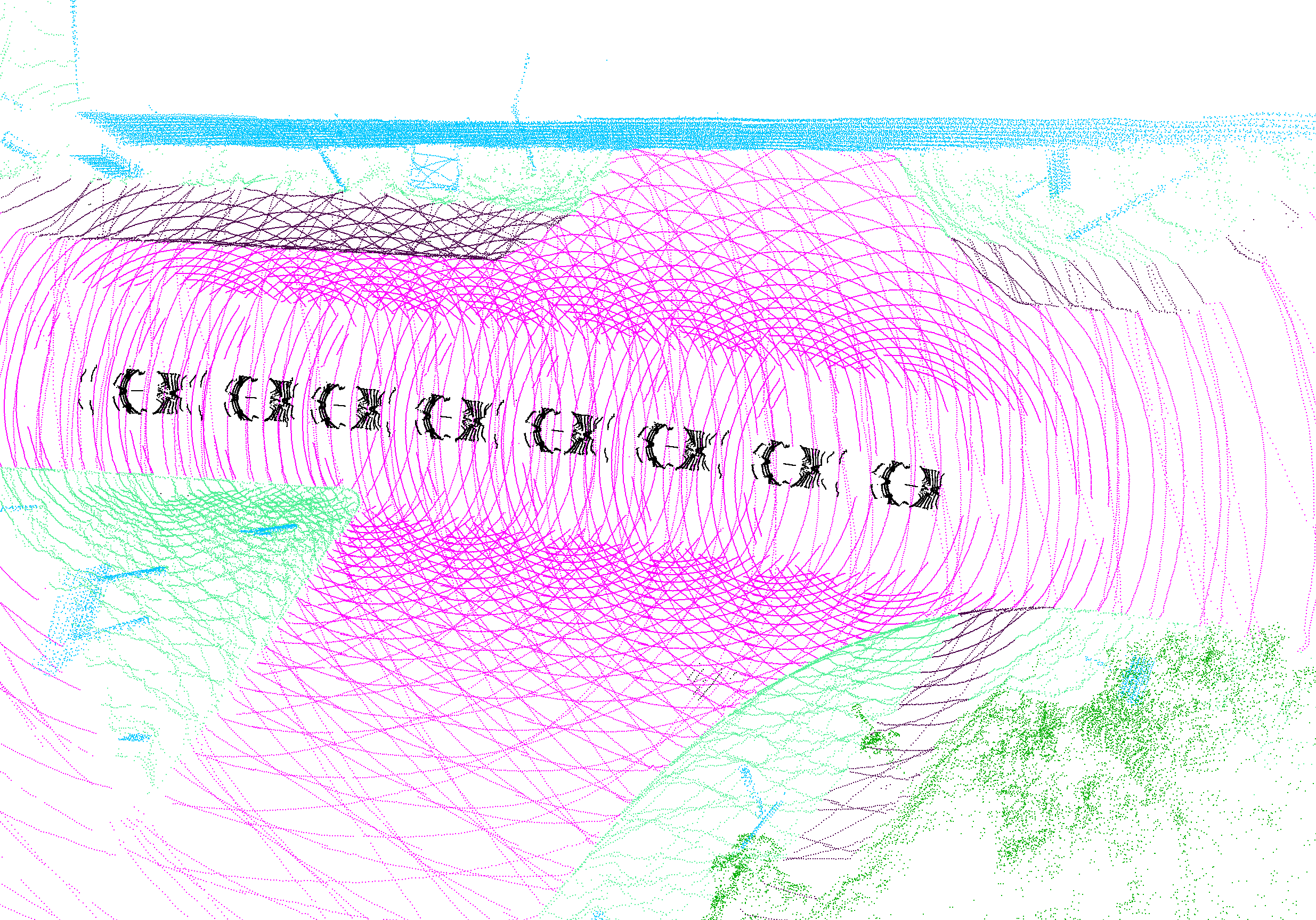}
    \includegraphics[width=0.250\textwidth]{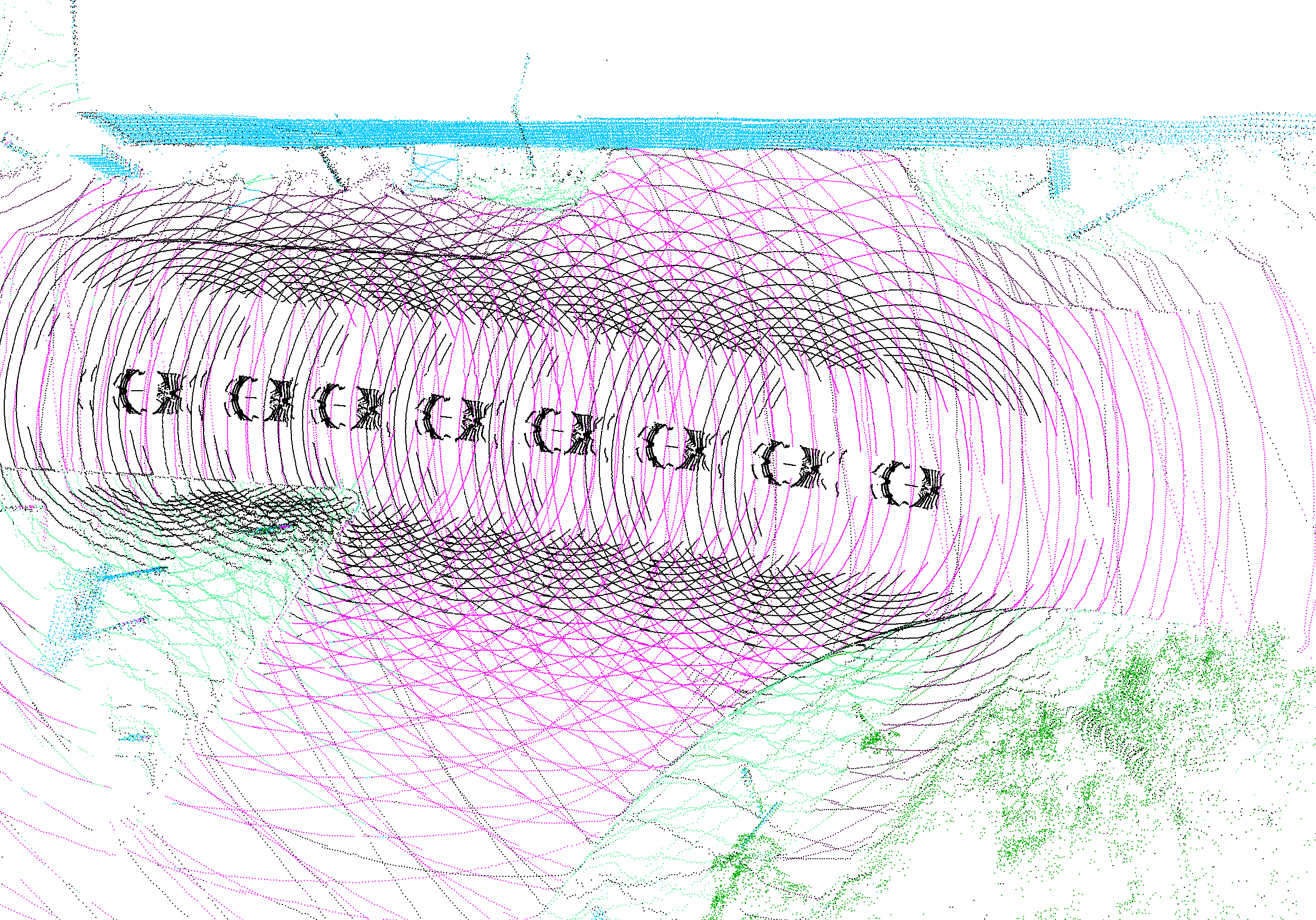}
    \includegraphics[width=0.250\textwidth]{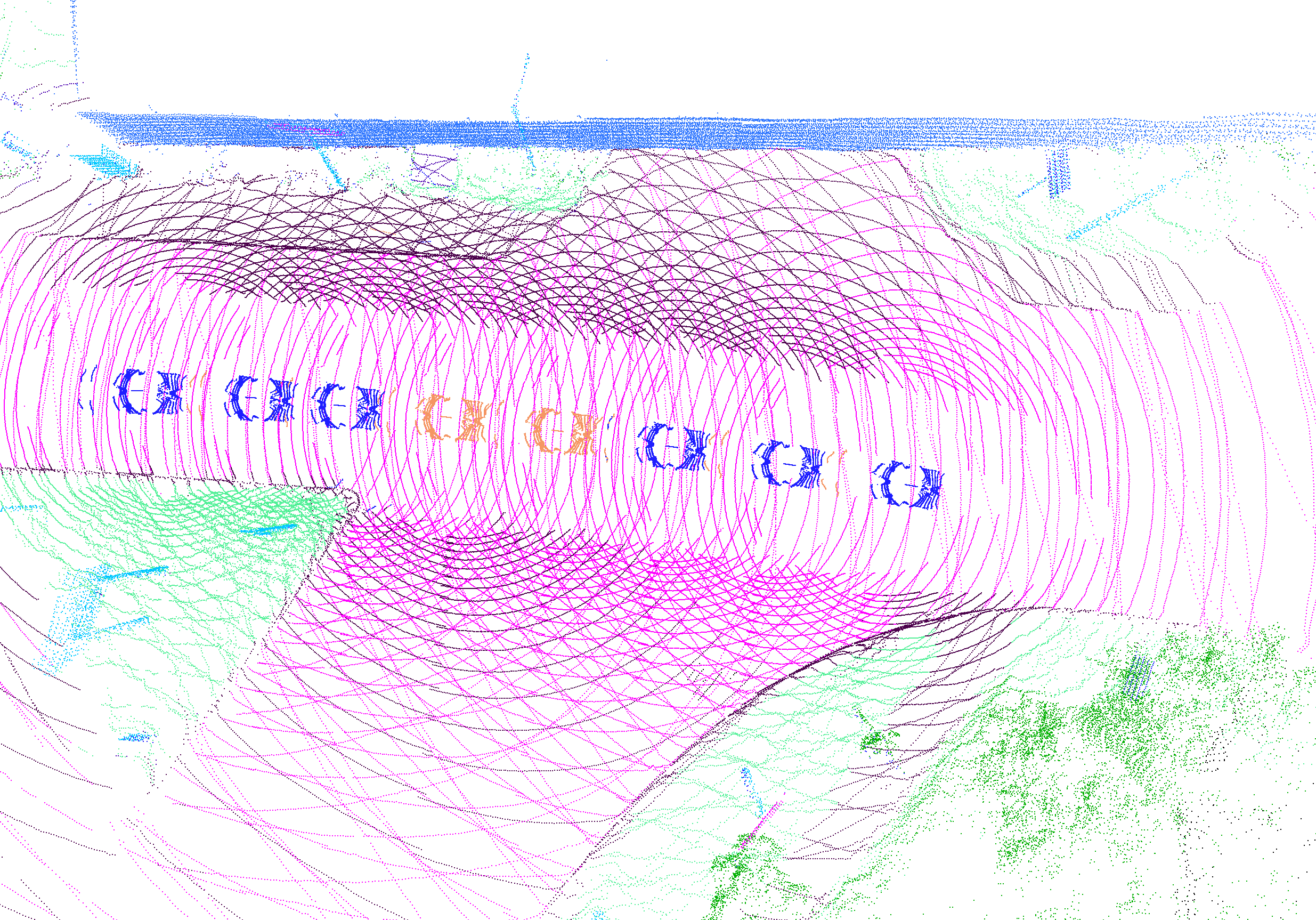}
    \\
    \includegraphics[width=0.250\textwidth]{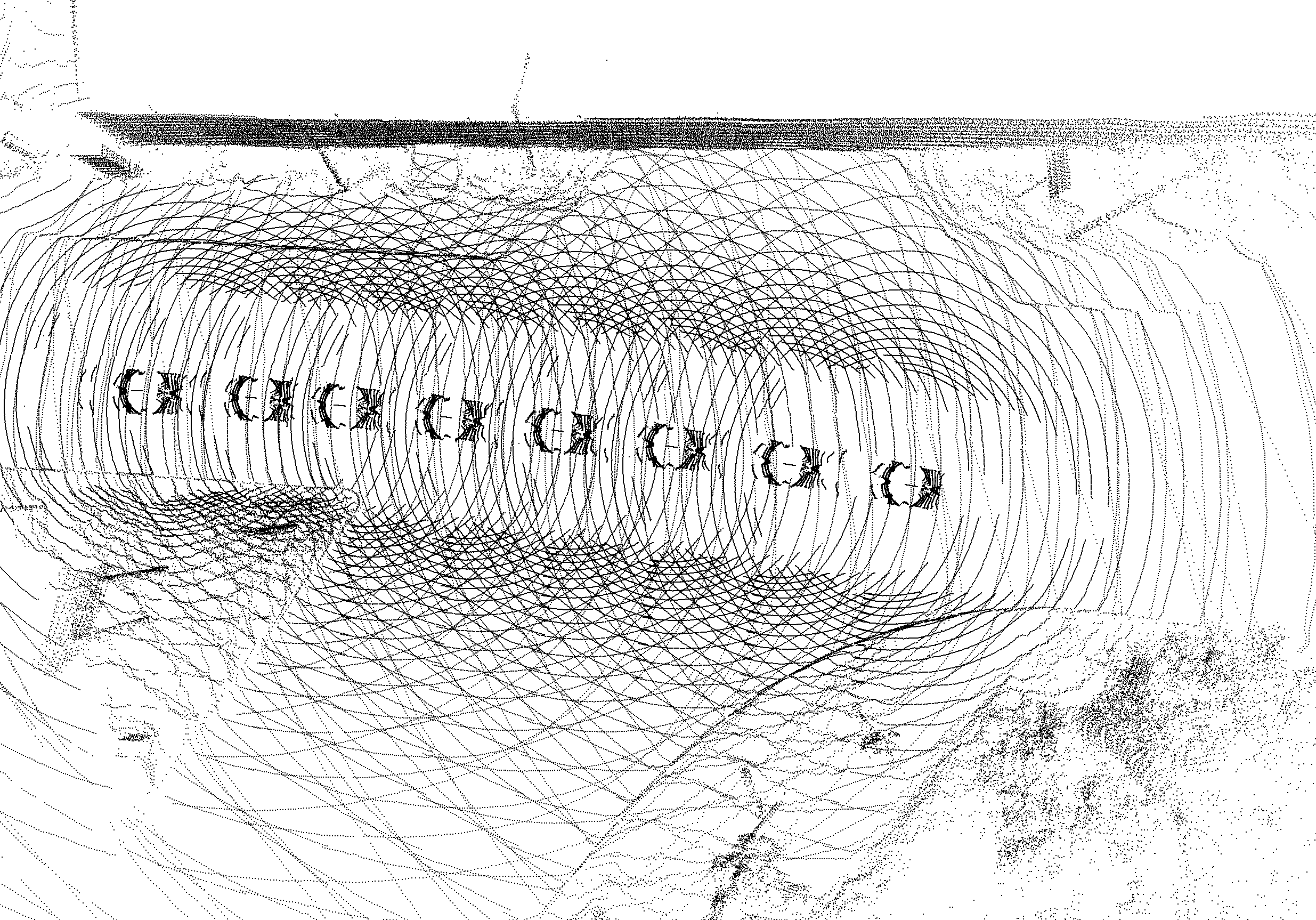}
    \includegraphics[width=0.250\textwidth]{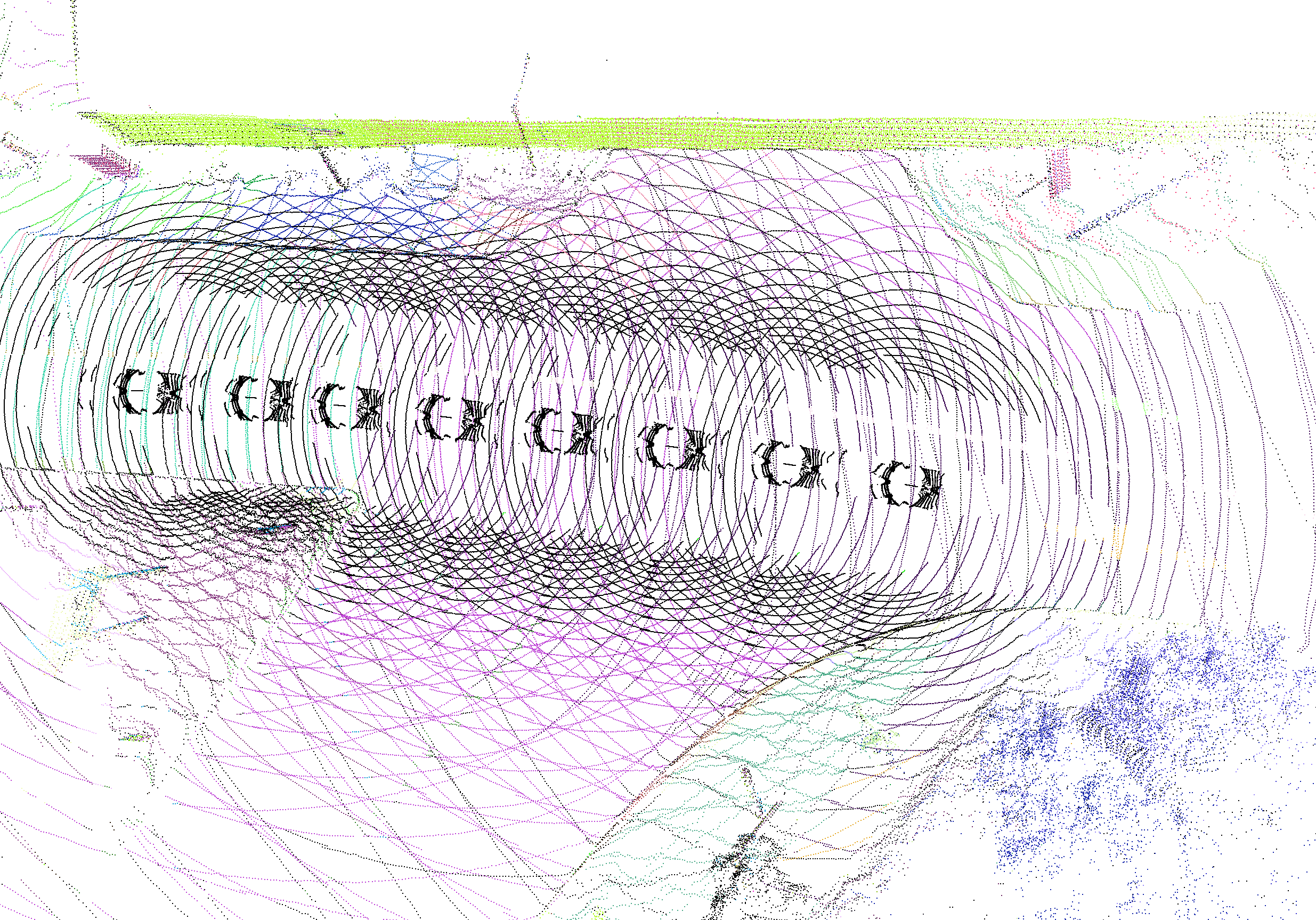}
    \includegraphics[width=0.250\textwidth]{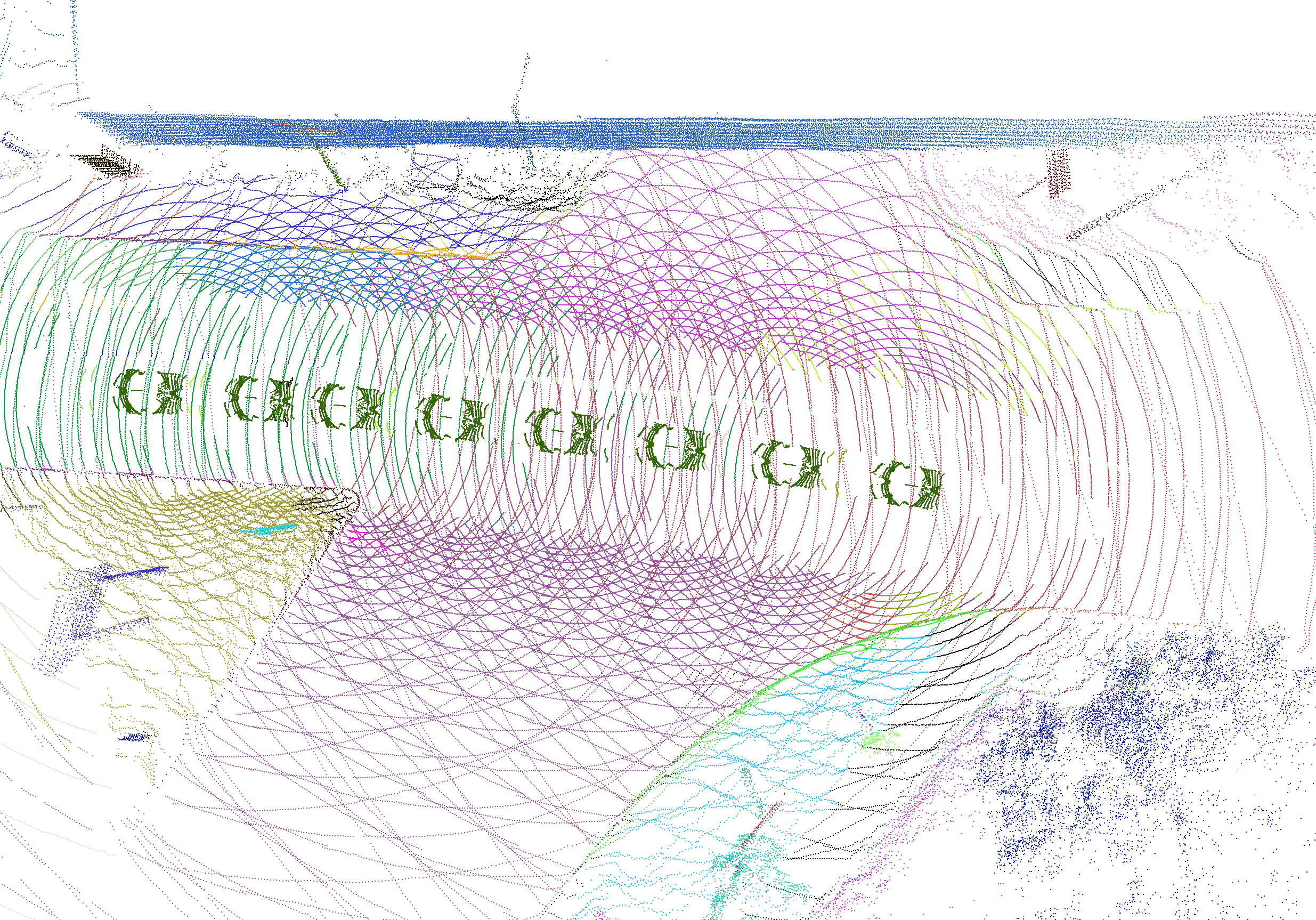}
    \\
    \makebox[0.250\textwidth]{\footnotesize GT}
    \makebox[0.250\textwidth]{\footnotesize Pseudo-labels}
    \makebox[0.250\textwidth]{\footnotesize \method}
    \\
    \includegraphics[width=0.250\textwidth]{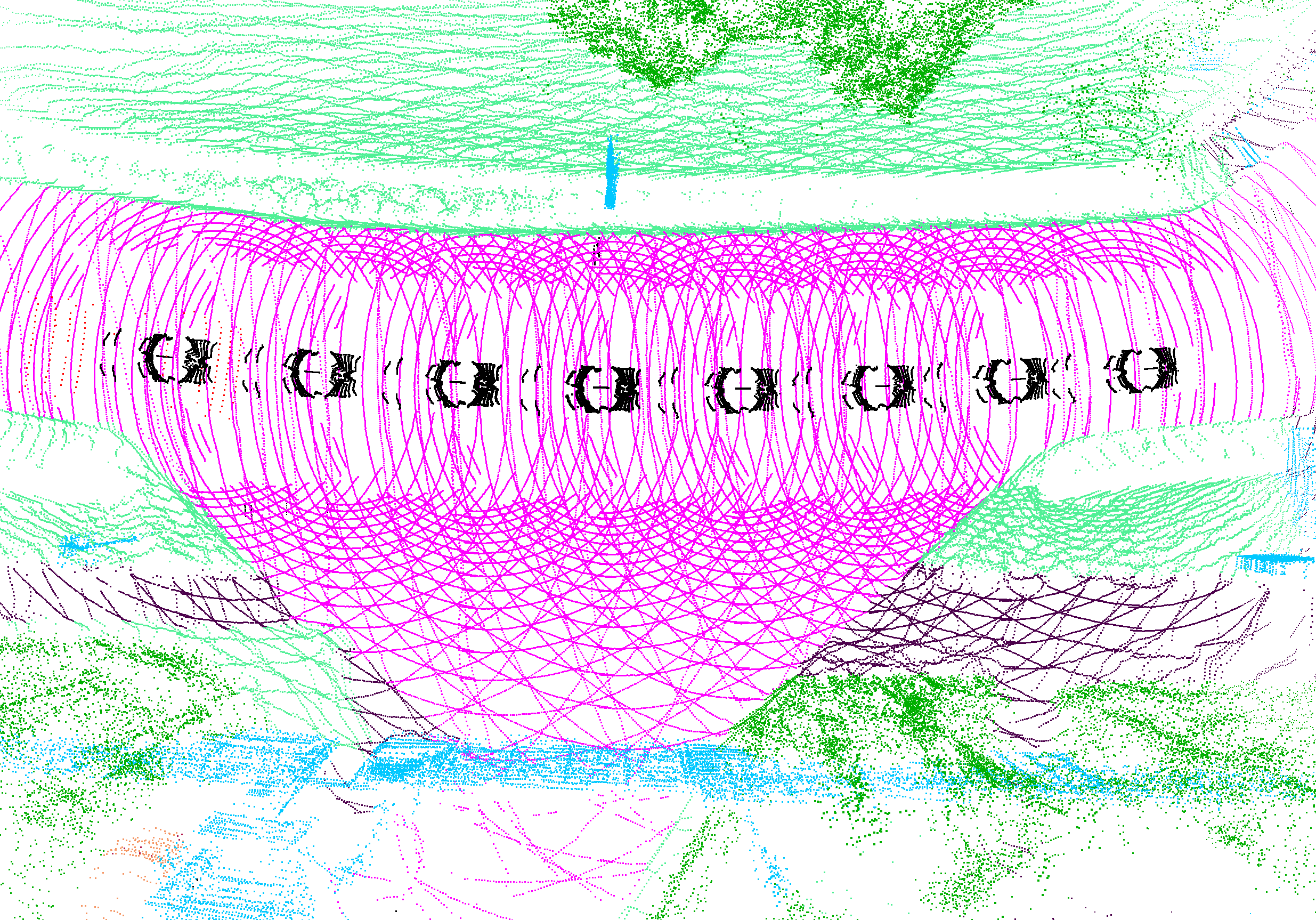}
    \includegraphics[width=0.250\textwidth]{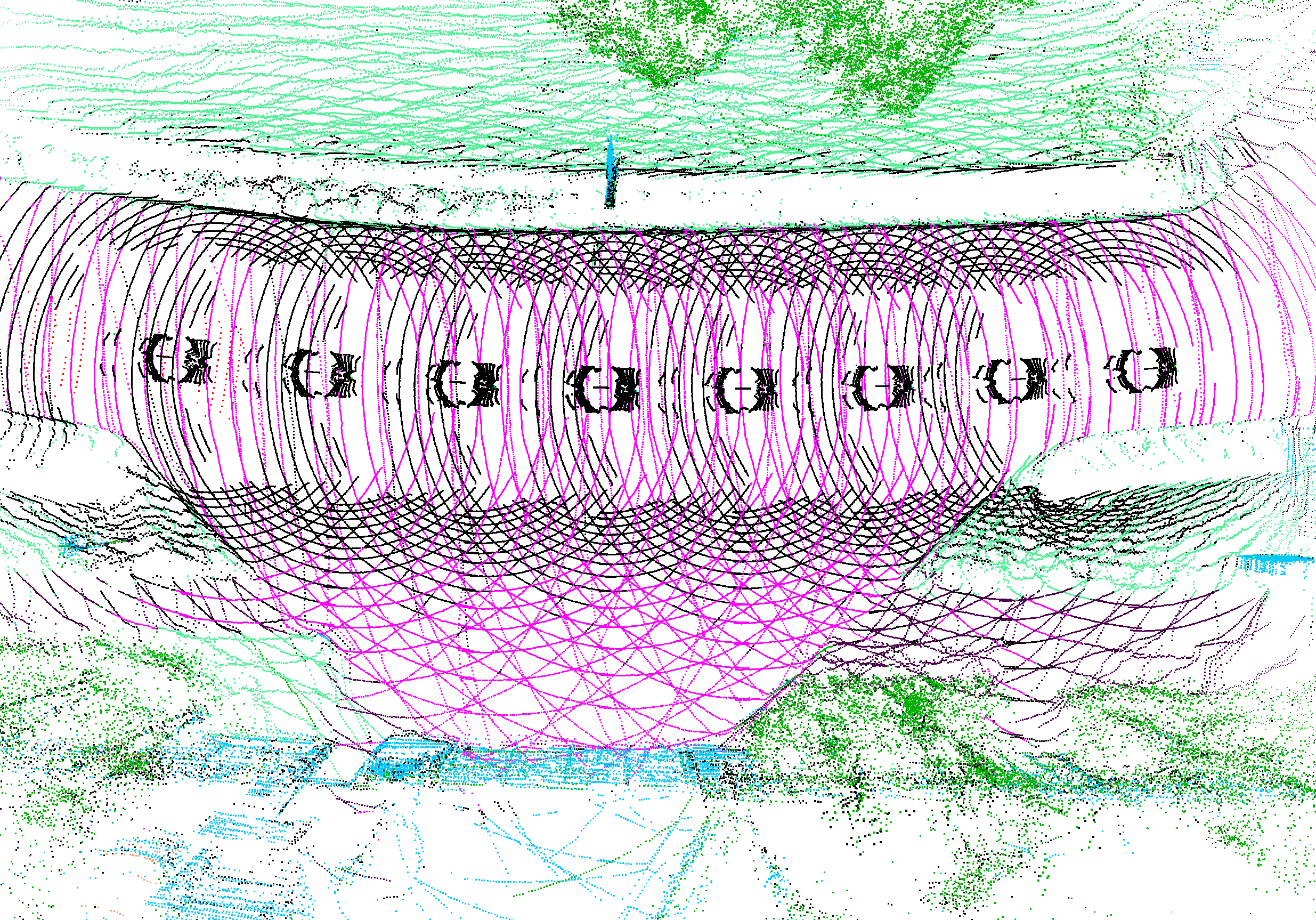}
    \includegraphics[width=0.250\textwidth]{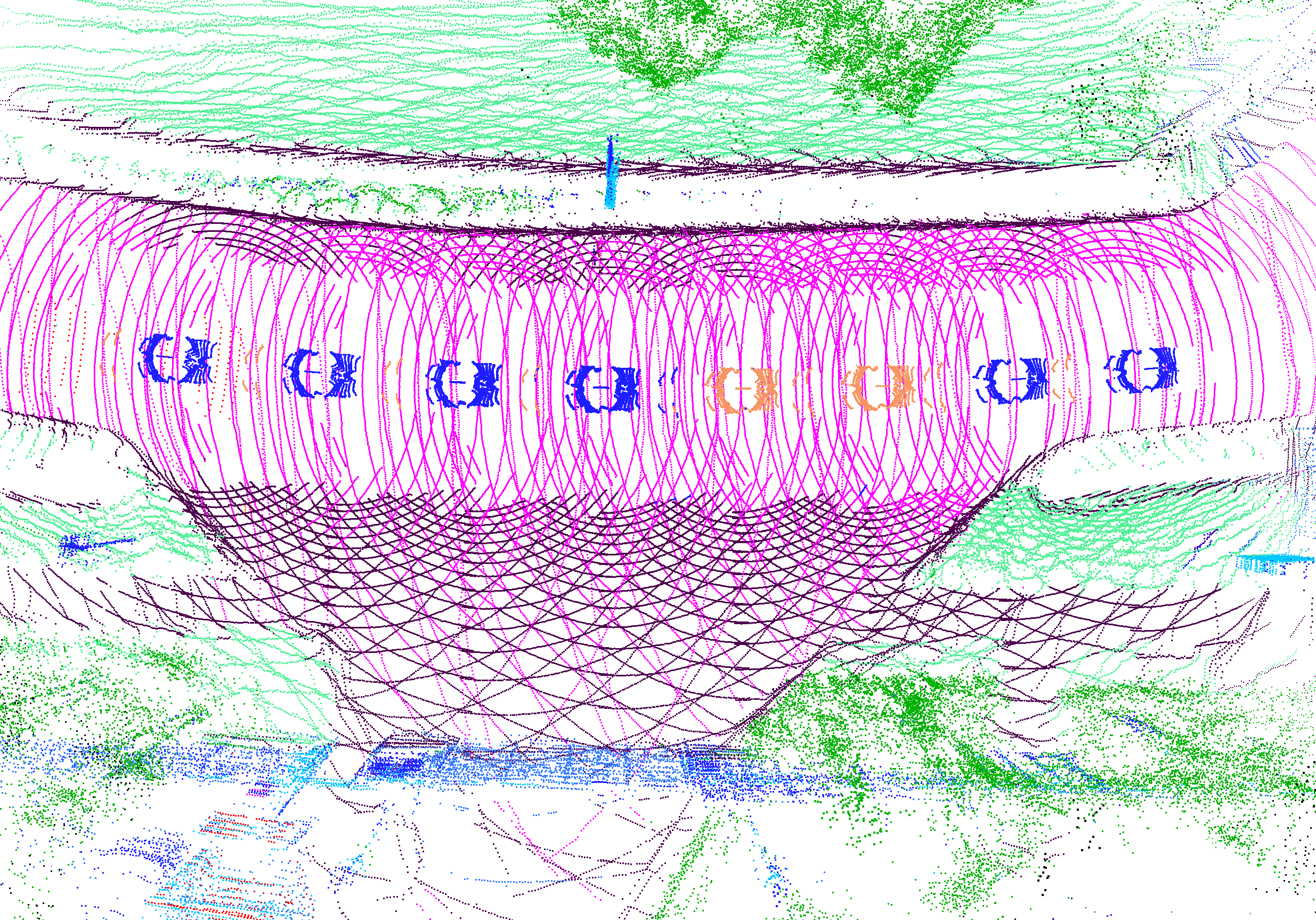}
    \\
    \includegraphics[width=0.250\textwidth]{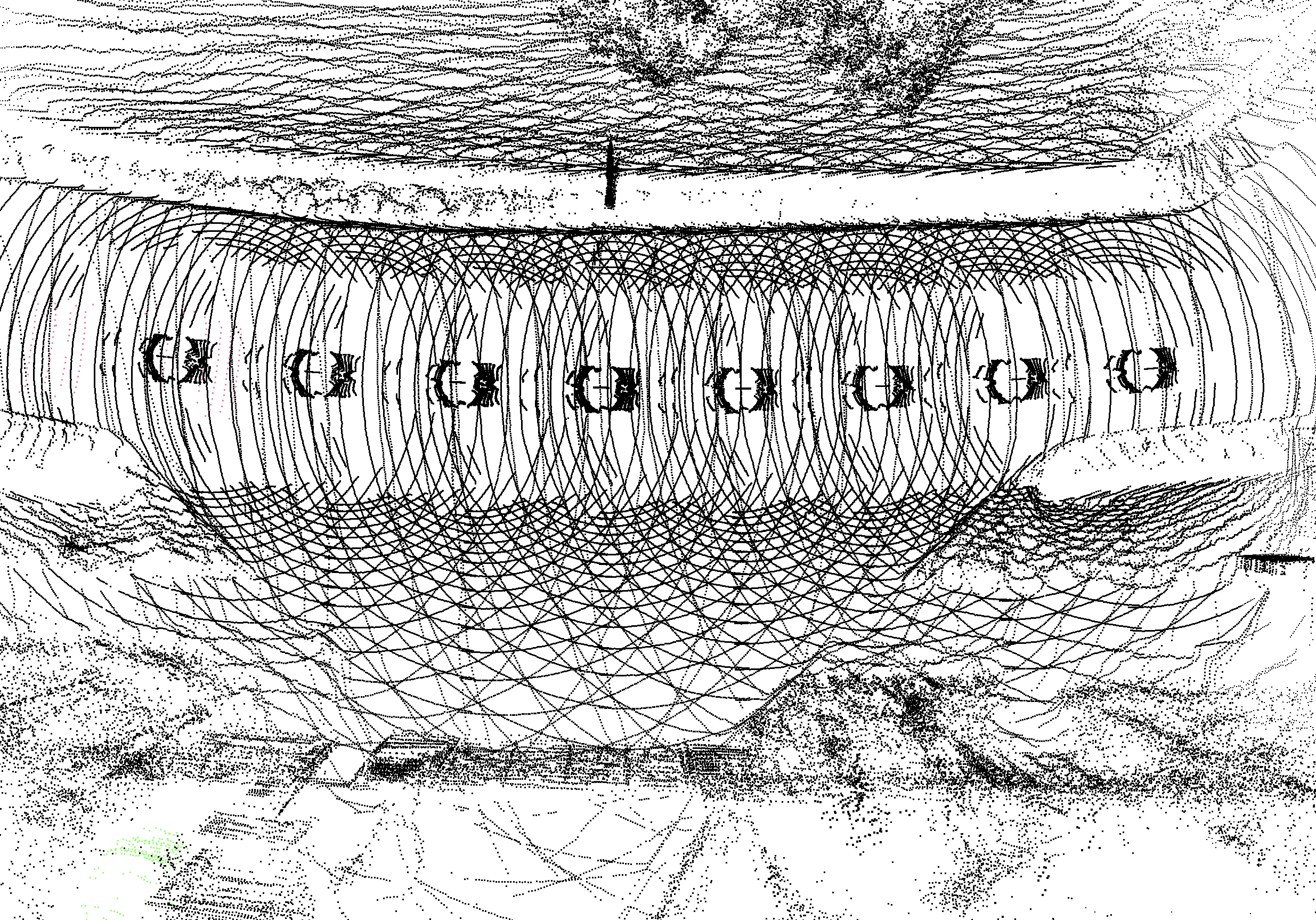}
    \includegraphics[width=0.250\textwidth]{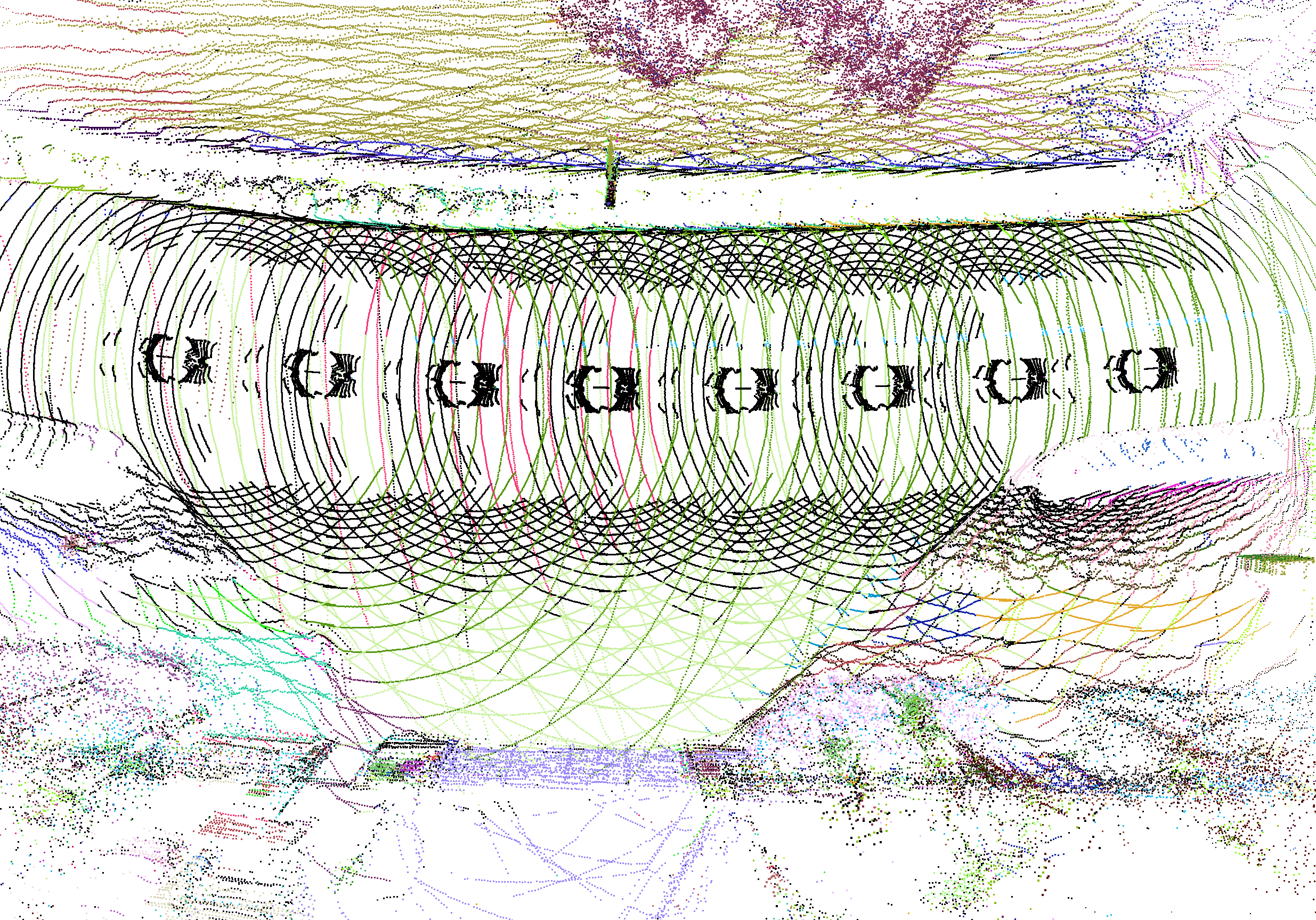}
    \includegraphics[width=0.250\textwidth]{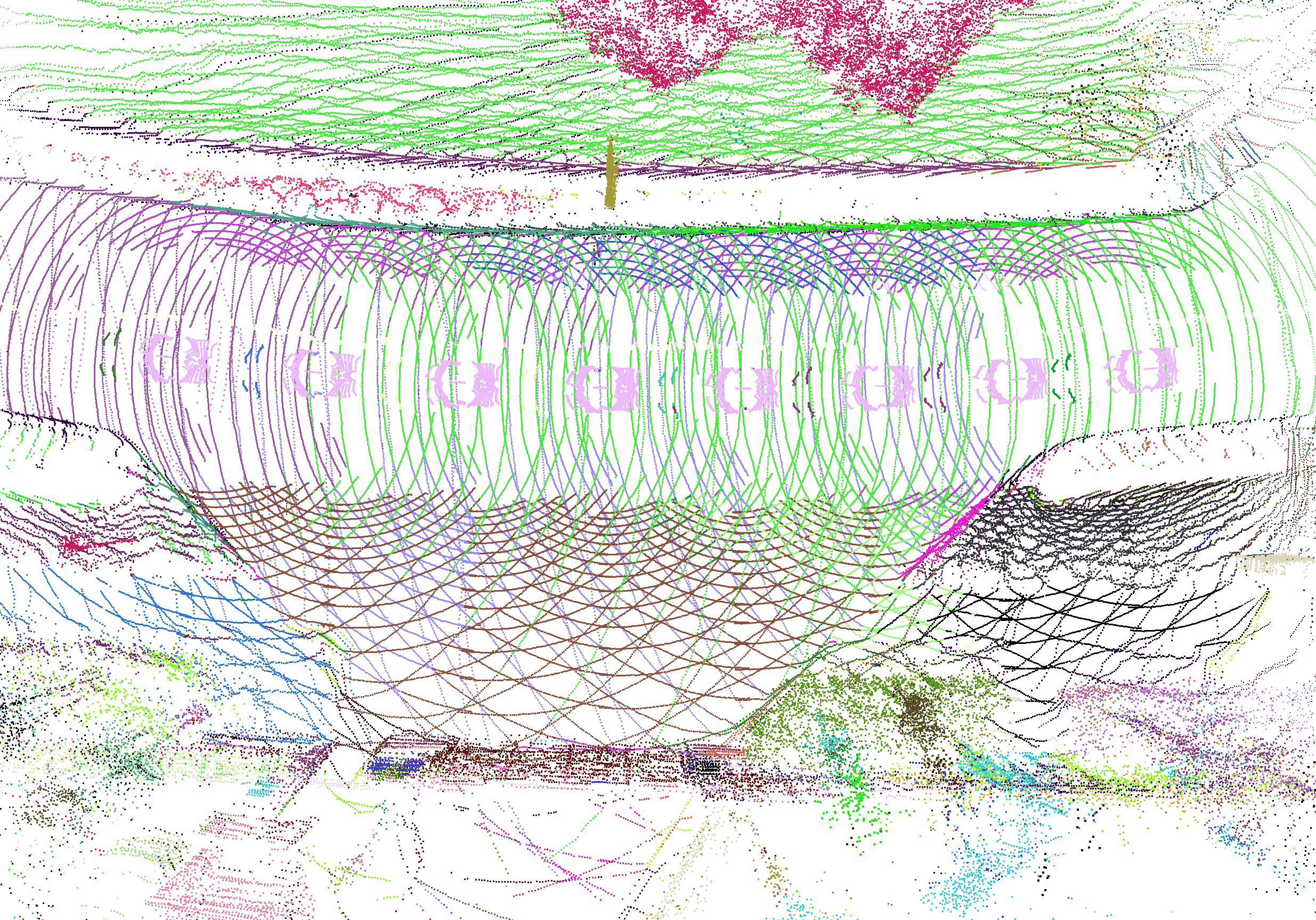}
    \\
    \caption{\textbf{Qualitative results on Panoptic nuScenes}. We show ground-truth (GT) labels (\textit{first column}), our pseudo-labels (\textit{middle column}), and \method results (\textit{right column}). We show three scenes (we superimpose point clouds). For each, we show semantics predictions in the \textit{first row} and instances predictions in the \textit{second row}. \textbf{Importantly, we visualize semantics for pseudo-labels via zero-shot prompting; pseudo-labels do not provide explicit semantic labels, only CLIP tokens.} In nuScenes, points also reflect from the ego-vehicle (seen as a car-shaped object in the center, replicated along the trajectory when the vehicle is moving; see $2^{nd}$ and $3^{rd}$ scene examples).}
    \label{fig:viz_nuscenes}
\end{figure*}

\PAR{Zero-Shot 4D Lidar Panoptic Segmentation.} In \cref{fig:viz_kitti} and \cref{fig:viz_nuscenes}, we visualize ground-truth labels (GT) (\textit{left}), pseudo-labels (\textit{center}), and \method results (\textit{right}) on \textit{SemanticKITTI} and \textit{Panoptic nuScenes}, respectively. 
We visualize three different scenes per dataset, shown as superimposed point clouds. In the \textit{top} row, we visualize semantics, and in the \textit{bottom} row, we visualize (4D) instances. 
\textbf{Importantly, to visualize semantic classes, we prompt individual instances with test-time specified prompts that conform to class vocabularies of \textit{SemanticKITTI} and \textit{Panoptic nuScenes}, respectively. Neither pseudo-labels nor our model has any explicit semantic information about these object classes.}
As can be seen, GT labels provide instance labels only for specific \thing classes, whereas our pseudo-labels and model predictions densely segment point clouds consistently in space and time. 

Our pseudo-labels only cover a small portion of the point cloud ($14\%$); however, our model learns to segment \textit{full} point clouds.
\cref{tab:4dsal_model_ablation_outsidefrustum} confirms that we can achieve such a generalization using suitable data augmentations. 

\begin{figure*}[t] \centering
    \makebox[0.24\textwidth]{\scriptsize \texttt{Advertising Stand}}
    \makebox[0.24\textwidth]{\scriptsize \texttt{Car}}
    \makebox[0.24\textwidth]{\scriptsize \texttt{Bicycle Rider}}
    \makebox[0.24\textwidth]{\scriptsize \texttt{Electric Street Box}}
    \\
    \includegraphics[width=0.24\textwidth]{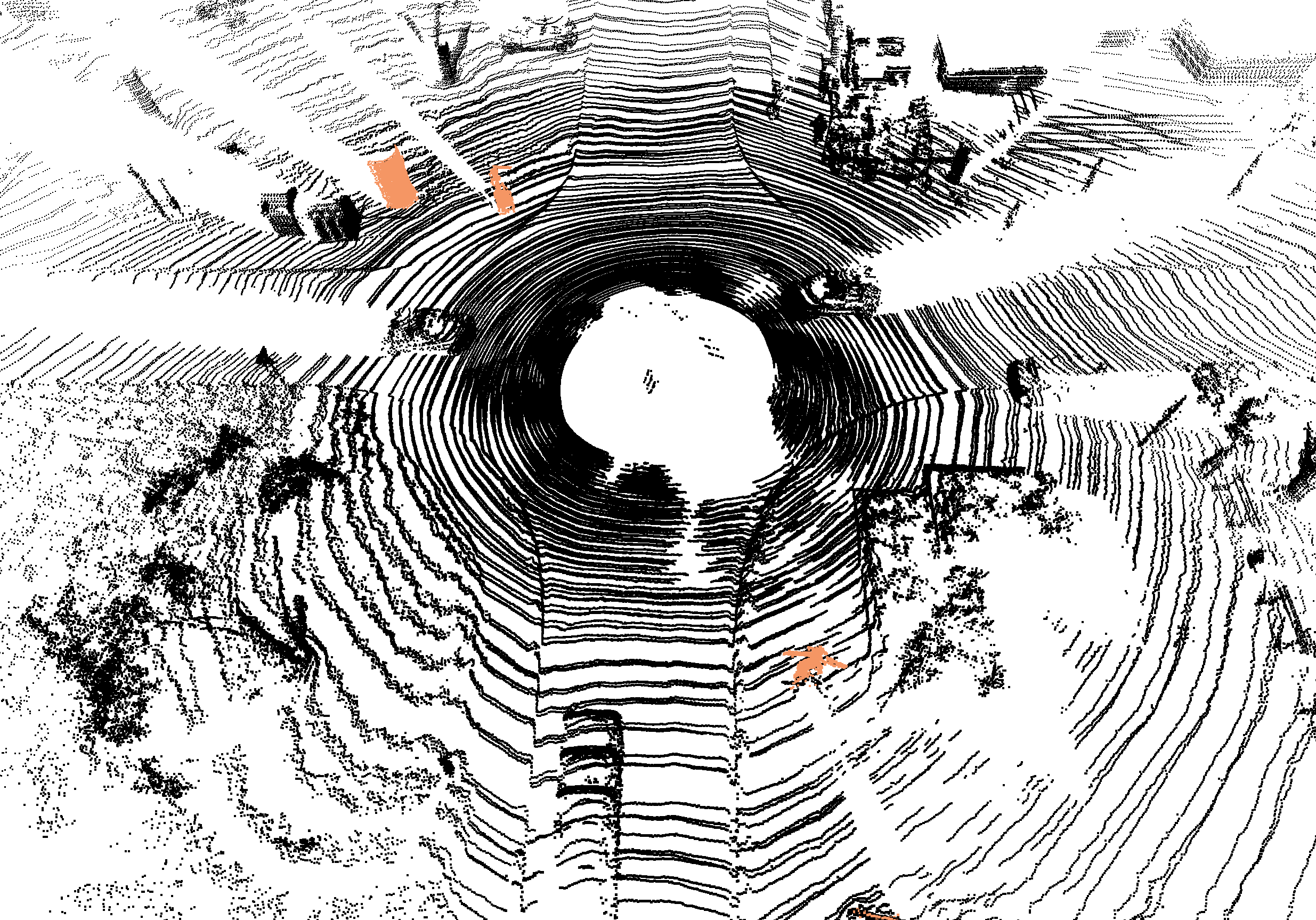}
    \includegraphics[width=0.24\textwidth]{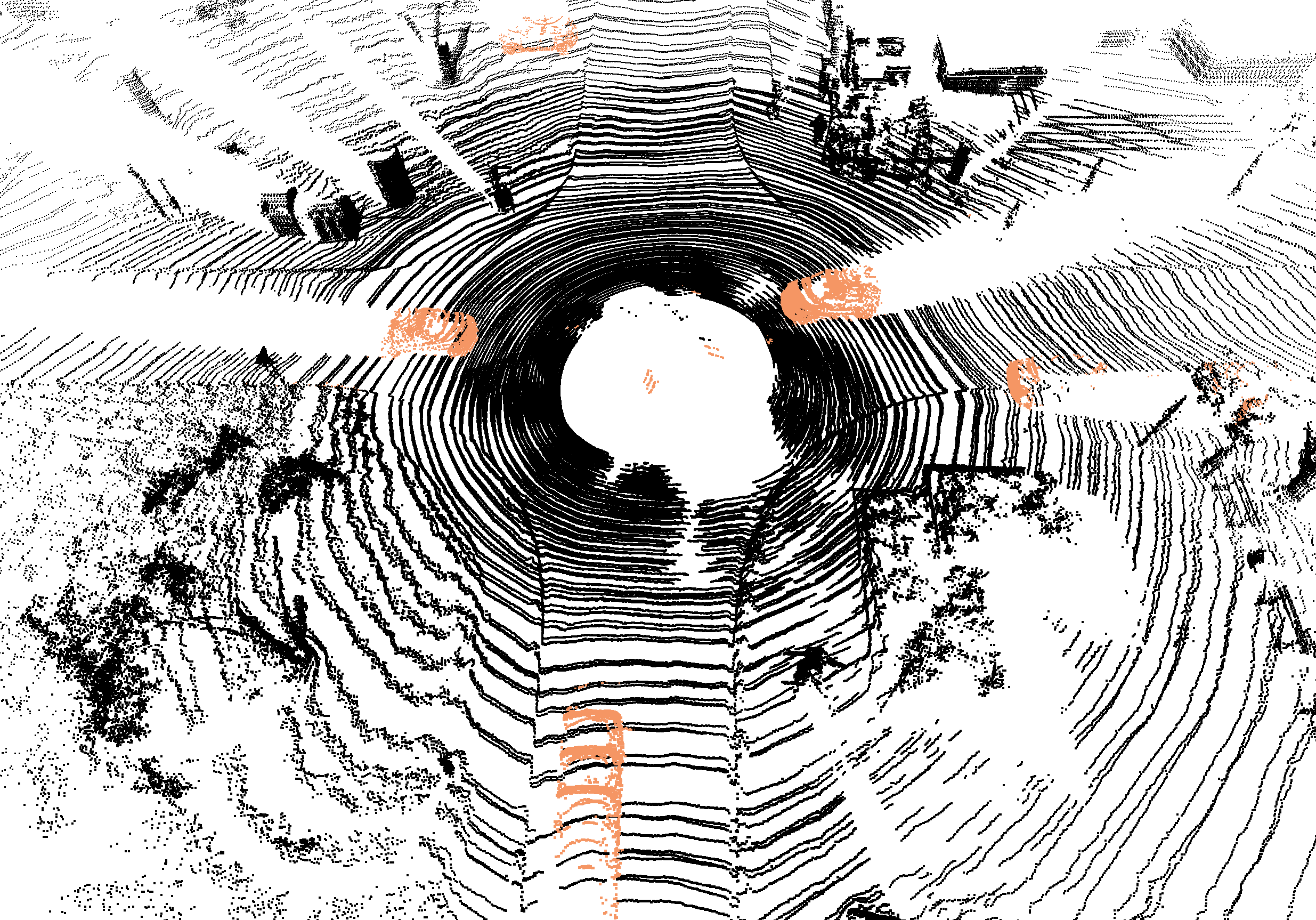}
    \includegraphics[width=0.24\textwidth]{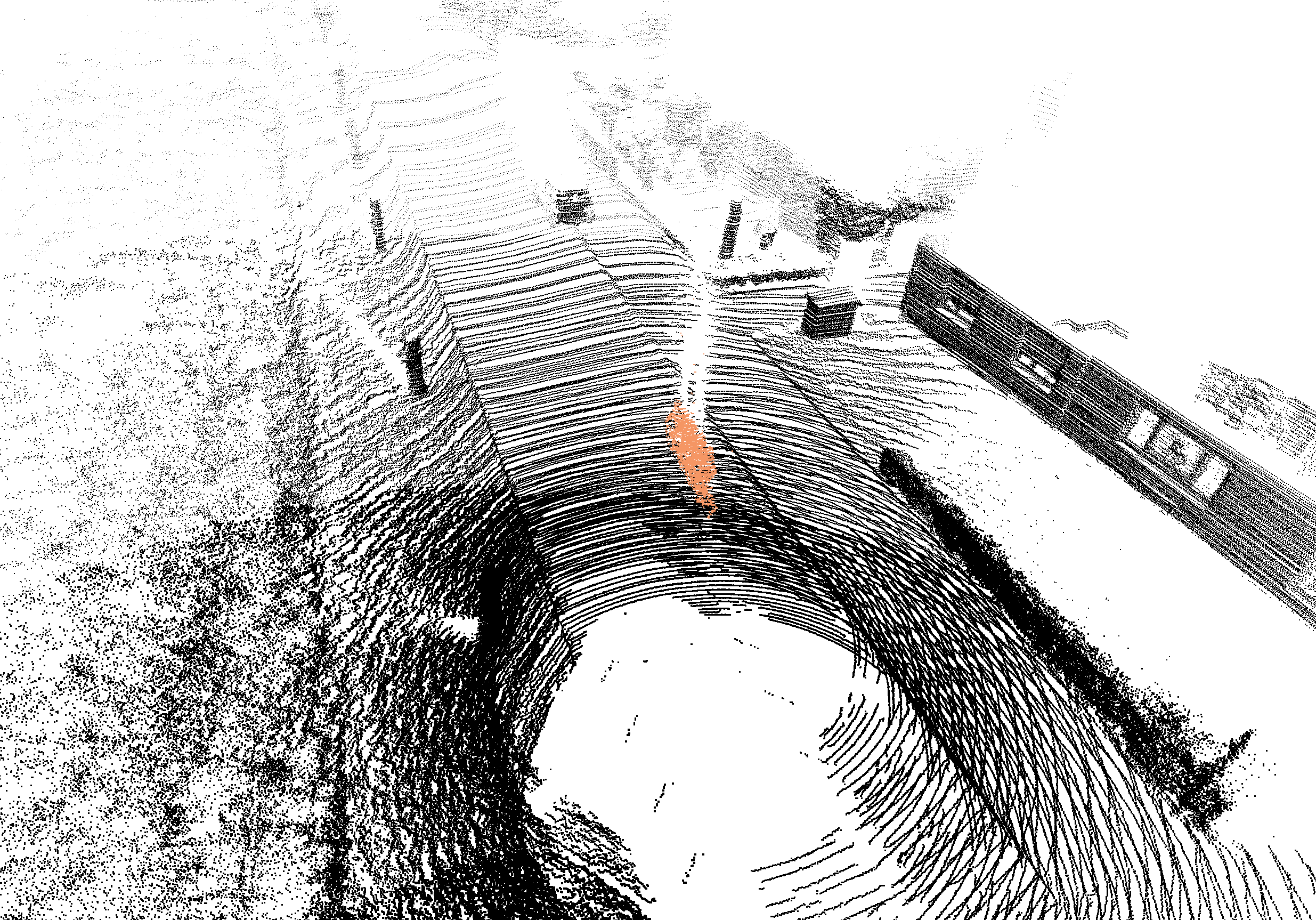}
    \includegraphics[width=0.24\textwidth] {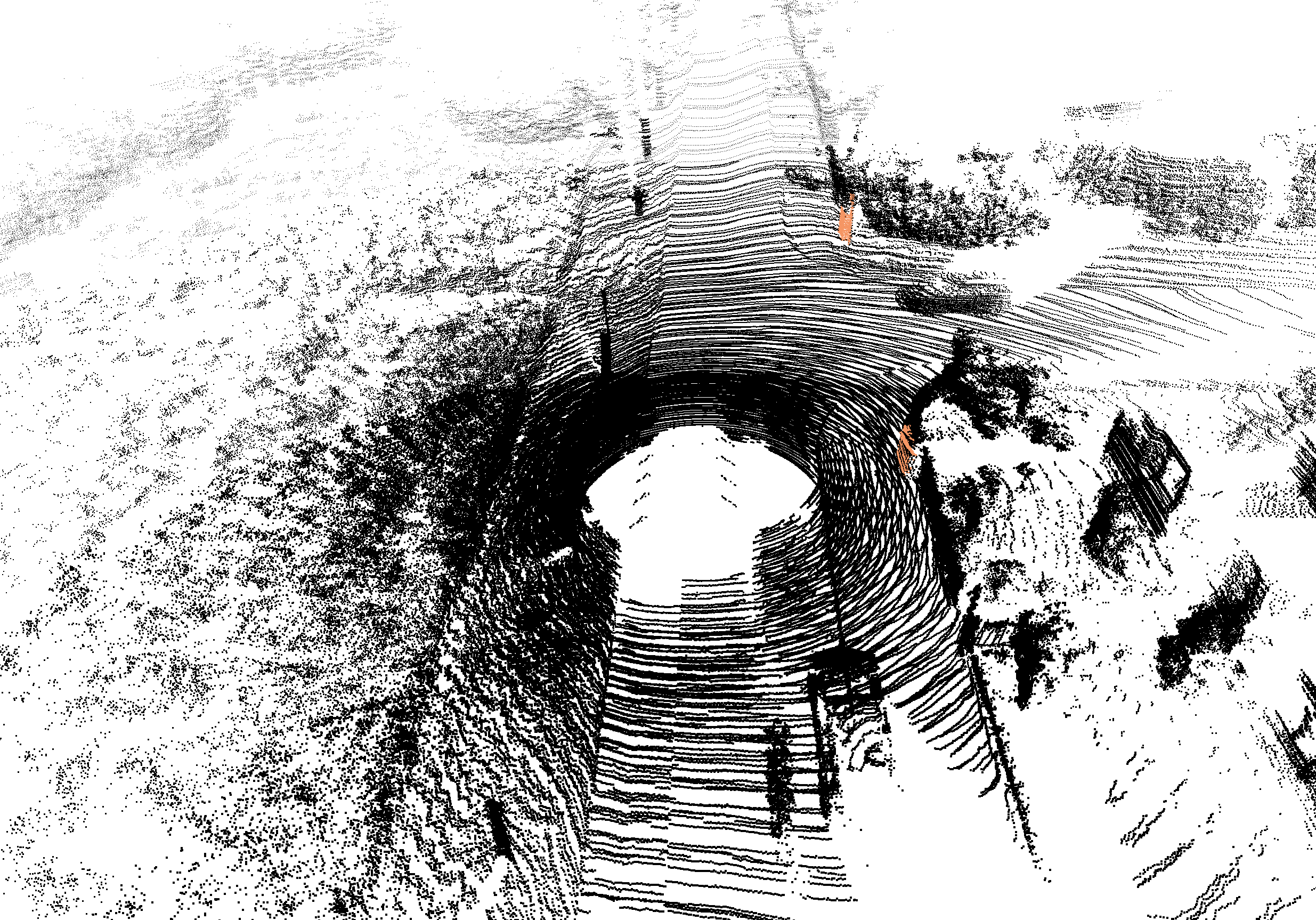}
    \\
    \includegraphics[width=0.24\textwidth]{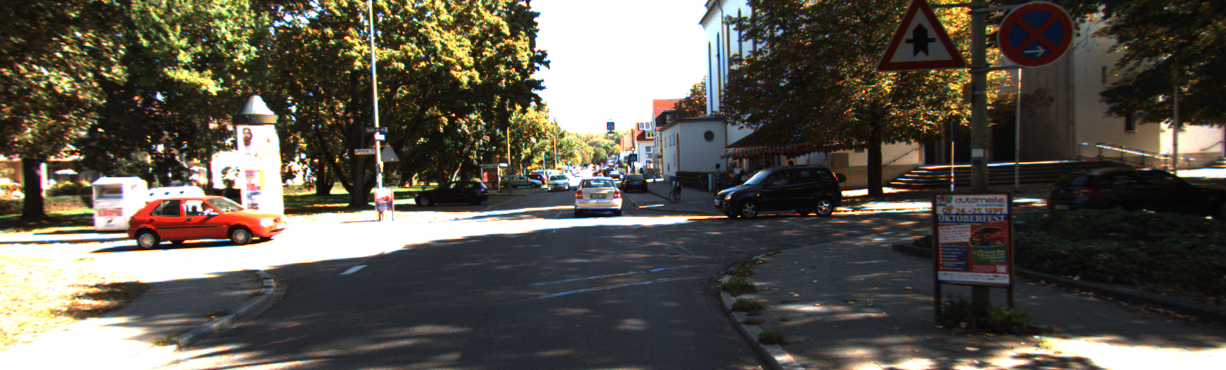}
    \includegraphics[width=0.24\textwidth]{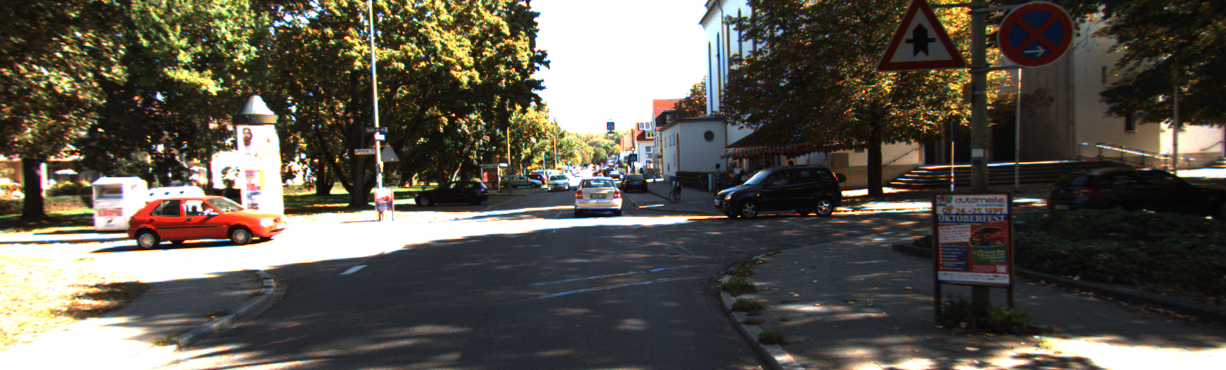}
    \includegraphics[width=0.24\textwidth]{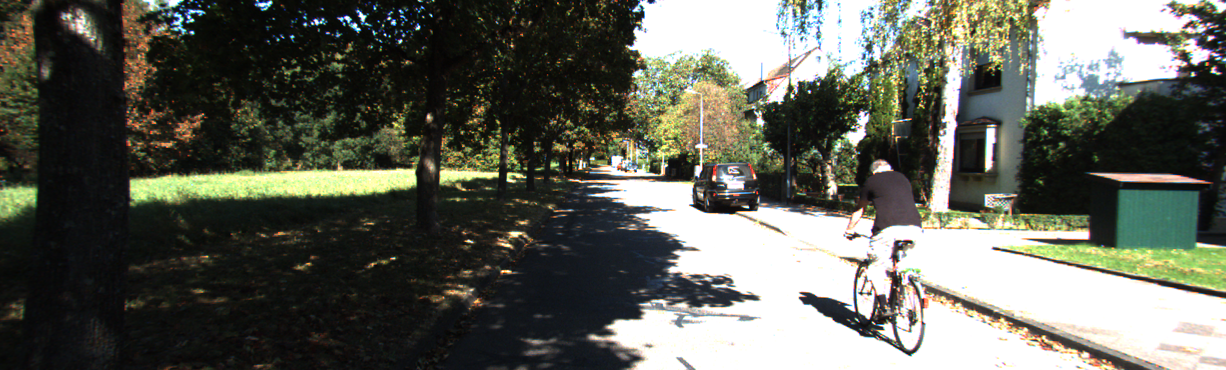}
    \includegraphics[width=0.24\textwidth]{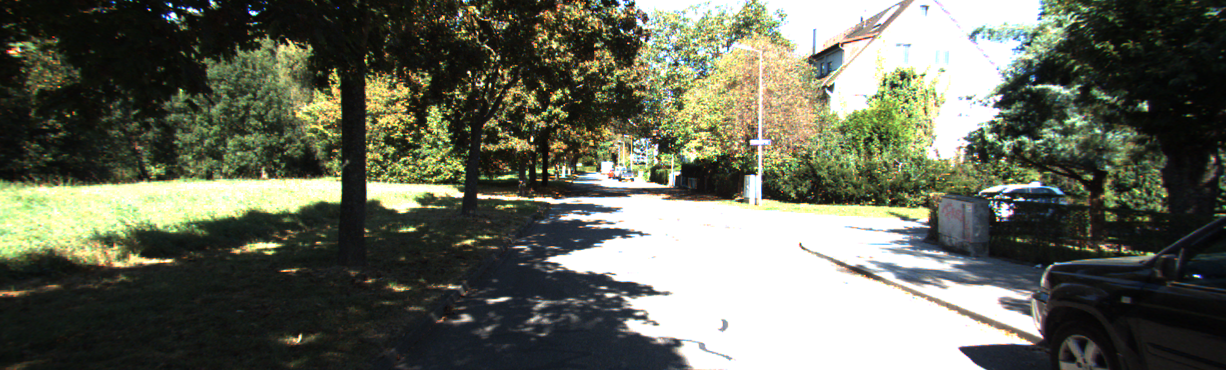}
    \caption{\textbf{Prompt examples}. We visualize the output of our model (we highlight objects in \textbf{\textcolor{orange}{orange}}) for four different prompts: two canonical \texttt{car} and \texttt{bicycle rider}, and two ``arbitrary'' object, \texttt{advertising stand} and \texttt{electric street box}. As can be seen, all are segmented correctly, including stationary and moving instances. \textbf{Remarkably, all three different types of \texttt{advertising stand}, and both instances of \texttt{electric street box} are correctly segmented.} We provide images for reference; images are \textit{not} used as input to our model. \textit{Best seen in color, zoomed.} } 
    \label{fig:prompt_visualization}
\end{figure*}

\PAR{Arbitrary prompts.} 
We report additional qualitative results with arbitrary text prompts in \cref{fig:prompt_visualization}. 
In particular, we specify single-class prompts and highlight objects in \textbf{\textcolor{orange}{orange}}) for four different prompts. Two are canonical objects (\texttt{car} and \texttt{bicycle rider}), and two are not parts of standard class vocabularies in Lidar segmentation: \texttt{advertising stand} and \texttt{electric street box}. Nevertheless, our \method segments all objects correctly (three different types of advertisement stands and two electric boxes). We provide images only for reference.

\vspace{10cm}
\begin{table*}[t]
    \centering
    \footnotesize
    \setlength{\tabcolsep}{2.0pt}
    \resizebox{\textwidth}{!}{
    \begin{tabular}{HlH | HHHH HHH HHH H c | ccccccccccccccccccc}
        \toprule
        & & & \multicolumn{31}{c}{SemanticKITTI~\cite{behley2019iccv}} \\
        \cmidrule{4-34}
        & Method & \rotatebox{90}{Supervision} & PQ & PQ\textsuperscript{\textdagger} & RQ & SQ & PQ\textsuperscript{Th} & RQ\textsuperscript{Th} & SQ\textsuperscript{Th} & PQ\textsuperscript{St} & RQ\textsuperscript{St} & SQ\textsuperscript{St} & mIoU & \rotatebox{90}{\texttt{all}} & \rotatebox{90}{\texttt{car}} & \rotatebox{90}{\texttt{bicycle}} & \rotatebox{90}{\texttt{motorcycle}} & \rotatebox{90}{\texttt{truck}} & \rotatebox{90}{\texttt{other-vehicle}} & \rotatebox{90}{\texttt{person}} & \rotatebox{90}{\texttt{bicyclist}} & \rotatebox{90}{\texttt{motorcyclist}} & \rotatebox{90}{\texttt{road}} & \rotatebox{90}{\texttt{parking}} & \rotatebox{90}{\texttt{sidewalk}} & \rotatebox{90}{\texttt{other-ground}} & \rotatebox{90}{\texttt{building}} & \rotatebox{90}{\texttt{fence}} & \rotatebox{90}{\texttt{vegetation}} & \rotatebox{90}{\texttt{trunk}} & \rotatebox{90}{\texttt{terrain}} & \rotatebox{90}{\texttt{pole}} & \rotatebox{90}{\texttt{traffic-sign}} \\
        \midrule
        & SAL  & ZS & - & - & - & - & - & - & - & - & - & - & - & $25.3$ & $78.8$ & $18.2$ & $20.3$ & $7.5$ & $8.7$ & $12.6$ & $0.0$ & $0.0$ & $70.3$ & $3.2$ & $28.7$ & $0.0$ & $44.6$ & $3.2$ & $76.5$ & $18.8$ & $30.0$ & $33.6$ & $24.6$ \\
        & \method  & ZS & - & - & - & - & - & - & - & - & - & - & - & $30.8$ & $84.3$ & $26.9$ & $26.7$ & $15.5$ & $16.2$ & $11.9$ & $21.0$ & $1.7$ & $74.1$ & $3.0$ & $33.4$ & $0.0$ & $62.5$ & $9.2$ & $82.4$ & $14.1$ & $35.7$ & $37.3$ & $28.9$ \\ 
        \midrule
        & & & \multicolumn{31}{c}{nuScenes~\cite{fong21ral}} \\
        \cmidrule{4-34}
        &  & & PQ & PQ\textsuperscript{\textdagger} & RQ & SQ & PQ\textsuperscript{Th} & RQ\textsuperscript{Th} & SQ\textsuperscript{Th} & PQ\textsuperscript{St} & RQ\textsuperscript{St} & SQ\textsuperscript{St} & mIoU & \rotatebox{90}{\texttt{all}} & \rotatebox{90}{\texttt{barrier}} & \rotatebox{90}{\texttt{bicycle}} & \rotatebox{90}{\texttt{bus}} & \rotatebox{90}{\texttt{car}} & \rotatebox{90}{\texttt{construction\_vehicle}} & \rotatebox{90}{\texttt{motorcycle}} & \rotatebox{90}{\texttt{pedestrian}} & \rotatebox{90}{\texttt{traffic\_cone}} & \rotatebox{90}{\texttt{trailer}} & \rotatebox{90}{\texttt{truck}} & \rotatebox{90}{\texttt{driveable\_surface}} & \rotatebox{90}{\texttt{other\_flat}} & \rotatebox{90}{\texttt{sidewalk}} & \rotatebox{90}{\texttt{terrain}} & \rotatebox{90}{\texttt{manmade}} & \rotatebox{90}{\texttt{vegetation}} \\
        \midrule
        & SAL  & ZS & - & - & - & - & - & - & - & - & - & - & - & $41.2$ & $0.6$ & $32.8$ & $60.3$ & $82.9$ & $26.4$ & $48.8$ & $57.3$ & $42.5$ & $31.3$ & $53.1$ & $63.1$ & $1.6$ & $16.3$ & $36.6$ & $33.3$ & $71.7$ \\
        & \method  & ZS & - & - & - & - & - & - & - & - & - & - & - & $45.7$ & $1.1$ & $68.1$ & $60.8$ & $85.3$ & $32.2$ & $73.7$ & $62.3$ & $37.2$ & $33.9$ & $56.4$ & $56.6$ & $0.1$ & $13.7$ & $39.4$ & $35.5$ & $75.0$ \\ 
        \bottomrule
    \end{tabular}
    } %
    \caption{
        \textbf{Per-class (zero-shot) results (PQ) for \method and SAL~\cite{sal2024eccv} on \textit{SemanticKITTI} and \textit{nuScenes-Panoptic} validation sets.} Our \method consistently outperforms SAL on (almost) all classes. Due to limited temporal context, SAL fails to segment smaller objects such as \texttt{motorcyclist}, \texttt{cyclist}, \texttt{barrier}. \method substantially improves segmentation of such objects.
    }
    \label{tab:kitti_nuscenes-baselines-per_class}
\end{table*}

\end{document}